\documentclass[journal]{IEEEtran}
\usepackage{cite}
\usepackage{graphicx}
\usepackage[hidelinks]{hyperref}
\usepackage{academicons}
\usepackage{svg}
\newcommand{\orcid}[1]{\href{https://orcid.org/#1}{\includesvg[width=10pt]{figs/orcid}}}
\usepackage{xcolor}
\usepackage{soul}
\usepackage{amsmath}
\usepackage{amssymb}

\usepackage{multirow}
\usepackage{tabularx}
\usepackage{pifont}
\newcommand{\cmark}{\ding{51}}%
\newcommand{\xmark}{\ding{55}}%

\begin{document}
\title{Survey of Graph Neural Network for Internet of Things and NextG Networks}

\author{Sabarish Krishna Moorthy and Jithin Jagannath, ~\IEEEmembership{Senior~Member,~IEEE}\\
$^{a}$Marconi-Rosenblatt AI Innovation Laboratory, ANDRO Computational Solutions LLC, NY, USA\\
$^{b}$Department of Electrical Engineering University at Buffalo, The State University of New York, NY, USA\\
Email: \{skrishnamoorthy, jjagannath\}@androcs.com
\thanks{ $^*$ Corresponding Author}
}

\maketitle

\begin{abstract}
The exponential increase in Internet of Things (IoT) devices coupled with 6G pushing towards higher data rates and connected devices has sparked a surge in data. Consequently, harnessing the full potential of data-driven machine learning has become one of the important thrusts. In addition to the advancement in wireless technology, it is important to efficiently use the resources available and meet the users' requirements. Graph Neural Networks (GNNs) have emerged as a promising paradigm for effectively modeling and extracting insights which inherently exhibit complex network structures due to its high performance and accuracy, scalability, adaptability, and resource efficiency. There is a lack of a comprehensive survey that focuses on the applications and advances GNN has made in the context of IoT and Next Generation (NextG) networks. To bridge that gap, this survey starts by providing a detailed description of GNN's terminologies, architecture, and the different types of GNNs. Then we provide a comprehensive survey of the advancements in applying  GNNs for IoT from the perspective of data fusion and intrusion detection. Thereafter, we survey the impact GNN has made in improving spectrum awareness. Next, we provide a detailed account of how GNN has been leveraged for networking and tactical systems. Through this survey, we aim to provide a comprehensive resource for researchers to learn more about GNN in the context of wireless networks, and understand its state-of-the-art use cases while contrasting to other machine learning approaches. Finally, we also discussed the challenges and wide range of future research directions to further motivate the use of GNN for IoT and NextG Networks.  
\end{abstract}


\IEEEpeerreviewmaketitle

\section{Introduction}

\IEEEPARstart{I}{nternet} of Things (IoT), generally defined as a network of interconnected devices, sensors, and other resources that communicate with each other over the internet. A simple IoT network consists of three layers namely the Perception or Sensing Layer, the Network Layer, and the Application Layer. The role of \textit{Perception or Sensing Layer} is to gather the data from connected sensors or actuators such as air quality monitoring sensors, motion sensors, cameras, and temperature sensors, to name a few. The \textit{Network Layer} is responsible for transmitting the collected data to the cloud or server through a centralized gateway. Examples of network layer protocols include ZigBee, Bluetooth, Long Range (LoRa), Wireless Fidelity (WiFi), Near-Field Communication
(NFC), etc. The cloud (or server) takes care of storing and transforming sensor raw data into human-understandable values as well as performing statistical analysis. The \textit{Application Layer} provides access to the processed sensor data as well as enables interaction between the user and the sensors. In a more sophisticated IoT network, there is another layer between the network and application layer called the \textit{Processing Layer}. 

\begin{figure*}[htbp]
\centering
\includegraphics[width=0.65\textwidth]{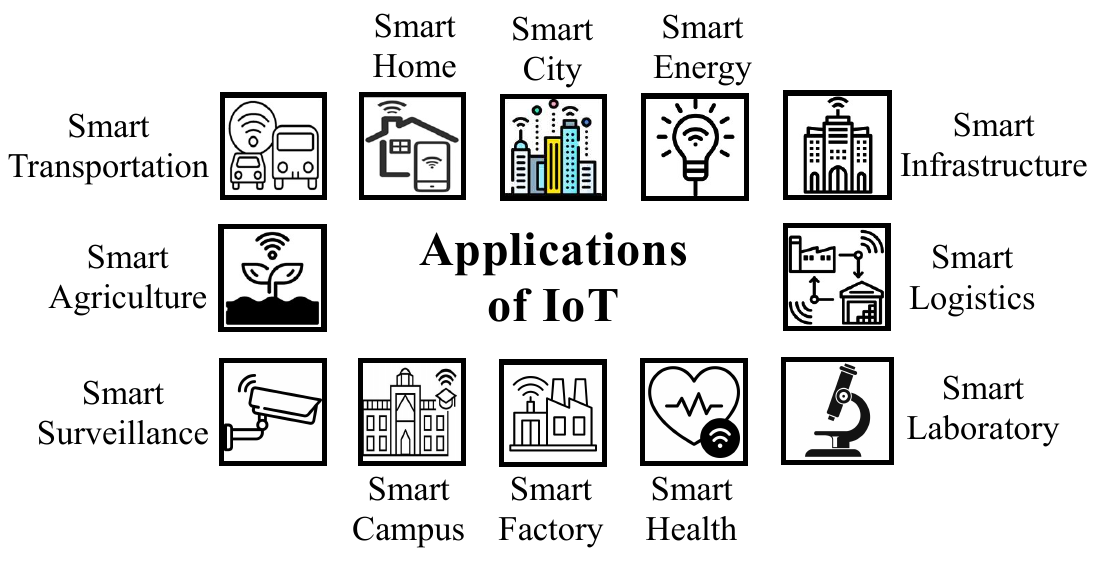}
\caption{Applications of IoT.}
\label{fig:iot_appln}
\end{figure*}

\subsection{Applications of IoT}

Over the past decade, the number of connected IoT devices has grown significantly. There are over 30 billion IoT devices, which are expected to quadruple by 2030 \cite{Zakria_icte_2023}. This can be attributed to its myriad of features such as improved quality of life \cite{Abdulmalek_Healthcare}, enhanced decision making \cite{song_cbi_2019}, safety and security \cite{Sogi_icirca_2018}, efficiency and automation \cite{Vishwakarma_iotsiu}, improved accessibility \cite{dara_assist}, business innovation and growth \cite{Timber_jbr_2021}, environmental sustainability \cite{Selwyn_epcs_2022}. These features have opened the door to the application of IoT across various domains in everyday life as shown in Figure~\ref{fig:iot_appln}. For example, IoT finds its application in smart home systems \cite{Mallikarjun_incet_2020, Mohamed_iemtronics}, smart city \cite{Bali_icacccn_2020, sheng_ieeeaccess_2020}, smart energy systems \cite{Orlando_iotj_2022, HamedaniWiseML21}, smart infrastructure \cite{Giacomo_aic_2020, Rahman_ieeeacc_2020, Le_iccc_2020}, smart logistics \cite{Maitra_ises_2020, Kamarudin_iccoins_2021}, smart laboratory \cite{Sholihul_icovet_2020,Poongothai_iciea_2018}, smart health \cite{Yeri_icirca_2020, health_mtp_2023}, smart factory \cite{islam_ictc_2020, Seiger_edocw_2020, hu_iiot_arxiv_2024}, smart campus \cite{Mircea_acc_2021}, smart surveillance \cite{Majumder_compsac_2020, Razalli_icspc_2019}, smart agriculture \cite{Haseeb_mdpis_2020, Boursianis_ieeesj_2021}, smart transportation \cite{Oliveira_iotj_2021,Sriratnasari_icimtech_2019}.

\subsection{Importance of 6G and NextG Networks}

While IoT enables a wide range of applications for consumers, it has a significant effect on the underlying network and spectrum (i.e., \textit{IoT network layer}) including but not limited to network and resource management, adversarial attacks, security and privacy concerns, and reliable quality of service (QoS). While 5G technology provided much-needed improvement over previous generations, the number of devices requiring enhanced capacity, ultra-low latency, and very high throughput continues to increase. To overcome these challenges, the research community has shifted focus to the 6G network as a necessary upgrade to match the user requirements \cite{chowdhury_ojcs_2020, mai_ccnc_2022, Ajagannath6G2020, jiang_ojcs_2021}. Compared to its predecessor, the 6G network is expected to have 10 times the peak data rate and 1 ms latency \cite{chowdhury_ojcs_2020}. It is also expected that the 6G system will have 1000 times higher simultaneous wireless connectivity compared to the 5G system \cite{chowdhury_ojcs_2020}. Millimeter wave (mmWave) and Terahertz (THz) bands will play a key role in enabling the high data rate and ultra-low latency communication links \cite{ahn_ieeewc_2023, sopin_tvt_2022, xue_comnst_2024, shafie_ieeen_2023, chen_commag_2021}. Compared to traditional cellular and WiFi networks that use 300 MHz to less than 30 GHz, mmwave will operate at 30–300 GHz while THz will operate at 0.1 THz to 10 THz.  While we move from 5G to 6G, it is important to effectively re-use the existing infrastructure instead of the need to re-deploy new hardware and other resources. Open Radio Access Network (O-RAN) is considered the solution to create more open, flexible, and interoperable mobile networks \cite{singh_wcncw_2020}. O-RAN introduces a set of open, standardized interfaces to interact, control, and collect data from every node in the network. O-RAN consists of RAN Intelligent Controllers (RICs) that execute third-party applications to control the RAN functionalities \cite{doro2022orchestran}. Common RAN functionalities include resource management, mobility management, and QoS management to name a few.

\subsection{Era of Machine Learning}

In the recent decade due to the exponential increase in computational resources and availability of digital data, Machine Learning (ML) has been making a significant impact in the research community \cite{Mahmudul_iot_19, Jagannath19MLBook, jagannath2019machine}. This is due to the vast potential of ML techniques such as automation of tasks \cite{Moubayed_mdpifi_2022}, enhanced user experience \cite{chi_commag_2023}, resource-efficient big data transfer \cite{silwa_tvt_2021} and intelligent decision making \cite{hu_cc_2022}. ML is also expected to enable improved energy efficiency and reduced human intervention \cite{mao_ieeecomst_2022}, secure 6G networks \cite{Siriwardhana_eucnc_2021}, intelligent 6G networks \cite{yang_ieeen_2020}. To this end, researchers have used different machine learning techniques such as Feed forward networks \cite{Jagannath18ICC},  Convolutional Neural Networks (CNN) \cite{chen_tmis_2021, AJagannath22PHYCOM, hijji_tits_2023}, Gated Recurrent Units (GRU) \cite{li_tvt_2024,
huang__tc_2024}, Echo State Networks (ESN) \cite{chang_tnnls_2022,
pricano_acmtit_2023}, Long-Short Term memory (LSTM) \cite{peng_icct_2021,
Tshakwanda_ojcs_2024} and even hybrid architectures \cite{AJagannath22GLOBECOM}. Additionally, there are techniques such as Generative Adversarial Networks \cite{Machumilane_ojcs_2023}, Autoencoder \cite{lu_vtc_2022}, and a category by itself referred to as Deep Reinforcement Learning \cite{arulkumaran2017deep, JagannathWiseML22_MRiNet, matsuo2022deep, SKafleComLet2023, luong2019applications} that uses a one of these machine techniques as its function approximator. These ML techniques fall under a general category called Neural Networks (NN). NNs also have several neural nodes that are connected and information is propagated between them. NNs have become increasingly popular due to its ability to learn complex patterns and relationships from data. This makes them suitable for a wide variety of tasks such as image and speech recognition, natural language processing, and even playing games. 

\begin{figure*}
\centering
\includegraphics[width=0.98\textwidth]{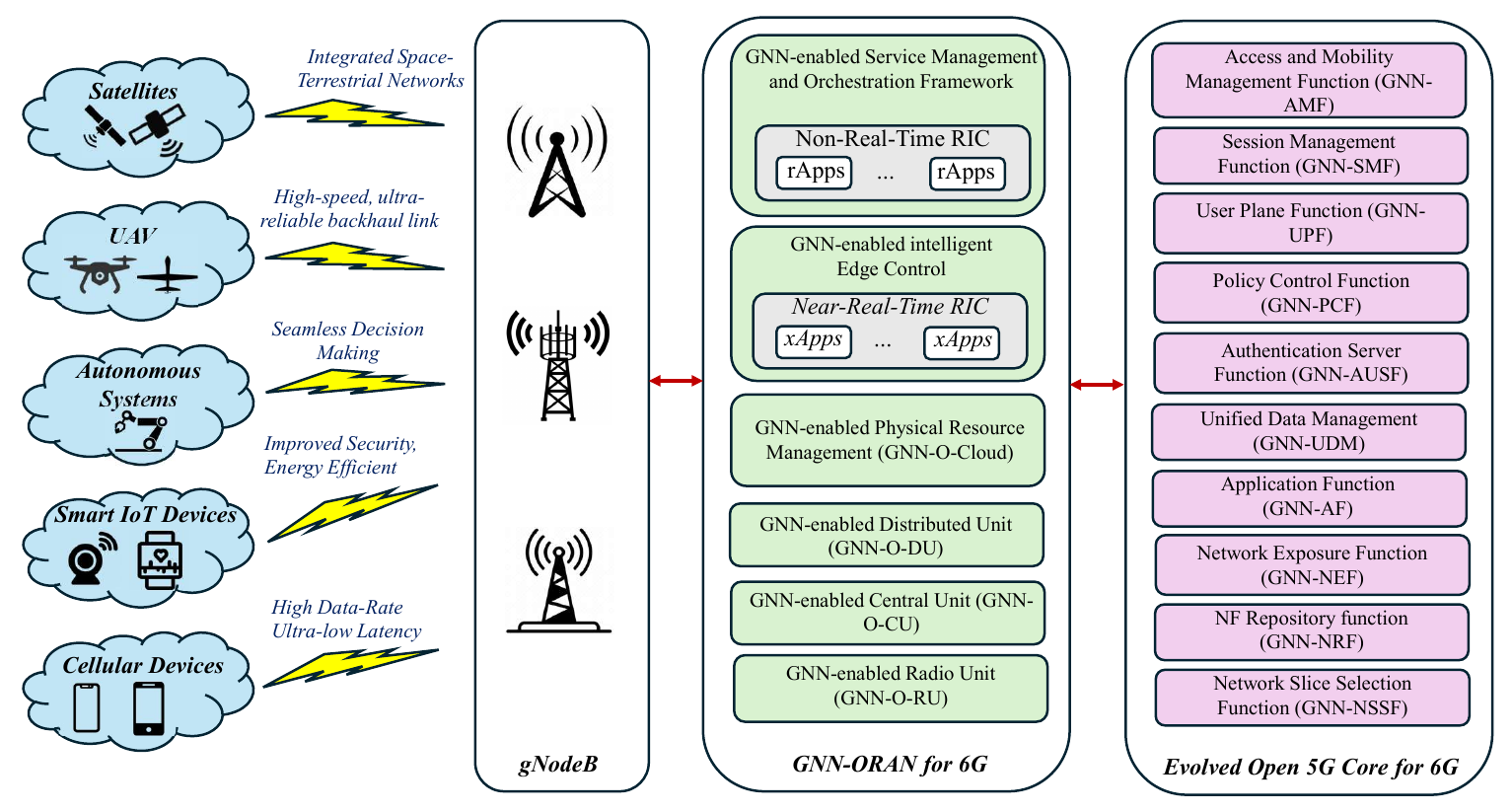}
\caption{Vision of GNN-enabled 6G Wireless Network.}
\label{fig:gnn_6g_vision}
\end{figure*}

\subsection{A Primer on Graph Neural Networks} \label{ssec:primer}
In addition to the aforementioned learning techniques, there is a relatively new class of neural networks called Graph Neural Networks (GNN) \cite{Gori_ijcnn_2005, Scarselli_TNN_2009}. GNNs are based on graphs that are modeled based on two entities, namely nodes (ie., objects) and edges (ie., relationships between nodes). 

\textbf{Uniqueness of GNNs.} GNNs are well-suited for capturing complex relationships and dependencies in graph data, which is challenging for traditional neural networks. They can learn representations of nodes and edges based on their local neighborhood information. Second,  GNNs can handle graphs of varying sizes and structures, making them applicable to a wide range of real-world problems. They can adapt to irregular and dynamic graphs while regular NNs fail. Third, pre-trained GNN models can be fine-tuned on specific graph-related tasks, enabling transfer learning. In this survey, we aim to provide a detailed study of GNN, discussing its design principles and architecture, state-of-the-art literature that covers how GNN has impacted various domains such as IoT, networking, spectrum awareness, and tactical systems, etc. GNNs are the go-to option for complex network control problems and applications with networks that are dynamic and need good generalizability and scalability with growing needs for better connectivity and high quality of service. Through this survey, we hope to make it easier for readers to learn and apply GNN to overcome some of the current research challenges of their respective domains of interest. 

\begin{table*}[t]

    \centering
    \caption{Comparison of recent surveys related to GNN.}
    \renewcommand{\arraystretch}{1.5}
    \begin{tabular}{ m{4cm} c c c c c c} 
         \hline \vspace{1mm}
         \multirow{2}{4cm}{\textbf{Survey of Surveys}} & \textbf{Detailed GNN} & \textbf{GNN for} & \textbf{GNN for Spectrum} & \textbf{GNN for} & \textbf{GNN for Tactical} & \textbf{Research } \\
         & \textbf{Architecture} & \textbf{IoT} & \textbf{Awareness} & \textbf{Networking} & \textbf{Systems} & \textbf{Directions} \\
        \hline
        \hline \vspace{2mm}
         \emph{Ruiz et al. 2021}\cite{ruiz_ieee_2021}& \cmark & \xmark &  \xmark & \cmark & \xmark & \xmark \\\vspace{2mm}
         \emph{Sato et al. 2020}\cite{sato_arxiv_2020}& \cmark & \xmark & \xmark & \xmark & \xmark & \xmark \\\vspace{2mm}
         \emph{Yuan et al. 2023}\cite{yuan_tpaml_2023}& \cmark & \xmark & \xmark & \xmark & \xmark & \xmark \\\vspace{2mm}
         \emph{Zhou et al. 2023}\cite{zhou_open_2020}& \cmark & \xmark & \xmark & \cmark & \xmark & \xmark \\\vspace{2mm}
         \emph{Abadal et al. 2022}\cite{Abadal_acmcs_2022}& \cmark & \xmark & \xmark & \xmark & \xmark & \xmark  \\\vspace{2mm}
          \emph{Wu et al. 2022}\cite{wu_acm_cs_2022}& \cmark & \xmark & \xmark & \xmark & \xmark & \xmark \\\vspace{2mm}
          \emph{Josephine et al. 2021}\cite{Josephine_arxiv_2023}& \cmark & \xmark & \xmark & \xmark & \xmark & \xmark \\\vspace{2mm}
          \emph{Tam et al. 2024}\cite{Tam_mdpielec_2024}& \xmark& \xmark & \cmark & \cmark& \xmark& \xmark \\\vspace{2mm}
         \emph{Dong et al. 2023}\cite{dong_acm_tsn_2023}& \cmark& \cmark & \xmark & \xmark& \xmark& \cmark \\\vspace{2mm}
          \emph{Liu et al. 2021}\cite{Liu_tnnls_2024}& \cmark & \xmark & \xmark & \xmark & \xmark &\cmark \\\vspace{2mm}
          \emph{Jiang et al. 2022}\cite{Jiang_eesa_2022}& \cmark & \xmark & \xmark & \xmark & \xmark & \cmark\\\vspace{2mm}
          \emph{Gao et al. 2023}\cite{Gao_2023_march}& \cmark & \xmark & \xmark & \xmark & \xmark & \cmark  \\\vspace{2mm}
          \emph{Bilot et al. 2023}\cite{Bilot_access_2023}& \cmark & \cmark & \xmark & \xmark &  \cmark &  \cmark \\\vspace{2mm}
          \emph{Waikhom et al. 2021}\cite{Lilapati2021arxiv}& \cmark & \xmark & \xmark & \xmark & \xmark & \cmark \\\vspace{2mm}
          \emph{Tam et al. 2022}\cite{tam_elect_2022}& \cmark & \cmark & \xmark & \cmark & \xmark &  \cmark \\\vspace{2mm}
          \emph{Gupta et al. 2021}\cite{atika_gnn_2021}& \xmark & \xmark & \xmark  & \xmark & \xmark & \cmark \\\vspace{2mm}
          \emph{Rahmani et al. 2023}\cite{Rahmani_tits_2023}& \cmark & \xmark & \xmark  & \xmark & \xmark & \cmark \\\vspace{2mm}
          \emph{Suárez-Varel et al. 2023}\cite{Varela_ieeenet_2023}& \cmark & \xmark &\xmark  & \cmark & \xmark & \cmark \\\vspace{2mm}
          \emph{Munikoti et al. 2023}\cite{Munikoti_tnnls_2023}& \cmark &\xmark & \xmark & \xmark& \xmark& \cmark \\\vspace{2mm}
          \emph{Wu et al. 2021}\cite{Wu_tnnls_2021}& \cmark  & \xmark & \xmark & \xmark & \xmark & \cmark\\\vspace{2mm}
          \emph{Zhou et al. 2022}\cite{Zhou_tist_2022}& \cmark & \xmark & \xmark & \xmark & \xmark & \cmark  \\\vspace{2mm}
          \emph{Longa et al. 2021}\cite{Antonio_arxiv_2021}& \cmark & \xmark & \xmark & \xmark &  \xmark & \cmark  \\\vspace{2mm}
          \emph{He et al. 2021}\cite{Shiwen2021}& \cmark & \xmark & \xmark & \cmark & \xmark & \cmark  \\\vspace{2mm}
          \textbf{This Survey} & \cmark & \cmark & \cmark & \cmark & \cmark & \cmark \\\\
          
        \hline
    \end{tabular}
    \label{tab:gnn_survey}
\end{table*}

\textbf{Advantages and Disadvantages of GNNs.} GNNs offer superior flexibility and modeling capabilities for data that is structured as graphs, which RNNs or CNNs are not equipped to handle. GNNs can naturally encode node features, edge attributes, and global graph-level properties through iterative message passing and neighborhood aggregation, allowing the model to learn representations that are aware of both local and global topology. More details about message passing and neighborhood aggregation will be discussed later (Section \ref{ssec:gnn_arch}). In contrast, CNNs operate on fixed grid structures (e.g., images), and RNNs on ordered sequences, making them unsuitable for scenarios where connectivity is irregular, such as wireless mesh networks, sensor networks, or social graphs. Additionally, GNNs are permutation-invariant, that is, the output does not depend on the order of nodes or edges. Furthermore, GNNs can integrate edge features such as link capacity, SNR, latency directly into the computation via attention or gating mechanisms, which is not possible with CNNs or RNNs. From a computational perspective, GNNs are comparatively computationally efficient even for large graphs, unlike CNNs which waste resources on irrelevant zero-padding in sparse spatial inputs. Additionally, advanced GNN variants like Graph Attention Networks (GATs), and Spatio-Temporal GNNs (ST-GCN) enable fine-grained control over information flow across the graph and time, outperforming RNNs in spatio-temporal forecasting tasks (e.g., traffic flow, node failures). More details about different types of GNN will be discussed later in this paper (Section \ref{ssec:gnn_types}). Finally, GNNs exhibit strong inductive learning capability, meaning a model trained on one graph can generalize to unseen graphs with similar structural patterns which is something CNNs and RNNs struggle with due to their tight coupling with input shape and size.

While GNN demonstrates capability to be applied for a variety of problems, they cannot be arbitrarily used as a solution for tasks that lack an inherent or meaningful graph structure, as forcing data into a graph format without natural relation or dependency can lead to poor performance, overfitting, and unnecessary computational overhead. Let us consider an example of RF spectrum monitoring scenario, where the objective to map the spectrum activity within a geographical area where a number of sensors are deployed. In this scenario, the spectrum activity observed at each sensor can be aggregated to form the global spectrum activity map. In this scenario, the sensors can be considered as nodes, the links between the nodes forms the edges, the sensed spectrum measurements are the node features and the the distance between the nodes are the edge features. However, if we modify this scenario to where there are a number of transmitters deployed within a geographical area and the objective is RF fingerprinting i.e, identifying the transmitter based on the received signal, modeling this problem as a graph with the transmitters as nodes is not ideal since signal from one transmitter (i.e, node) does not provide useful information about another sensor's fingerprint. The fingerprinting depends only on individual transmitter characteristics, not on any neighboring transmitters. This is a simple example to show how a minor change in the network and the network problem could deem GNN an infeasible option and instead RNNs can be used \cite{Ajagannath2022ComST2022,
AJagannathTCCN2023}. 

\textbf{General Applications of GNNs.}
GNNs have been widely used to address a variety of network management issues such as resource allocation \cite{Perera_wcnc_2023, Chen_iotj_2022, Eisen_spawc_2019, Wang_icassp_2021, Rahman_icece_2022}, spectrum sensing and sharing \cite{Zhang_twc_2022, Srinath_access_2022,hu_ieeeton_july_2025}, privacy \cite{Lee_twc_2023}, edge computing \cite{Pamuklu_ieeenl_2023, ying_phycom_2023}, trajectory optimization \cite{Chen_comlet_2022, li_iotj_2022}. Flex-Net \cite{Perera_wcnc_2023} uses GNN to jointly optimize the direction of communication and transmission power with low computational complexity as well as enable scalability and generalization capabilities. A GNN variant called Inductive Graph Neural Network Kriging (IGNNK) is used for secure cooperative sensing \cite{Zhang_twc_2022}. The advantages of IGNNK are that a trained GNN model can be transferred to a new graph structure as well as be used to generate data for virtual sensors for a given location as discussed in \cite{Wu2020InductiveGN}. There is another inductive learning method called GraphSAINT \cite{zhang_iclr_2020} which samples the training graphs into smaller mini-batches. This approach reduces the “neighbor explosion” problem, maintains training efficiency even for deeper networks, and seamlessly supports architectural enhancements like attention mechanisms and skip connections. GraphSAINT has been used for applications such as Botnet Detection \cite{yin_mdpi_mat_2024}, Hardware Trojan Detection \cite{lashen_iscas_ts_2023}, among others. There is a relatively new variant of GNN called Factor Graph Neural Network (FGNN) \cite{FGNN_neurips2020,FGNN_JMLR_2023} that has been used for cooperative combinatorial optimization problems \cite{hu_kbs_2020}, localization \cite{dai_sensors_2022}, route planning \cite{li_aamas_2024}.

Decentralized management and control is one of the key features of next-generation wireless communication. However, privacy leakage is a major issue. To address this issue, in \cite{Lee_twc_2023} the authors use GNN to design a privacy-preserving and privacy-guaranteed training algorithm for information exchanges among neighbors in wireless networks. In \cite{Pamuklu_ieeenl_2023}, the authors use a combination of GNN and Reinforcement learning (RL) to optimize the task offloading from IoT devices to mobile nodes that is robust to changes in network topology and mobile node failure. In \cite{li_iotj_2022}, the authors present a GNN-enabled Advanced Actor-Critic (GNN-A2C) algorithm for aerial-edge IoT (EdgeIoT) system to jointly optimize trajectory and task offloading. One of the important research directions for energy-efficient network management is trajectory optimization, which refers to the set of steps taken by any autonomous system to optimize a given network control problem. These network control problems can be maximizing throughput, minimizing end-to-end delay, IoT data fusion, and swarm UAV deployment.
With the continuous rise in the use of GNNs for a wide range of applications, it is important to keep track of existing literature works as well as have a clear idea of the challenges and opportunities provided by GNNs. Figure~\ref{fig:gnn_6g_vision} shows our vision of GNN-enabled 6G networks with integrated ORAN architecture.

\begin{table}[htbp]
    \centering
    \caption{List of notations and its definition.}
    \renewcommand{\arraystretch}{1.55}
    \begin{tabular}{ m{2cm} c c } 
         \hline \vspace{1mm}
         \textbf{Notation} & \textbf{Definition} \\
        \hline
        \hline \vspace{1mm}
         \textit{N} & Number of Nodes in a graph \\
         \textit{F} & Number of Features \\
         \textbf{F} & Feature Matrix\\
         \textit{G} & Graph \\
         \textit{V} & Vertices\\
         \textit{E} & Edges\\
         $i$ & Index of node of interest \\
         $j$ & Index of neighboring node\\
         $\textbf{x}_{i}$ & Node Feature vector\\
         $l$ & Layer index\\
         $L$ & Total number of layers\\
         $m_{i}^{(l+1)}$ & Aggregated message of node $i$ at layer $l+1$\\
         $\textbf{h}_{i}^{(l)}, \textbf{h}_{j}^{(l)}$ & Feature vectors of node $i$, node $j$ at layer $l$\\
         $\textbf{h}_{i}^{(l+1)}, \textbf{h}_{j}^{(l+1)}$ & Feature vectors of node $i$, node $j$ at layer $l+1$\\
         $\textbf{W}_{i}^{(l)}$ & Trainable weight matrix of node $i$ at layers $l$\\
         $\textbf{W}_{i}^{(l+1)}$ & Trainable weight matrix of node $i$ at layers $l+1$\\
         $\textbf{W}_{j}^{(l)}$ & Trainable weight matrix of node $j$ at layers $l$\\
         $\textbf{W}_{j}^{(l+1)}$ & Trainable weight matrix of node $j$ at layers $l+1$\\
         $\textit{N}_{i}$ & Set of Neighbors of node $i$\\
         $A(\cdot)$ & Aggregator Function\\
         $U(\cdot)$ & Update Function\\
         $\sigma(\cdot)$ & Activation Function\\
         $R(\cdot)$ & Readout Function\\
         $(\cdot)^T$ & Transpose Operation\\
         $\textbf{A}$ & Adjacency Matrix\\
         $\textbf{D}$ & Degree Matrix\\
        $e_{ij}$ & Attention coefficient of node $i$ for neighbor $j$\\
        $a(\cdot)$ & Attention Mechanism Function \\
        $\alpha_{ij}$ & Normalized attention coefficient of \\ &node $i$ for all neighbors $j\in N_{i}$\\
        $\textbf{a}$ & Attention Mechanism Vector\\
        $||$ & Concatenation Operator\\
        $\rm{inp}$ & Input Vector\\
        $\rm{enc}$ and $\rm{dec}$ & Encoder and Decoder Output Vector\\
        $f_{\rm{enc}}$ and $f_{\rm{dec}}$ & Encoder and Decoder Functions\\
        $\theta_{\rm{enc}}$ and $\theta_{\rm{dec}}$ & Parameters associated with  $f_{\rm{enc}}$ and $f_{\rm{dec}}$\\
        $\mathcal{N}$ & Number of elements in $\rm{\textbf{inp}}$\\
        $S_{\rm{inp}}$ and $S_{\rm{dec}}$ & Sum of elements in $\rm{inp}$ and $\rm{dec}$\\
        $||(\cdot)||^2$ &Squared Euclidean Distance\\
        $\epsilon$ & Learnable or fixed scalar parameter\\
        $f_i(\cdot)$ & Graph Capsule Function\\
        $\mathcal{F}_{i}$ & Neighborhood feature value set\\
        $\lambda_{ij}$ & Edge weight between nodes $i$ and $j$\\
        $x_{i}$ & Node feature value\\
        $p$ & Number of instantiation parameters\\
        \hline
    \end{tabular}

    \label{tab:notations}
\end{table}

\begin{table}[htbp]
    \centering
    \caption{List of commonly used acronyms and their definitions.}
    \renewcommand{\arraystretch}{1.5}
    \begin{tabular}{ m{2cm} c c } 
         \hline \vspace{1mm}
         \textbf{Acronym} & \textbf{Definition} \\
        \hline
        \hline \vspace{1mm}
         6G & Sixth-Generation \\
         NextG & Next-Generation\\
         IoT & Internet of Things\\
         5G & Fifth Generation\\
         GHz & Gigahertz\\
         ML & Machine Learning\\
         GNN & Graph Neural Networks\\
         QoS & Quality of Service\\
         GRU & Gated Recurrent Unit\\
         ESN & Echo State Networks\\
         LSTM & Long-Short Term Memory\\
         CNN & Convolutional Neural Networks\\
         RL & Reinforcement Learning\\
         GCN &  Graph Convolutional Networks\\
         GAT &  Graph Attention Networks \\
         GraphSAGE & Graph Sample and Aggregate\\
         GAE & Graph Autoencoders\\
         GIN & Graph Isomorphic Networks\\
         RNN & Recurrent Neural Network\\
         GCapsNet & Graph Capsule Networks\\
         STGN & Spatial-Temporal Graph Networks\\
         MANET & Mobile Ad Hoc Network\\
         GCNN & Graph Capsule Convolution Neural Network \\
         NID & Network Intrusion Detection\\
         RSSI &Received Signal Strength Indicator\\
         GTCN & Graph-based Temporal Convolutional Network\\
         DRL & Deep Reinforcement Learning\\
         MEC & Mobile Edge Computing\\
         DT & Digital Twin\\
         DDQN & Deep Distributed Q-Network\\
         UAV & Unmanned Aerial Vehicle\\
         UAN & Unmanned Aerial Network\\
         SAR & Synthetic Aperture Radar\\
         FL & Federated Learning\\
        \hline
    \end{tabular}
    \label{tab:acronyms}
\end{table}

\subsection{Scope of the article} 

The scope of the article is to survey the applications of GNN for key areas of research in IoT and NextG wireless networks. The applications of GNN for wireless domain is still in its infancy. In this survey we explored and compiled applications of GNN for wireless domain that is worth reviewing to enable maturity of GNN as well as future wireless networks. To this end, we selected IoT, RF Spectrum awareness, networking and tactical systems given these are our area of research focus.Collectively, these form the most important and key problems that needs to be addressed to deliver intelligent, secure, and adaptive connectivity solutions for IoT and nextG networks.

Table \ref{tab:gnn_survey} summarizes all the recent surveys related to GNN. Compared to the above works, in this survey, we focus on the following aspects:
\begin{itemize}
    \item Provide a detailed background about GNN and its architecture by discussing its components and functions. We also discuss the different types of GNNs and discuss related literature that uses different GNN types to solve a wide variety of problems. This is essential to provide the relatively newer reader with a great guide to start their journey. 
    \item In today's world, all \textbf{\textit{IoT smart applications}} rely on \textbf{\textit{IoT data fusion}} to aggregate and process vast amounts of heterogeneous data, facilitating real-time decision-making in domains such as healthcare, smart cities, and industrial automation. As IoT ecosystems expand, ensuring network security becomes crucial to protect the vast amount of data collected from any external attacks, making \textbf{\textit{IoT network intrusion detection}} essential for identifying cyber threats and safeguarding sensitive data. For this reason in this article we first give a brief introduction to the Internet of Things and discuss the application of GNN for IoT smart applications, IoT data fusion and Intrusion Detection.
    \item While small-scale indoor IoT networks may rely on wired network setup, there are many IoT applications that share data to a cloud or other data processing centers wirelessly. This makes it important to efficiently utilize the radio spectrum is critical for sustaining the high-speed, low-latency communication demanded by nextG networks. \textbf{\textit{Radio frequency spectrum sensing}} and \textit{\textbf{radio frequency signal classification}} optimize spectrum allocation, minimize interference, and enhance dynamic access to limited spectrum resources, thereby ensuring seamless wireless connectivity. To this end, we provide an overview of spectrum awareness and elucidate the application of GNN for Radio Frequency Spectrum Sensing and Radio Frequency Signal Classification.
    \item In addition to spectrum allocation and dynamic access, it is also important to manage the overall network to keep up with growing number of IoT devices. \textbf{\textit{Network and signal characteristics prediction}} enables proactive adjustments to network parameters, improving \textbf{\textit{routing optimization}} and \textbf{\textit{congestion control}} for efficient data transmission. \textbf{\textit{Mobile edge computing}} enhances real-time processing by decentralizing computation, reducing the load on cloud infrastructure and enabling rapid decision-making for IoT applications. \textbf{\textit{Digital twin networks}} create virtual replicas of physical systems, which helps in mimicking the real-world application performance and aids in developing and testing virtually before integrating them to the real-world scenario. Additionally, \textbf{\textit{unmanned aerial networks}} extend connectivity to remote and underserved areas, supporting disaster response and autonomous systems. With this goal in mind, we delve into the basics of networking and discuss the applications of GNN for networking focusing on Network and Signal Characteristics Prediction, Routing Optimization, Congestion Control, Mobile Edge Computing, Digital Twin Networks, and Unmanned Aerial Networks.
    \item In mission-critical environments, \textbf{\textit{tactical communication and sensing systems}} ensure secure and resilient operations, with \textbf{\textit{target recognition}} and \textbf{\textit{localization}} playing key roles in defense, surveillance, and autonomous navigation. To further shed some light on this area of research we explore the basics of tactical systems and provide a qualitative discussion of the application of GNN for Tactical Communication and Sensing Systems and Localization.
    \item For each section, we provide a table that shows the type of ML (i.e,, GNN models used), the application as well as the dataset or simulator used for performance evaluation. Through these tables we hope to provide a quick start-guide to researchers who are interested in using GNN in their research area of interest.
    \item Finally we deep-dive into the research opportunities and future research direction of GNN to motivate more exploration in the domains specific to next-generation wireless networks.
\end{itemize}

\subsection{Survey Organization.} \label{ssec:sur_org}
\begin{figure}[t]
\centering
\includegraphics[width=0.48\textwidth]{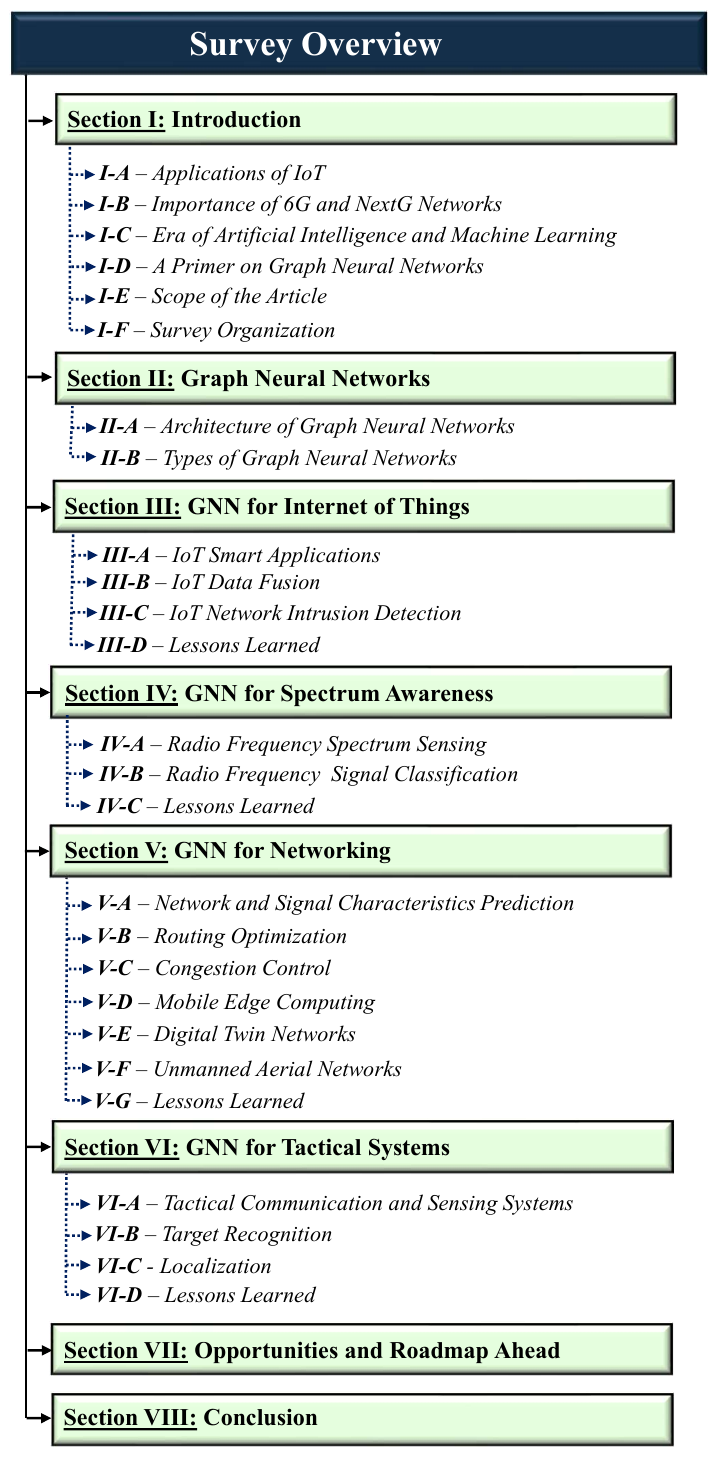}
\caption{Overall Organization of the Survey.}
\label{fig:survey_organization}
\end{figure}

The rest of the survey is organized as follows. We formally introduce GNN by discussing its inception and providing a detailed description of the architecture and types of GNN in Section~\ref{sec:gnn}. In Section~\ref{sec:gnn_iot}, we discuss the applications of GNN for the Internet of Things, followed by applications of GNN for spectrum awareness, networking, and tactical systems in Section~\ref{sec:gnn_spec}, Section~\ref{sec:gnn_net} and Section~\ref{sec:gnn_tact}, respectively. Finally, we provide our perspective on the opportunities that could motivate novel research directions utilizing GNN in Section~\ref{ssec:opp_roadmap} before concluding in Section~\ref{sec:conclusion}. List of notations and its definition, and commonly used acronyms are tabulated in Table.\ref{tab:notations} and Table \ref{tab:acronyms}, respectively. The overall survey overview is shown in Figure~\ref{fig:survey_organization}.

\begin{figure}[t]
\centering
\includegraphics[width=0.48\textwidth]{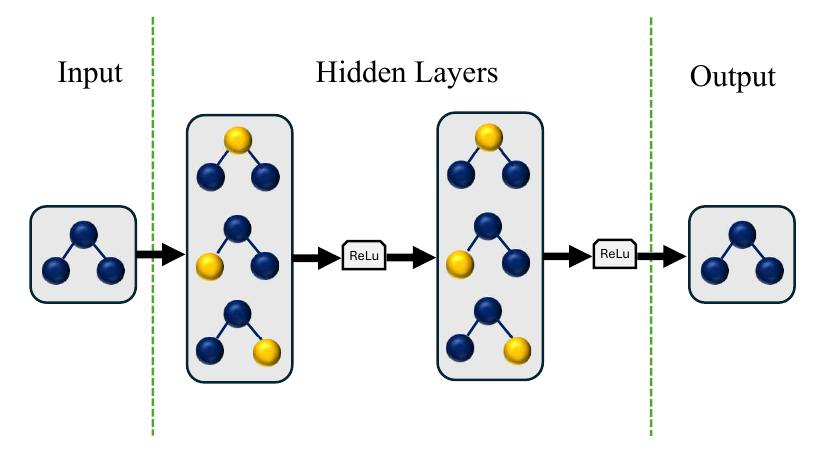}
\caption{Architecture of Simple GNN.}
\label{fig:gnn_arch_v1}
\end{figure}

\begin{figure*}[t]
\centering
\includegraphics[width=0.9\textwidth]{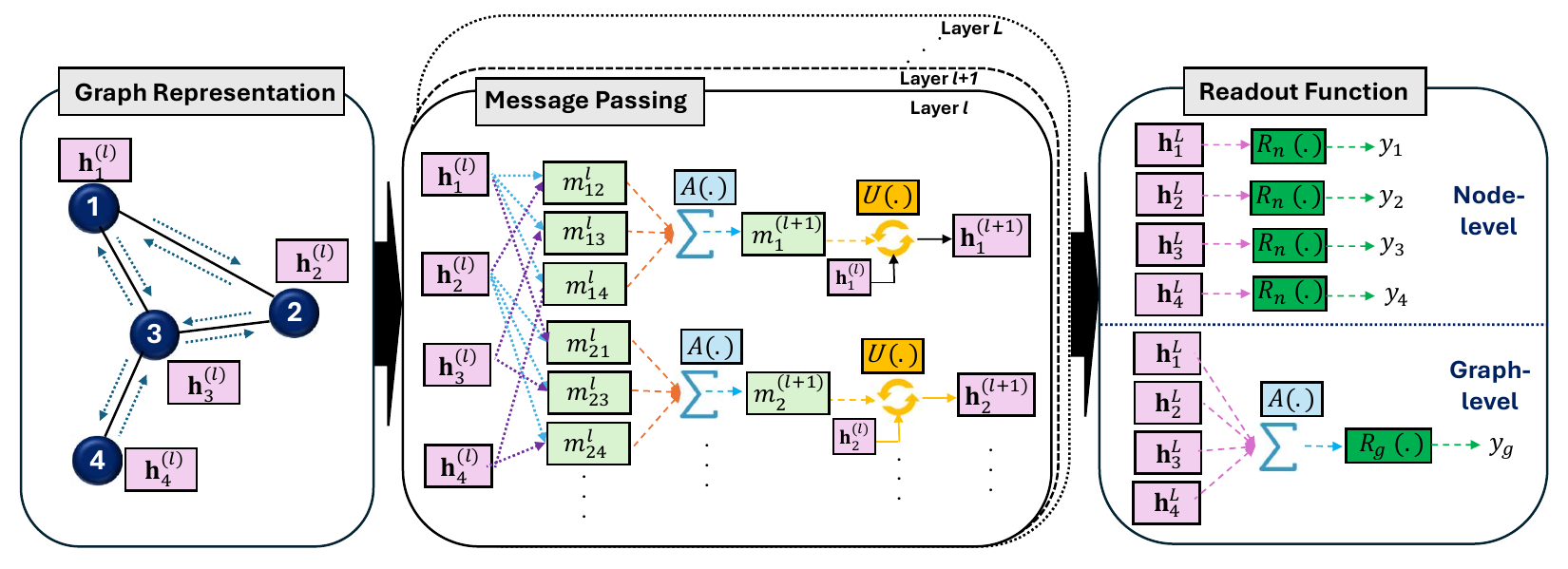}
\caption{Architecture of GNN based on Message Passing Neural Networks.}
\label{fig:gnn_arch}
\end{figure*}

\begin{figure*}[t]
\centering
\includegraphics[width=0.9\textwidth]{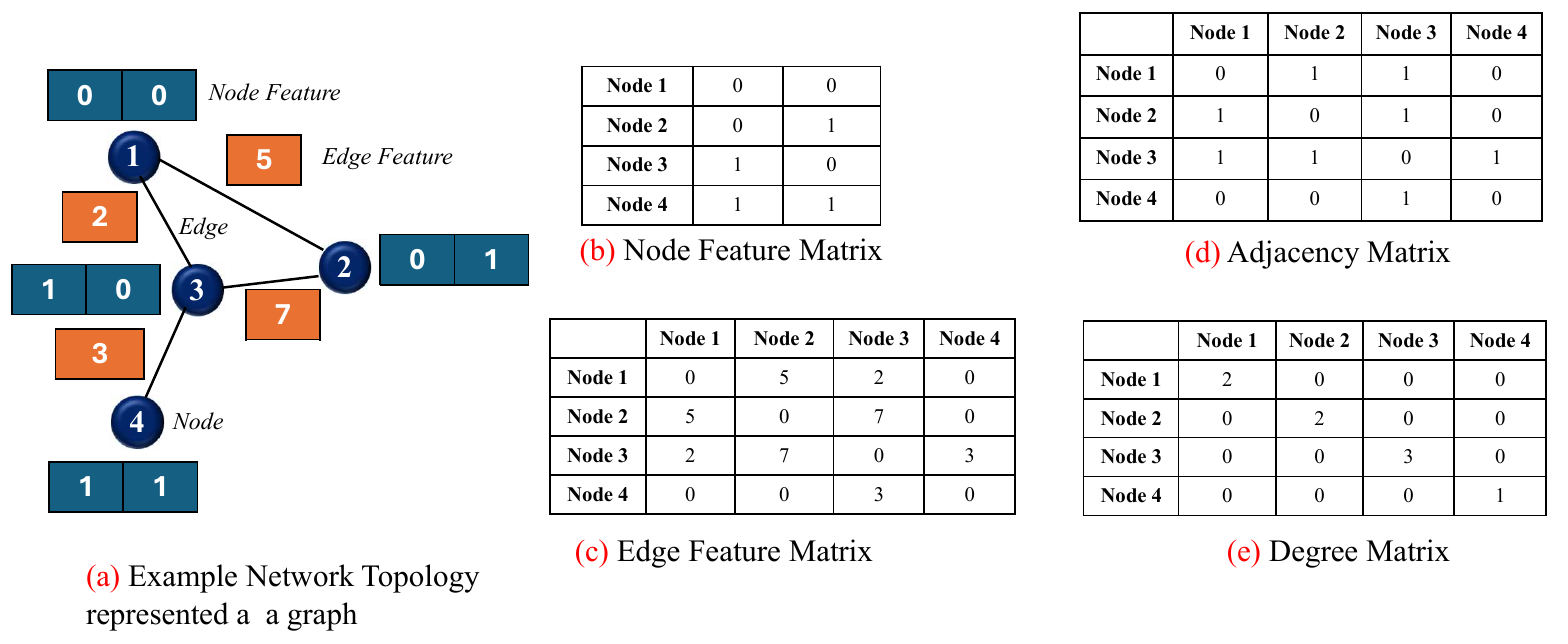}
\caption{Examples of GNN graph representation and different matrices.}
\label{fig:terminology}
\end{figure*}

\section{Graph Neural Networks} \label{sec:gnn}

In this section, we will provide a detailed introduction to Graph Neural Networks (GNN). To this end, we first present the architecture of GNN in Section~\ref{ssec:gnn_arch} and then discuss different types of GNN variants in Section~\ref{ssec:gnn_types}.

\begin{figure*}[t]
\centering
\includegraphics[width=1\textwidth]{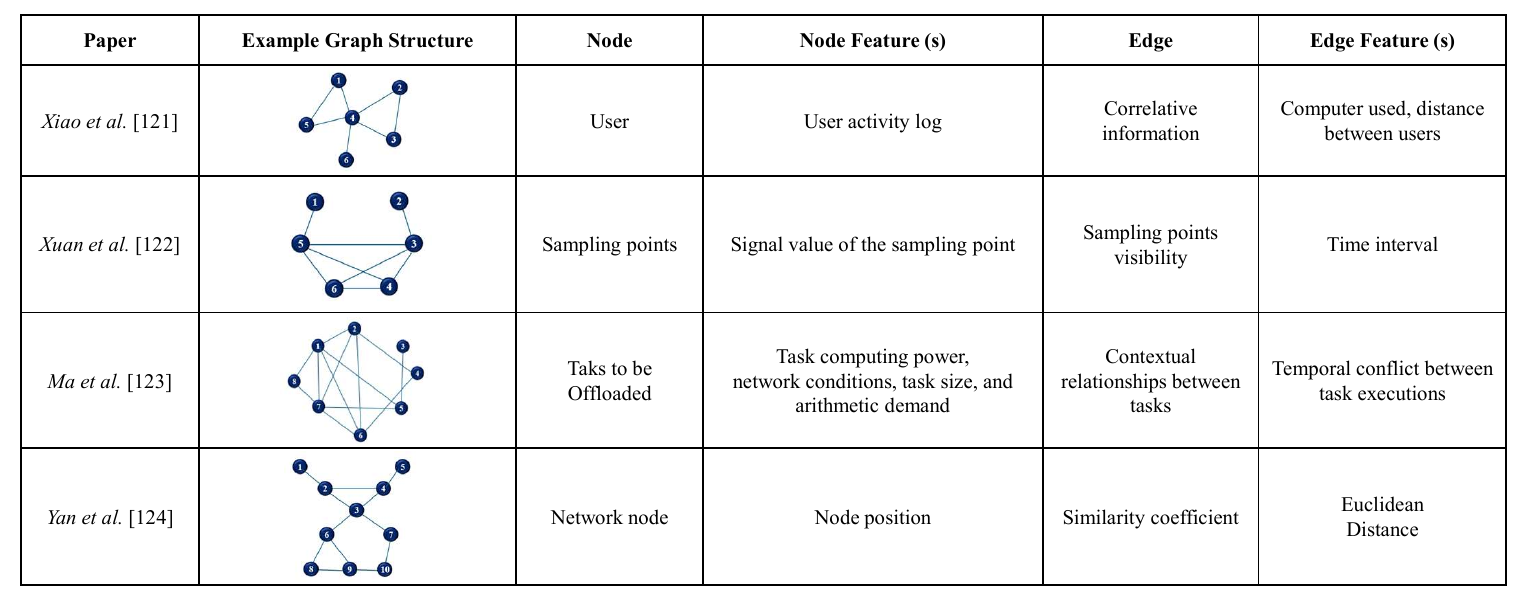}
\caption{Structure of Graphs with Nodes, Node Features, Edges, and Edge Features. In the example graph structure, the solid blue circle with numbers denote the node and the blue lines in-between denotes the edges.}
\label{fig:str_feat_details}
\end{figure*}

\begin{figure*}[t]
\centering
\includegraphics[width=0.7\textwidth]{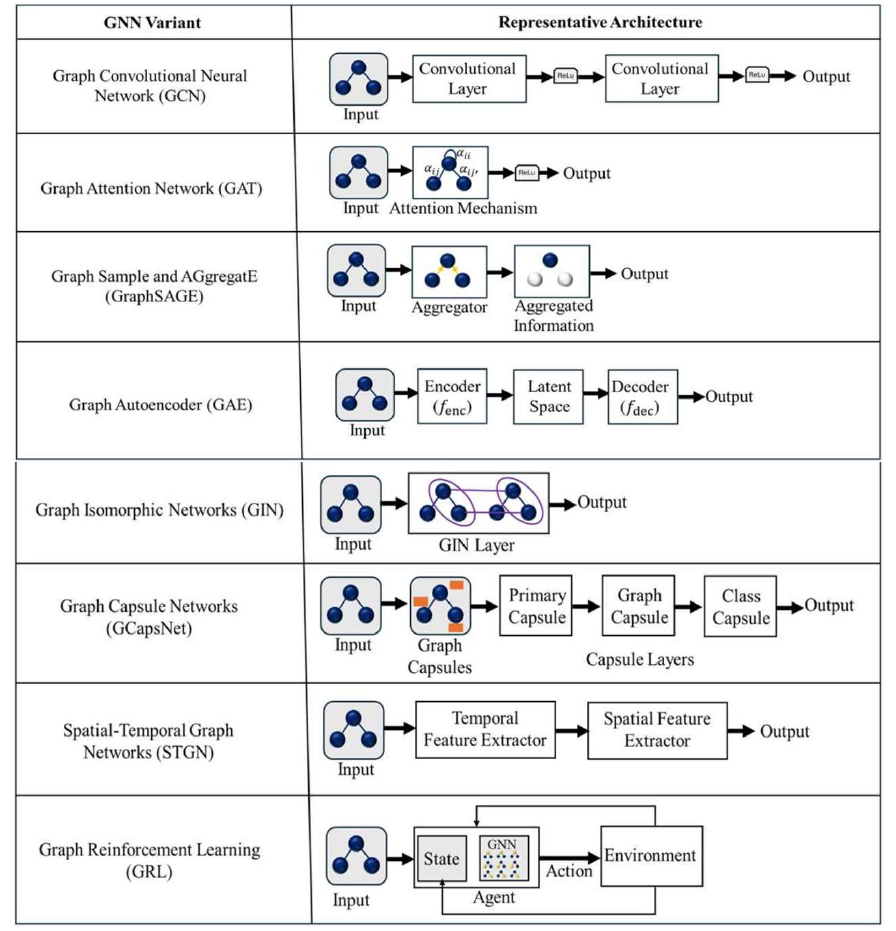}
\caption{Representative Architecture of GNN variants.}
\label{fig:gnn_arch_vari}
\end{figure*}

\subsection{Architecture of Graph Neural Networks}\label{ssec:gnn_arch}
Graph Neural Networks (GNNs) are a class of neural networks designed to work with graph-structured data. Graphs consist of nodes and edges, where nodes represent entities (e.g., users in a social network or atoms in a molecule), and edges represent relationships or connections between these entities. GNNs are specifically designed to process and extract meaningful information from these graph structures, making them well-suited for a wide range of applications. 

A simple GNN consists of the following three main layers namely the input layer, hidden layers, and output layer as shown in Fig.~\ref{fig:gnn_arch_v1}. To enable information exchange and aggregation from neighboring nodes in the graph, a more sophisticated GNN is designed using the concept of message passing. With the help of message passing through the graph, GNNs can leverage the local neighborhood information of all the nodes in the graphs and help in refining and updating the global representation of the network. This allows for GNN to capture both the structural information and the features of all the nodes of the graph. The architecture of GNN designed based on message passing is shown in Fig.~\ref{fig:gnn_arch}. The working of message passing-based GNN is as follows.

\begin{itemize}
    \item \textbf{Graph Representation.} A graph is represented as  $G=(V,E)$, where $V$ is the set of nodes (vertices) and $E$ is the set of edges (connections) between nodes. Each node $i \in V$ typically has an associated feature vector $\textbf{x}_i$ that encodes information about the node. A GNN aims to learn representations of nodes that capture the graph's structural and semantic information. The general principle behind generating a graph dataset (i.e., representation). First, identify the problem of interest, this can be energy optimization, reducing latency, improve throughput, among others. The next step is to understand the network topology and the different entities present. The different entities can vary depending on the application. For example, it can be a transmitter, receiver, relay, access point, base station, moving vehicles, etc., Each of these entities is called a node of the graph (refer Fig.~\ref{fig:terminology}(a)). The next step is to understand the connections between entities. These connections are called edges. The edges denote the relationship between the entities (or nodes). For example, if one transmitter may be connected to a receiver. Now there is an edge between the transmitter and receiver. This receiver may in turn have connections to two relay nodes, and so on. Now we have the details of the nodes and edges. The next step is to identify the features of the node and the edges. First, let's begin with the node features. The node features are nothing but the attributes of the node. These features can include details such as node name, ID, location information, speed of travel, transmission power, maximum transmission power, and so on. Next is the edge feature which are attributes of the connection between two nodes. These attributes include link distance, latency, throughput, link capacity, and so on. Finally, all this information is represented in the form of matrices that can be used as the graph representation for the GNN. Figure \ref{fig:str_feat_details} shows a table of different GNN structure used in representative works discussed later in Sections~\ref{sec:gnn_iot}-\ref{sec:gnn_tact} \cite{haitao_capitd_2023,Xuan_tnse_2022,ma_icc_mogr_2024,yan_2021_icassp}. 
    
    There are four types of matrices namely feature matrix, adjacency matrix, degree matrix, and weight matrix.
    The \textit{\textbf{feature matrix}} represents the 2-D representation of features of all the nodes in the graph. For example, if a graph has $N$ nodes, and each node has $F$ number of features, then the feature matrix $\textbf{F}$ has a dimension of $N \times F$. There are different types of features such as numeric features, categorical features, and binary features, among others. An example of binary features can be a multi-hop network, where each node will have a value of 0 or 1 based on its feature such as it can either be a transmitter (0) or receiver (1), it may be static (1) or mobile (0) and so on. In this case, as an example, the number of features will be 2 and the feature vector will comprise of either 1 or 0 and will be of length 2. If there are 4 nodes in the network, then the feature matrix $\textbf{F}$ will be of size $4\times2$ (refer Fig.~\ref{fig:terminology}(b)). The relationship between the nodes are represented in the form of \textbf{\textit{edge feature matrix}} (refer Fig.~\ref{fig:terminology}(c)). Examples of edge features include the distance between the two nodes, path loss, etc. The \textbf{\textit{adjacency matrix}} represents the connection between the nodes in a graph and takes a value of 1 or 0 if there is an edge between a pair of nodes or not. It is worth mentioning here that the adjacency matrix is always a square matrix, meaning the number of rows and columns is the same. If there are $N$ nodes in the network, then the shape of the adjacency matrix will always be $N \times N$ (refer Fig.~\ref{fig:terminology}(d)). The adjacency matrix can also be a weighted adjacency matrix which is used to represent graphs in which the edges have weights. These weights can represent various attributes such as the strength of a connection, distance, cost, among others. The \textbf{\textit{degree matrix}} is a diagonal matrix containing information about the degree of each node. Here, the degree of the node indicates the number of terminating edges for a node. For example, a node may have $C$ connections associated with it (refer Fig.~\ref{fig:terminology}(e)). Similar to the adjacency matrix, the degree matrix is also a square matrix of shape $N \times N$. Finally, we have the \textbf{\textit{weight matrix}} which is the trainable parameters used to transform node features during the learning process.
    
    \item \textbf{Message Passing.} The core operation in GNNs is message passing, which enables nodes to exchange information with their neighbors in the graph and can be bi-directional in nature \cite{Yujiao_bgnn_2021, zhen_fgnn_2023}.  The message passing step is typically defined in two parts:
    \begin{itemize}
        \item  \textbf{Aggregation Function:} Aggregation refers to the process of collecting information at the node of interest from all the nodes in the neighborhood where neighborhood refers to the set of nodes that are connected to the node $i$. Each node aggregates information from its neighbors by computing an aggregation function (e.g., sum, mean, max) over the features of its neighbors. Mathematically, this can be represented as:
        \begin{align}\label{eq:agg}
            \textbf{m}_{N_i}^{(l)} &= A^{(l)}(\textbf{h}_{j}^{(l)}, \forall j \in N_i),\\
            \textbf{m}_{N_i}^{(l)} &= \sum_{j \in N_i} \textbf{h}_{j}^{(l)}
        \end{align}
        where, $\textbf{m}_{N_i}^{(l)}$ represents the aggregated message at node $i$ at layer $l$, $\textbf{h}_{j}^{(l)}$ represents the hidden state of neighbor node $j$ at layer $l$, $A^{(l)}(\cdot)$ represents the aggregation function, $N_i$ represents the set of neighbors of node $i$ at layer $l$.
        \item \textbf{Update Function:} Once messages are aggregated, each node updates its own representation based on the aggregated messages. This update function (U) can be implemented using neural networks (typically multi-layer perceptrons), and it can be represented as:
        \begin{align}\label{eq:updt}
            \textbf{h}_{i}^{(l+1)} &= U^{(l)}(\textbf{h}_{i}^{l}, \textbf{m}_{N_i}^{(l)}),\\
            \textbf{h}_{i}^{(l+1)} &= \sigma(\textbf{W}_{i}^{(l+1)}\textbf{h}_{i}^{(l)}+\textbf{W}_{j}^{(l+1)}\textbf{m}_{N_i}^{(l)} + \mathbf{b}^{(l+1)})
        \end{align}
        where, $\textbf{h}_{i}^{(l)}$ and $\textbf{h}_{i}^{(l+1)}$ are feature vectors of node $i$ at layers $l$ and $l+1$, respectively, $\textbf{h}_{j}^{(l)}$ is the feature vector of node $j$ at layer $l$ , $\textbf{W}_{i}^{(l+1)}$ and $\textbf{W}_{j}^{(l+1)}$ are the trainable weight matrices for nodes $i$ and $j$ at layers $l+1$, $\textbf{m}_{N_i}^{(l)}$ represents the aggregated message at node $i$ at layer $l$, and $\mathbf{b}^{(l+1)}$ is the bias term which is usually a constant added to the product of features and weights and $\sigma$ is the element-wise non-linearity activation function. Activation Functions allow the defining of a nonlinear family of functions to capture the relationship between the input graph data and its representation. Some examples of non-linear activation functions include Sigmoid or Logistic, Hyperbolic Tangent, Rectified Linear Unit (ReLU), LeakyReLU, Parametric ReLU, Exponential Linear Units, Softmax, Scaled Exponential Linear Unit (SELU) and Gaussian Error Linear Unit (GELU), among others. GNNs typically have multiple layers (or iterations) of message passing and updating. Each layer refines the node representations by considering information from a broader neighborhood in the graph.
    \end{itemize}
    
    \item \textbf{Readout Function:} After several message-passing layers, the final node representations are often aggregated to produce a graph-level representation or prediction. The readout phase computes a feature vector for each node $i$ (node-level) and is expressed as
    \begin{align}\label{eq:readout_node}
            y_{i} = R_n(\textbf{h}_{i}^{L^{T}}|i\in G)
    \end{align}
    Similarly, the readout phase can also compute the feature vector for the entire graph $G$ (graph-level) and is expressed as
    \begin{align}\label{eq:readout_graph}
            y_{g} = R_g(A({h}_{i}^{L^{T}}|i\in G)),
    \end{align}
    where $R_n$ and $R_g$ are the readout function per-node and graph-level, respectively, $L$ is the total number of layers and $(\cdot)^T$ is the transpose operation.
\end{itemize}

\subsection{Types of Graph Neural Networks}\label{ssec:gnn_types}

In this section, we will discuss some of the most popular types of variations of GNN architectures namely Graph Convolutional Networks (GCNs),  Graph Attention Networks (GAT), Graph Sample and Aggregated (GraphSAGE), Graph Autoencoders (GAE), Graph Isomorphic Networks (GIN), Graph Capsule Networks (GCaps), Spatial-Temporal Graph Networks (STGNs) and Graph Reinforcement Learning (GRL). Figure~\ref{fig:gnn_arch_vari} shows the representative architecture of each of the GNN variants to help understand the basic operations.

\textbf{Graph Convolutional Network.} 
Graph Convolutional Networks (GCNs) are a type of deep learning model used for processing and analyzing data represented as graphs or networks. They were introduced to extend the power of CNNs to irregular data structures like graphs. CNNs are commonly used for tasks like image classification, object detection, image segmentation, and various computer vision tasks where spatial relationships are important while GCNs are used for tasks like node classification, link prediction, community detection, and other graph-based problems where understanding the relationships between data points is crucial. A GCN \cite{gcn_csn_2019} consists of several convolutional layers and a single convolutional layer can be expressed as 
\begin{align}\label{eq:conv_eq}
    \textbf{h}^{(l+1)} = \sigma(\textbf{D}^{-\frac{1}{2}} \textbf{A }\textbf{D}^{-\frac{1}{2}}\textbf{h}^{(l)}\textbf{W}^{(l)})
\end{align}
where $\textbf{h}^{(l)}$ and $\textbf{h}^{(l+1)}$ are the feature vectors at layers $l$ and $l+1$, respectively. $\textbf{A}$ is the adjacency matrix of the graph with self-connections, $\textbf{D}$ is the degree matrix of $\textbf{A}$ defined as a diagonal matrix where each diagonal element $D_{ii}$ is the sum of the corresponding row of $\textbf{A}$, $\textbf{W}^{(l)}$ is the weight matrix of layer $l$, and $\sigma$ is an activation function, such as ReLU or sigmoid. The applications of GCNs include Intrusion Detection \cite{Deng_tnsm_2023}, Cooperative Spectrum Sensing \cite{Janu_tvt_2023}, Channel and Resource Allocation \cite{Nakashima_access_2020,Zhao_sensor_2020}, Modulation Recognition \cite{liu_wcl_2020} and Joint Routing and Scheduling \cite{yang_iotj_2022}.

\textbf{Graph Attention Network.}
The Graph Attention Network (GAT) \cite{Petar_arxiv_2017} is a type of neural network architecture designed for processing data structured as graphs or networks. GAT is particularly useful for tasks such as node classification, link prediction, and graph classification, where information propagation and aggregation across graph nodes are crucial. GAT was introduced to improve how nodes in a graph aggregate information from their neighbors by applying attention mechanisms. Attention mechanisms are a crucial component in modern neural network architectures, and they have significantly improved the performance of various deep learning models across a wide range of tasks, especially in natural language processing and computer vision. At their core, attention mechanisms enable models to focus on specific parts of the input data when making predictions or generating output. The mathematical expression for an attention mechanism coefficient for GAT \cite{Petar_arxiv_2017} can be expressed as  
\begin{align}
    e_{ij}^{(l)} = a(\textbf{W}\textbf{h}^{(l)}_i, \textbf{W}\textbf{h}^{(l)}_j)
\end{align}
where $a$ is the shared attentional mechanism. This equation indicates the importance of node $j$'s feature to node $i$. To easily compare the coefficients across different nodes, the coefficient is normalized across all choices of $j$ using a softmax function and is given as
\begin{align}\label{eq:aij}
 \alpha_{ij}^{(l)} = \rm{softmax}(\textit{e}_{ij}^{(l)}) = \frac{\rm{exp}(\textit{e}_{\textit{ij}}^{(l)})}{\sum_{\textit{j} \in N_{i}}\rm{exp}(\textit{e}_{\textit{ij}}^{(l)})}
\end{align}
A LeakyReLU nonlinearity is applied to the Attention mechanism $a$ and can be expressed as 
\begin{align}\label{eq:eij}
    e_{ij}^{(l)} = \rm{LeakyReLU}(\textbf{a}^{(l)^{T}}[\textbf{W}^{(l)}\textbf{h}_i^{(l)}||\textbf{W}^{(l)}\textbf{h}_j^{(l)}]
\end{align}
where $\textbf{a}$ is the weight vector for attention mechanism $a$, $(\cdot)^{T}$ and $||$ are transpose and concatenation operators, respectively. Applying \eqref{eq:eij} in \eqref{eq:aij}, we get
\begin{align}\label{eq:aij2}
 \alpha_{ij}^{(l)} =  \frac{\rm{exp}(\rm{LeakyReLU}(\textbf{a}^{(l)^{T}}[\textbf{W}^{(l)}\textbf{h}_i^{(l)}||\textbf{W}^{(l)}\textbf{h}_j^{(l)}])}{\sum_{j \in N_{i}}\rm{exp}(\rm{LeakyReLU}(\textbf{a}^{(l)^{T}}[\textbf{W}^{(l)}\textbf{h}_i^{(l)}||\textbf{W}^{(l)}\textbf{h}_j^{(l)}])}
\end{align}
GATs are used in applications such as Channel Estimation \cite{Tekbiyik_icc_2021}, Malware Detection \cite{Catal_mal_elect_2021}, Malicious Traffic Classification\cite{zhang_apnoms_2023} Network Slicing Management \cite{shao_wcnc_2021}, Task Offloading \cite{Zhong_wcsp_2021}, Beamforming \cite{li_wcl_2023},  Predictive Channel Modelling \cite{li_vtc_2023}.  

\textbf{Graph Sample and AGgregatE (GraphSAGE).}
GraphSAGE is a graph representation learning technique that aims to generate embeddings for nodes in a graph. Embedding refers to the process of representing discrete variables as continuous, low-dimensional vectors. There are 3 types of embedding namely Node Embedding, Edge Embedding, and Graph Embedding. Node embedding involves learning a vector representation for each node in the graph such that nodes with similar network neighborhoods or connectivity patterns have similar embeddings. These embeddings capture the structural and relational information of nodes in the graph. GraphSAGE starts by sampling a fixed-size neighborhood for each node in the graph. This neighborhood typically consists of the node itself and its immediate neighbors. The sampling process can be random or based on some specific strategy. For each node, GraphSAGE aggregates information from its sampled neighborhood based on an aggregator function. After aggregating information from the neighborhood, GraphSAGE generates a new embedding for each node. This embedding is learned through a neural network layer, which can be shared across all nodes or specific to each node. GraphSAGE combines techniques from graph sampling and aggregation to create node embeddings that capture both local and global graph structures. The mathematical modeling for GraphSAGE embedding generation \cite{Will_Graphsage_2017} can be expressed as follows. First, each node $i$ aggregates the representations of its neighboring nodes into a single vector $\textbf{h}_{N(i)}^{l}$. Then, GraphSAGE concatenates the node's current representation $\textbf{h}_{i}^{l}$ with the aggregates neighborhood vector $\textbf{h}_{N(i)}^{l}$ which is fed through a fully connected layer with non-linear activation function $\sigma$ to get $\textbf{h}_{i}^{l+1}$. This can be expressed as
\begin{align}
    \textbf{h}_{N(i)}^{(l+1)} = A({\textbf{h}_{j}^{l}, \forall j \in N_{i}}),\label{eq:gsage1}\\
    \textbf{h}_{i}^{(l+1)} = \sigma(\textbf{W}^{(l)} \cdot C(\textbf{h}_{i}^{(l)},\textbf{h}_{N_{i}}^{(l+1)}))\label{eq:gsage2}
\end{align}
where $N_{i}$ is the set of neighboring nodes of $i$, $\textbf{W}^{(l)}$ is the weight matrix of layer $l$, $A(\cdot)$ and $C(\cdot)$ are the aggregator and concatenator functions. There are three common types of aggregators namely Mean Aggregator, LSTM aggregator, and Pooling Aggregator. GraphSAGE is advantageous because it combines the power of both local and global information in graph data. GraphSAGE has been widely used for Intrusion Detection \cite{wai_gsage_2022}, MANET traffic analysis \cite{Tekdogan_ICMLANT_2022}, Access Point Selection \cite{Ranasinghe_globecom_2021}, Routing \cite{lu_sensor_2023}, Recommender Systems \cite{Yassine_eaai_2023} and Network Traffic Prediction \cite{liu_tits_2023}.

\textbf{Graph Autoencoders.}
Graph Autoencoders (GAEs) are a type of neural network architecture designed for learning graph representations by encoding and decoding the graph structure. GAEs are used in unsupervised or self-supervised learning settings to capture meaningful latent representations of nodes or graphs. They enable the discovery of meaningful latent representations that can reveal the underlying structure and relationships within complex graphs. Researchers continue to explore variations and extensions of GAEs to improve their performance and applicability in different domains. GAE consists of two main components namely the \textbf{\textit{Encoder}} which takes graph-structured data as input and maps it into a lower-dimensional latent space. It typically consists of graph convolutional layers that aggregate information from neighboring nodes while considering the graph's topology. The output of the encoder is a compressed representation of the input graph and \textbf{\textit{Decoder}} which takes the latent representation from the encoder and attempts to reconstruct the original graph. It maps the latent space back to the graph structure. The decoder's goal is to generate a graph that is as close as possible to the input graph while minimizing reconstruction loss. The mathematical model for the encoder, decoder, and the loss functions \cite{Miao_arxiv_2022} can be expressed as follows,
\begin{align}
    \rm{\textbf{enc}} &= f_{\rm{enc}}(\rm{\textbf{inp}};\theta_{\rm{enc}}),\\
    \rm{\textbf{dec}} &= f_{\rm{dec}}(\rm{\textbf{enc}};\theta_{\rm{dec}}),
\end{align}
where $\rm{\textbf{enc}}$ and $\rm{\textbf{dec}}$ are the encoder and decoder output vector, $f_{\rm{enc}}$ and $f_{\rm{dec}}$ are encoder and decoder functions, $\rm{\textbf{inp}}$ is the input vector, $\theta_{\rm{enc}}$ and $\theta_{\rm{dec}}$ are parameters associated with $f_{\rm{enc}}$ and $f_{\rm{dec}}$, respectively. The reconstruction loss function is expressed as 
\begin{align}
    \rm{Loss} &= \frac{1}{\mathcal{N}}\sum_{i=0}^{\mathcal{N}}(||S_{\rm{inp},\textit{i}}-S_{\rm{dec},\textit{i}}||^2)
\end{align}
where $\mathcal{N}$ is the number of elements in $\rm{\textbf{inp}}$, $S_{\rm{inp}}$ and $S_{\rm{dec}}$ are the sum of elements in $\rm{\textbf{inp}}$ and $\rm{\textbf{dec}}$, respectively, $||(\cdot)||^2$denotes the squared Euclidean distance between the input and reconstructed features. GAEs are widely used for Anomaly Detection \cite{Sun_ict_2021,Miao_arxiv_2022}, Wireless Sensor Network Topology Reconstruction \cite{Zhang_mse_2023}, Indoor Localization \cite{Zahra_sp_2021}, Intrusion Detection \cite{Venturi_arganids_2023} and Task Assignment \cite{Liu_tagae_iotj_2023}.

\begin{table*}[htbp]
\caption{Comparison of key features, strengths, and weaknesses of different GNN variants.}
\renewcommand{\arraystretch}{2.2} 
\begin{tabular}{|m{2cm}|m{3cm}|m{3cm}|m{3cm}|m{2cm}|m{2cm}|}
\hline
\textbf{GNN Variant} & \textbf{Key Features} & \textbf{Strengths} & \textbf{Weaknesses} & \textbf{Dynamic Environment Suitability} & \textbf{Expressive Power}
\\

\hline \hline

\textit{Graph Convolutional Network (GCN)} 
& Spectral-based method; 
Aggregates features from neighboring nodes. 
& Simple and effective;
Performs well on homophilic graphs. 
& Assumes homogeneity in neighborhoods;
Computationally expensive for large graphs.
&Limited 
&Low \\
\hline

\textit{Graph Attention Network (GAT)} 
& Employs attention mechanism to weight neighbors;
 Learnable importance for neighbors.
\newline 
Handles varying neighborhood sizes. 
&  Differentiates between neighboring nodes;
Flexible across diverse graph structures. 
& High computational cost due to attention calculations; 
 Prone to overfitting on small datasets.
& Limited
& Low \\
\hline

\textit{Graph Sample and AGgregatE (GraphSAGE)} 
& Inductive learning method;
 Samples neighbors for aggregation;
Designed for large-scale graphs. 
& Scales well to large and dynamic graphs;
Suitable for real-time and inductive settings. 
& Sampling strategy affects performance;
 May be less accurate than transductive approaches. 
& Suitable
& Moderate \\
\hline

\textit{Graph Autoencoder (GAE)} 
& Encoder-decoder architecture;
Learns node embeddings unsupervised;
Common for link prediction and clustering. 
& Effective for unlabelled graphs;
Simple design with strong applications in unsupervised tasks. 
& Depends on well-designed loss functions;
Limited in handling complex graph dynamics;
May require large datasets.
& Not ideal
& Moderate \\
\hline

\textit{Graph Isomorphism Network (GIN)} 
& Uses MLP for neighborhood aggregation;
High expressive power;
Designed to distinguish graph structures. 
& Powerful for graph classification; 
Strong discriminative capability. 
& Requires careful hyperparameter tuning;
Computationally intensive;
May be over-parameterized for simple graphs. 
& Limited
& High \\
\hline

\textit{Graph Capsule Network (GCapsNet)} 
& Uses capsule networks for graphs;
Captures hierarchical and part-whole relationships. 
& Handles complex and hierarchical graph structures;
Potentially more robust to noise. 
& Complex architecture;
High computational cost.
& Limited
& High \\
\hline

\textit{Spatial-Temporal Graph Network (STGN)} 
& Integrates spatial and temporal dependencies;
Combines well with RNNs/LSTMs;
Designed for dynamic graphs. 
& Effective for time-series prediction on graphs;
Models evolving graph behavior well. 
& Requires extensive preprocessing;
Training is complex and data-intensive.
& Best suited
& High\\
\hline

\textit{Graph Reinforcement Learning (GRL)} 
& Learns graph policies via reinforcement learning;
Can be integrated with various GNNs;
Focused on task-specific embeddings. 
& Enables multi-task learning;
Facilitates transfer learning;
Effective for high-dimensional state spaces. 
& Embedding quality depends heavily on training;
Requires large datasets; 
Complex to tune and deploy. 
& Varies based on underlying GNN architecture
& Varies based on underlying GNN architecture \\
\hline

\end{tabular}
\label{tab:comp_gnn}
\end{table*}

\textbf{Graph Isomorphic Networks.}
The Graph Isomorphism Network (GIN) is a type of neural network architecture designed for solving the graph isomorphism problem. The graph isomorphism problem involves determining whether two given graphs are isomorphic, meaning they have the same structure but might differ in the labels assigned to their nodes. A simple test called the Weisfeiler-Lehman test \cite{wl_iso_test_1968} can be conducted to determine if the graphs are isomorphic or not. The idea behind this test is to iteratively aggregate the labels of the nodes and their neighbors and hash the aggregated labels into unique new labels. If the labels of the nodes differ at some point in the iteration, then the graph is said to be non-isomorphic.

GINs are built upon GNNs and the key idea behind GINs is to aggregate information from neighboring nodes iteratively, updating the representation of each node based on its neighbors. By stacking multiple layers of such aggregation, GINs can capture increasingly complex graph patterns. Additionally, GINs incorporate techniques to ensure that they are permutation-invariant, meaning they can handle graphs with nodes in arbitrary order without affecting the result. A single layer of GIN \cite{keyulu_gin_2019} can mathematically expressed as  
\begin{align}
    \textbf{h}_{i}^{(l+1)} = \rm{MLP}^{(l+1)}((1+\epsilon^{(l+1)}h_{i}^{(l)}+\sum_{j \in N_{i}}\textbf{h}_{j}^{(l)}))
\end{align}
where $\textbf{h}_{i}^{(l)}$ and $\textbf{h}_{i}^{(l+1)}$ are the representation of node $i$ at layers $l$ and $l+1$, respectively. $N_{i}$ is the set of neighboring nodes of $i$, $\epsilon$ is a learnable or a fixed scalar parameter and $\rm{MLP}$ is a feed-forward multi-layer perceptron. GIN has shown promising results and has been applied to various tasks related to learning and classification \cite{gin_bert_wcnc_2023, IsoNN_lin_rxiv}, network authentication \cite{Tran_gin_2008}, malicious node detection \cite{wai_xgbot_2023}, network traffic prediction \cite{Wang_iiot_tvan_2022}.

\textbf{Graph Capsule Networks.} Graph Capsule Networks (GCapsNet) are a type of neural network architecture inspired by capsule networks originally introduced for image analysis tasks. Graph Capsule Networks extend this concept to graph-structured data, allowing them to handle relational reasoning and hierarchical representations within graph data. In traditional CNNs, convolutional layers typically extract features from images by applying convolutional filters across spatial dimensions. Capsule networks introduce the idea of capsules, which are groups of neurons that represent instantiation parameters of entities within an image, such as object pose, scale, orientation, etc. These capsules allow for more robust and interpretable representations of objects compared to traditional CNNs. An example of GCapsNet called Graph Capsule Convolution Neural Network (GCNN) is introduced in \cite{Saurabh_Gcaps_2018}. For node $i$ with node feature value of $x_{0}$ and a set of neighborhood node feature values $\mathcal{F}_{i} = \{x_0, x_1, x_2, ..., x_j\}$ including node $i$ the output of graph capsule function is $f:\mathbb{R}^{j}\rightarrow\mathbb{R}$ can be expressed as 
\begin{align}\label{eq:gcf}
    f_{i}(x_0, x_1, ..., x_j) = \frac{1}{|\mathcal{F}_i|} \sum_{j\in \mathcal{F}_i}\lambda_{ij}x_j
\end{align}
where $\lambda_{ij}$ is the edge weight between nodes $i$ and $j$. For a graph capsule network that captures higher-order statistical moments, $f_{i}(x_0, x_1, ..., x_k)$ is replaced with a vector values capsule function $f:\mathbb{R}^{j}\rightarrow\mathbb{R}^{p}$, where $p$ is the number of instantiation parameters described in \cite{instantiation_ann}. Then Equation~\eqref{eq:gcf} can be rewritten as
\begin{align}\label{eq:gcf2}
    f_{i}(x_0, x_1, ..., x_j) = \frac{1}{|\mathcal{F}_i|} \begin{bmatrix}
\sum_{j\in \mathcal{F}_i}a_{ij}x_j \\
\sum_{j\in \mathcal{F}_i}a_{ij}x^2_j \\
\vdots \\
\sum_{j\in \mathcal{F}_i}a_{ij}x^p_j
\end{bmatrix}
\end{align}
where each row denotes statistical moments, such as mean, standard deviation, skewness, and other higher statistical moments. Due to the extensive features, GCaps finds its application in various domains, such as Recommendation systems \cite{Chen_KHGCN_2023}, Classification \cite{Yang_NCGNN_2024}, Malware Detection \cite{Deng_MDHE_2023}.

\textbf{Spatial-Temporal Graph Networks.}
Spatial-Temporal Graph Networks (STGNs) are a type of deep learning model designed to handle data that exhibit both spatial and temporal dependencies. They are particularly suited for tasks where understanding the interactions and dynamics across space and time is essential. STGNs operate on data represented as spatio-temporal graphs, where nodes represent spatial entities such as locations, sensors, and edges capture temporal dependencies such as timestamps, and time intervals.

STGNs integrate techniques from graph neural networks (GNNs) and temporal modeling to effectively capture spatial relationships and temporal dynamics within the data. They typically consist of multiple layers, including graph convolutional layers, temporal modeling layers, and attention mechanisms, to process and learn from the spatio-temporal graph data. A single convolutional layer can be expressed as in Eq.~\ref{eq:conv_eq}, which can then be modified to account for spatial and temporal dependencies as follows \cite{ghosh_wacv_2020}.
\begin{align}
    \textbf{h}^{(l+1)} = \sigma(\textbf{D}_{t}^{-\frac{1}{2}} \textbf{A}_{t}\textbf{D}_{t}^{-\frac{1}{2}}\textbf{h}_{s}^{(l)}\textbf{W}_{t}^{(l)})\\
    \textbf{h}_{s}^{(l)} = \textbf{D}_{s}^{-\frac{1}{2}} \textbf{A}_{s}\textbf{D}_{s}^{-\frac{1}{2}}\textbf{h}^{(l)}\textbf{W}_{s}^{(l)}
\end{align}
where $\textbf{h}^{(l)}$, $\textbf{h}^{(l+1)}$ are the feature vectors, at layers $l$ and $l+1$, respectively, $\textbf{h}_{s}^{(l)}$ is the spatial feature vector, $\textbf{A}_{t}$ and $\textbf{A}_{s}$ are temporal and spatial adjacency matrices, $\textbf{D}_{t}$ is the degree matrix of $\textbf{A}_{t}$, and $\textbf{D}_{s}$ is the degree matrix of $\textbf{A}_{s}$. $\textbf{W}_{t}^{(l)}$ and $\textbf{W}_{s}^{(l)}$ are temporal and spatial weight matrix of layer $l$, and $\sigma$ is an activation function, such as ReLU or sigmoid. Finally, STGN (ie.e, STGCN) can be expressed as
\begin{align}
    \textbf{h}_{t}^{(l+1)} = \sigma(\textbf{D}_{s}^{-\frac{1}{2}} \textbf{A}_{s}\textbf{D}_{s}^{-\frac{1}{2}}\textbf{h}^{(l)}\textbf{W}_{s}^{(l)}\textbf{W}_{t}^{(l)})
\end{align}

STGNs are commonly used for Multi-object Tracking \cite{Xue_stt_2021}, IoT Intrusion Detection \cite{Wang_stgam_tii,Sharadqh_Hybrid_2023} and Cellular Traffic Prediction \cite{Wang_tii_2023, Zhao_ieeecl_2022}.

\textbf{Graph Reinforcement Learning.} Graph Reinforcement Learning (GRL) is a special type of GNN that uses RL for real-time interaction with the environment and decision-making. RL is used with GNN due to its ability to learn by interacting with the environment making it suitable for complex, dynamic, and uncertain environments where there is no prior knowledge of the network to train the GNN model. GNNs are commonly combined with RL techniques such as Q-Learning \cite{watkins_qlearn}, REINFORCE \cite{rj_reinforce}, Actor-Critic \cite{konda_actor}, among others. This has motivated researchers to apply GRL for various problems such as Multiaccess Edge Computing \cite{sun_iotj_2023}, Joint Cruise Control and Task Offloading \cite{li_iotj_2022}, Resource Allocation \cite{Nakashima_access_2020,Zhao_sensor_2020}. In addition to the combination of simple GNN with RL, there is also the combination of different variants of GNNs with RL such as GCN-RL \cite{Jiechuan_arxiv_2020}, GAT-RL \cite{shao_wcnc_2021} and GraphSAGE-RL \cite{jiaqi_icaiic_2024}.

In Table~\ref{tab:comp_gnn}, we compare the key features, strengths, weaknesses, suitability for dynamic environment and expressive power of different GNN variants to help readers choose the best variant based on the application's needs.

\begin{figure*}
\centering
\includegraphics[width=0.6\textwidth]{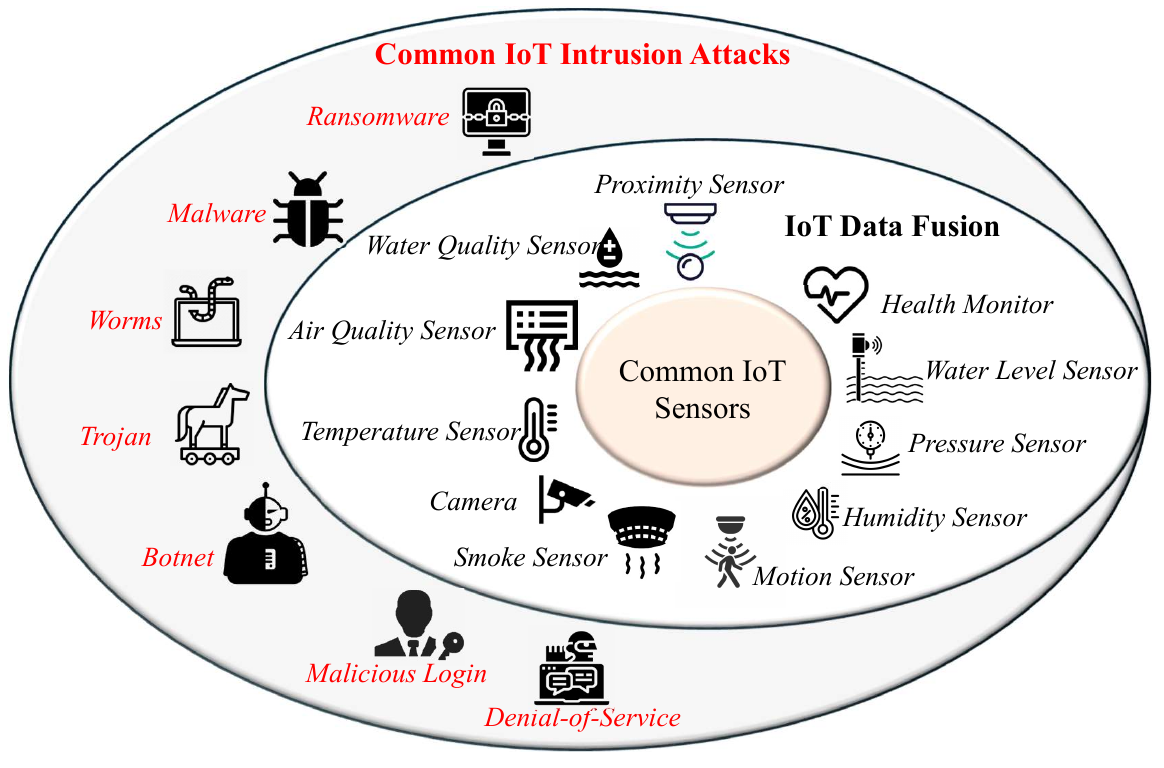}
\caption{Common IoT Sensors and Intrusion Attacks.}
\label{fig:iot_fig}
\end{figure*}

\section{GNN for Internet of Things}\label{sec:gnn_iot}
Internet of Things refers to a network of interconnected devices that can communicate and share data with each other over the internet, without requiring direct human interaction. These devices are embedded with sensors, software, and other technologies that enable them to collect and exchange data, as well as perform specific functions. In Section~\ref{ssec:iot_smart} we first discuss the application of GNN for different IoT smart applications. There are two major challenges with IoT systems. First, due to the vast number of devices that are heterogeneous in nature, it is important to collect all the data and combine it to extract the relevant information. Second, it is also important to safeguard the devices from adversaries that may infiltrate the network and spoof or steal valuable data. Therefore it is important to address these two challenges. With this challenge as the motivation, IoT Data Fusion and IoT Network Intrusion have been extensively studied. Unlike CNNs, which work best on grid-like structures (e.g., images), and RNNs, which excel at sequential data (e.g., time series data), GNNs can process non-grid-like and non-sequential unstructured data where relationships are more complex and are not fixed. This makes them ideal for aggregating and using data from heterogeneous IoT applications. To this end, we discuss the applications of GNN for IoT Data Fusion in Section~\ref{ssec:iotdatafusion} and Network Intrusion Detection in Section~\ref{ssec:iot_nid}. Figure~\ref{fig:iot_fig} shows representative IoT sensors and intrusion attacks. Table~\ref{tab:gnn_iot_table} provides a high-level comparison of surveyed paper using GNNs for IoT applications. We also provide the dataset or the simulator used for performance evaluation.

\begin{table*}[htbp]
\centering
\caption{Comparison of surveyed papers using GNNs for Internet of Things applications.}
\renewcommand{\arraystretch}{1.7}
{\begin{tabular}{|m{2.3cm}|m{2.5cm}|m{3.5cm}|m{3.5cm}|m{3.5cm}|}
\hline

\textbf{Section} & \textbf{Paper} & \textbf{Application} & \textbf{ML Algorithm Used} &  \textbf{Dataset Used} \\
\hline
\hline
\multirow{5}{=}{\textbf{Section \ref{ssec:iot_smart}}\\IoT Smart Applications} 
& \textit{Li et al.} \cite{li_tosn_2021} & Human Pose Forecasting System & Graph Convolutional Neural Network & Human 3.6M dataset \cite{human36M_2014}, Penn Action dataset \cite{penn_action_2013} \\
\cline{2-5}
& \textit{Yanmei et al.} \cite{jiang_smart_energy_2023} & Short-term load prediction & Graph Attention Network + LSTM & Short-term power load data - Shijiazhuang
power supply company\\
\cline{2-5}
& \textit{Zhang et al.} \cite{shiyu_smart_campus_2022} & Crowd flow prediction & Spatio-temportal GNN  & University of Macau Campus WiFi Data \\
\cline{2-5}
& \textit{Sharma et al.} \cite{sharma_smart_city_2023} & Real-time traffic-speed estimation &   Spatio-Temportal Graph Attention Network  & PeMS Data \cite{Chao_pems} \\
\cline{2-5}
& \textit{Pamuklu et al.} \cite{Pamuklu_smart_agri_2023} & Smart agriculture & GNN-RL & Simulated dataset \\
\hline

\multirow{2}{=}{\textbf{Section \ref{ssec:iotdatafusion}}\\IoT Data Fusion} 
& \textit{Ma et al.} \cite{Zhengjing_jksu_2022} & Air quality prediction & Spatio-Temporal Graph Convolutional Network & Air Pollution and Meteorological Data \cite{Wang_aqi} \\
\cline{2-5}
& \textit{Yang et al.} \cite{yang_mdpijgi_2023} & Pedestrian trajectory prediction & Spatio-Temporal Graph Convolutional Network & ETH \cite{Pellegrini_eth}, UCY \cite{Pellegrini_eth} \\
\hline

\multirow{7}{=}{\textbf{Section \ref{ssec:iot_nid}}\\IoT Network Intrusion Detection} 
& \textit{Lo et al.} \cite{wai_gsage_2022} & IoT intrusion detection & GraphSAGE & ToN-IoT \cite{Alsaedi_toniot}; BoT-IoT \cite{Nickolaos_botiot}; NF-TON-IoT and NF-BoT-IoT \cite{Sarhan_nfdata} \\
\cline{2-5}
& \textit{Caville et al.} \cite{evan_anomal_2022} & Anomaly and intrusion detection & GraphSAGE & NF-UNSW-NB15-v2 \cite{sarhan_mna_2022}; NF-CSE-CIC-IDS2018-v2 \cite{Moustafa_mcis} \\
\cline{2-5}
& \textit{Zhang et al.} \cite{Zhang_tnse_2022} & Industrial IoT Intrusion Detection & Graph Convolutional Network & Industrial Control System Traffic Data Sets \cite{morris_iccip_2014}  \\
\cline{2-5}
& \textit{Pujol-Perich et al.} \cite{pujal_sigm_2022} & Network Intrusion Detection & MPNN GNN & CIC-IDS2017 \cite{Sharafaldin2018TowardGA} \\
\cline{2-5}
& \textit{Xiao et al.} \cite{haitao_capitd_2023} & Malicious insider threat detection & Graph Capsule Networks & Universal insider threat dataset CERT v4.2\cite{glasser_bridge_2013} \\
\cline{2-5}
& \textit{Liu et al.} \cite{Liu_trustcom_2020} & Malicious login detection & GNN+CNN & Los Alamos National Lab’s (LANL’s) comprehensive cybersecurity
events dataset \cite{kent_dataset}, Own corporate dataset \\
\cline{2-5}
& \textit{Busch et al.} \cite{busch_ssdbm_2021} & Malware detection and classification & Graph Autoencoder & CICAndMal2017 \cite{Lashkari_iccst_2018} \\
\hline

\end{tabular}}
\label{tab:gnn_iot_table}
\end{table*}
\subsection{IoT Smart Applications}\label{ssec:iot_smart}
In simple terms, IoT smart applications are those that leverage data gathered from the users to improve their experience and quality of life through the process of data analytics and use AI/ML for recommending insights based on user's past activity. As shown in Figure~\ref{fig:iot_appln} there are several IoT smart applications. In this section, we will focus on a subset of applications that leverage GNN for the data analytics and forecasting \cite{li_tosn_2021, jiang_smart_energy_2023, shiyu_smart_campus_2022}, real-time estimation and decision-making \cite{sharma_smart_city_2023}, efficient task offloading and planning \cite{Pamuklu_smart_agri_2023}. In \cite{li_tosn_2021} the authors propose a GCN-based hybrid cloud-edge system called GPFS, a graph-based pose forecasting system that is capable of online learning that can significantly improve the forecasting accuracy in new environments. This system is specifically implemented for a smart home deployment. In \cite{jiang_smart_energy_2023} the authors propose a GAT-LSTM-based prediction algorithm called EnGAT-BiLSTM for load prediction in smart energy applications where GAT is used to extract high-quality load spatio-temporal features while bi-directional LSTM (BiLSTM) is used for lifelong learning of load prediction. CrowdTelescope \cite{shiyu_smart_campus_2022} is a Wi-Fi positioning-based multi-grained spatio-temporal crowd flow prediction framework that leverages the extracted human mobility trace from the Wi-Fi connection records on the university campus which can then be used to predict the crowdflow within the campus with different granularities.
In \cite{sharma_smart_city_2023} the authors propose a Spatio-Temporal Graph Attention Network-based solution for real-time traffic-speed estimation for smart city applications by considering the structure and characteristics of road networks. In \cite{Pamuklu_smart_agri_2023} the authors propose a GNN-RL-based solution to optimize the task offloading from IoT devices to the UAVs for smart agriculture applications. The objective of the solution is to reduce the task deadline violations and optimize the energy usage of the UAVs to extend their mission time.

\subsection{IoT Data  Fusion}\label{ssec:iotdatafusion}

In the previous section, we discussed several IoT smart applications that gather different types of data and how it is used for different applications. There is one important step within different IoT smart applications which is called IoT data fusion. IoT data fusion, also known as information fusion, sensor data fusion, or multi-sensor data fusion, refers to the process of integrating and combining data from multiple IoT devices and sensors to obtain a more comprehensive and accurate understanding of a physical or digital environment. The primary goal of IoT data fusion is to extract valuable insights, make informed decisions, and improve the overall efficiency and effectiveness of IoT systems. To implement IoT data fusion, various techniques and technologies can be used, including data preprocessing, sensor calibration, data alignment, feature extraction, ML algorithms \cite{Zhengjing_jksu_2022, yang_mdpijgi_2023}, and data visualization tools. The choice of fusion method depends on the specific use case and the characteristics of the data sources involved. 

For example, in \cite{Zhengjing_jksu_2022} the authors propose a deep learning method for fusing heterogeneous data collected from multiple monitoring stations using a Graph Convolutional Network specifically a spatiotemporal graph convolutional network. The overall idea behind this approach is to use the location of the stations and fuse the collected heterogeneous data and then predict the future trend based on global information rather than local information. Through this approach, better performance is guaranteed when compared to using data from individual stations. Applications such as surveillance systems and autonomous driving require the prediction of trajectories of multiple agents. These are affected by many factors such as each agent's historical trajectory, and interactions between agents. In  \cite{yang_mdpijgi_2023} the authors propose a trajectory prediction model based on the fusion of prior awareness and past trajectory information. Similar to \cite{Zhengjing_jksu_2022}, this paper also leverages the capabilities of a spatio-temporal convolutional network. 

\subsection{IoT Network Intrusion Detection}\label{ssec:iot_nid}
In the previous section we discussed the applications of GNN for IoT data fusion where data from different heterogeneous data are collected and analyzed for various applications such as surveillance, and autonomous driving among others. But the major challenge associated with these kinds of data generating and data storing applications is that they are susceptible to various attacks as shown in Fig.~\ref{fig:iot_fig}. These kinds of attacks are also called Network Intrusion attacks. It is important to design countermeasure approaches to overcome these challenges. These approaches are broadly called IoT Network Intrusion Detection approaches. IoT Network Intrusion Detection (NID) refers to the process of monitoring and analyzing the traffic and behavior within an IoT network to identify and respond to unauthorized or malicious activities. IoT devices are increasingly being integrated into various industries and applications, ranging from smart homes and industrial automation to healthcare and transportation. However, the proliferation of IoT devices also presents new security challenges and vulnerabilities, making it crucial to have effective intrusion detection mechanisms in place for safeguarding IoT ecosystems, as these devices often have limited computing resources and may lack the robust security features found in traditional computing devices. By implementing effective intrusion detection systems, organizations can better protect their IoT networks from various threats, including malware, unauthorized access, data breaches, and other security risks. Below we discuss a few interesting approaches proposed for IoT NID. These are divided into general intrusion detection applications  \cite{wai_gsage_2022, evan_anomal_2022, Zhang_tnse_2022,pujal_sigm_2022, haitao_capitd_2023} and Malicious logins and Malware detection \cite{Liu_trustcom_2020,busch_ssdbm_2021}.

In \cite{wai_gsage_2022}, the authors present a GNN-based approach called Edge-Graph SAmple and aggreGatE aka E-GraphSAGE that allows capturing both the edge features as well as the topological information for network intrusion detection in IoT networks. Compared to traditional GraphSAGE \cite{Will_Graphsage_2017} which mainly focuses on node features for node classification but falls short on edge classification using edge features. Classification based on edge features allows us to classify different network flows to the nodes into benign and attack flows which can be achieved using E-GraphSAGE. Similarly, in \cite{evan_anomal_2022}, the authors propose Anomal-E, a E-GraphSAGE-based approach to detect both anomaly and intrusion in computer networks using edge features and a graph topological structure in a self-supervised manner. In \cite{Zhang_tnse_2022}, the authors present a general framework for intrusion detection called Graph Intrusion Detection designed based on GNN for industrial IoT applications that are highly susceptible to cyber-attacks. This framework is designed to learn both the network topology as well as the GNN network parameter at the same time. While there are several ML-based intrusion detection techniques, they are not robust to common adversarial attacks. To address this challenge, in \cite{pujal_sigm_2022} the authors present a novel GNN-based model designed to process and learn from graph-structured information that keeps flow records and their relationships compared to other techniques that focuses only on the different flows between the source and the destination hosts. In \cite{haitao_capitd_2023}, the authors propose CapsITD, a novel user-level threat detection method using GCapsNet. The GCapsNet is used to learn the statistical features of users' daily activities and then use the learned features to detect any malicious activity and threats.

Another possible network intrusion scenario is malicious logins and the use of malware to gain access to the network. In \cite{Liu_trustcom_2020}, to address the challenge of malicious login, the authors present a GNN-based solution called MLTracer. This framework consists of two key features. First, it enables differentiating crucial attributes of logins such as the source/destination IDs, roles, port, and authentication type, among others. Second, a GNN model in conjunction with CNN is used to detect malicious logins. In \cite{busch_ssdbm_2021}, the authors present an autoencoder-based Network Flow GNN for malware detection and classification. The framework extracts the flow graphs to classify them using a novel edge feature-based graph neural network model. Three derived model variants are presented namely a graph classifier, a graph autoencoder, and a one-class graph neural network.

\subsection{Lessons Learned}
In this section, we present the key takeaways and lessons learned from the applications of GNN for IoT applications discussed in Sections~\ref{ssec:iot_smart} - \ref{ssec:iot_nid}.
\begin{enumerate}
    \item Traditional ML models including deep learning architectures like CNNs fail to leverage the relational information present in IoT networks due to structured grid-like data whereas GNN are specifically designed to process graph-structured data, allowing them to capture complex inter-dependencies, dynamically update node representations based on their neighbors, and better model IoT communication patterns. This capability is crucial in IoT applications, where network effects, communication patterns, and distributed sensing play a significant role in data processing and decision-making.
    \item  As discussed earlier, GNNs propagate information across edges in a graph, aggregating and transforming node features. However, as the number of nodes and edges in an IoT network grows, the computational burden increases significantly. Each node's representation is influenced by its neighbors, and in deep GNN architectures, this information propagates across multiple layers, leading to an exponential increase in computation. This scalability challenge makes it difficult to deploy GNNs in large-scale IoT networks, where thousands or even millions of devices may be interconnected.
    \item Many IoT applications, such as autonomous driving, industrial automation, and smart healthcare, require real-time decision-making. GNNs, however, involve iterative message-passing mechanisms that increase inference time. Unlike simpler models that can quickly process incoming data, GNNs may struggle to meet the low-latency requirements of real-time IoT systems. This limitation necessitates optimizations such as model compression, approximate inference techniques, or hardware acceleration to make GNNs viable for real-time scenarios.
    \item IoT ecosystems generate diverse data types such as temperature and humidity readings from environmental sensors, images from surveillance cameras, text-based logs from communication protocols, and more. GNNs must integrate and process this heterogeneous data efficiently while maintaining meaningful graph representations. Achieving this requires specialized architectures, such as multi-modal GNNs, that can incorporate different data modalities into a unified learning framework. Traditional ML models treat different data types separately and often require handcrafted feature fusion techniques, which are inefficient and inflexible. GNNs, however, naturally accommodate heterogeneous data by encoding different modalities into node and edge features. Models like Heterogeneous GNNs (H-GNNs) \cite{ba_iotj_2023, wei_iotj_2024} are designed specifically to handle multi-modal IoT data, making them far more adaptable and effective.
    \item IoT systems are often deployed in open environments, making them susceptible to cyber threats such as data spoofing, node compromise, and adversarial attacks on machine learning models. GNNs, like other deep learning models, can be fooled by subtle perturbations in the input data, leading to incorrect predictions or system failures. Developing robust GNN architectures that can detect and mitigate adversarial attacks is essential for maintaining the security and reliability of IoT networks.  Traditional ML models lack built-in robustness against adversarial threats because they assume independent and identically distributed (IID) data points. GNNs, however, can leverage their relational structure to detect anomalous patterns and use graph-based anomaly detection methods to identify malicious behavior. Additionally, techniques such as adversarial training \cite{li_tkde_2023, liang_arxiv_2022} and robust graph learning algorithms \cite{zhu_kdd_2019, qian_enn_2024} can enhance GNN’s resilience against security threats in IoT networks.
    \item Many IoT devices run on battery power or energy-harvesting techniques, limiting their computational capacity. The high computational requirements of GNNs pose a challenge in such environments, as frequent model updates and inference operations can quickly deplete energy reserves. Techniques like model pruning, quantization, and edge computing strategies are needed to adapt GNNs for energy-efficient deployment in resource-constrained IoT settings. Additionally, deploying tiny GNNs \cite{ahmed_arxiv_2024, yan_kdd_2020} on low-power IoT nodes can help reduce energy consumption while maintaining high performance.
    \item While GNNs excel at capturing relational structures, they may not be ideal for all aspects of IoT data processing. For instance, CNNs are highly effective at processing image data, while RNNs and transformers are well-suited for sequential and time-series data. By integrating GNNs with these models, hybrid architectures can be developed to leverage the strengths of each, enhancing overall performance in IoT applications. Such integrations enhance the overall accuracy and efficiency of IoT analytics. 
    \item Despite the growing interest in applying GNNs to IoT networks, there is a lack of widely accepted benchmarks tailored to this domain. Standardized datasets and evaluation protocols are crucial for comparing different GNN architectures, measuring their real-world effectiveness, and fostering innovation. Traditional ML models have well-established datasets, but GNNs in IoT lack such standardized resources. Creating open-source IoT-specific datasets incorporating factors such as device mobility, communication patterns, and heterogeneous data streams would enable more rigorous testing and validation of GNN-based solutions. 
    \item In conclusion, GNNs offer powerful tools for IoT applications, particularly in dealing with complex, interconnected systems. The future of GNNs in IoT lies in optimizing these models for practical, real-world deployment, ensuring they are both effective and efficient in a variety of IoT scenarios.
\end{enumerate}

\section{GNN for Spectrum Awareness}\label{sec:gnn_spec}

As the number of wireless devices continues to surge, the radio spectrum is becoming increasingly congested, leading to unwanted interference and a decline in Quality of Service (QoS). This challenge is particularly critical as modern communication systems must coexist within shared frequency bands, often competing with both licensed and unlicensed users. Without effective spectrum awareness, networks risk inefficient spectrum utilization and disruptive interference, ultimately degrading performance for essential applications. To tackle this challenge, advanced ML techniques have been leveraged to enhance spectrum awareness, enabling intelligent and adaptive spectrum management. GNNs have emerged as a powerful tool due to their unique ability to model complex, dynamic spectral environments. Unlike traditional ML approaches, GNNs can effectively capture spatial and relational dependencies among spectrum users, making them well-suited for tasks such as spectrum sensing, and adaptive signal classification. To this end, in this section, we shed light on the use of GNN for Radio Frequency Spectrum Sensing in Section~\ref{ssec:rfss} and Radio Frequency Signal Classification in Section~\ref{ssec:rfsc}. Table~\ref{tab:gnn_sa_table} provides a high-level comparison of surveyed paper using GNNs for spectrum awareness applications. We also provide the dataset or the simulator used for performance evaluation.

\begin{table*}[htbp]
\centering
\caption{Comparison of surveyed papers using GNN for Spectrum Awareness Applications.}
\renewcommand{\arraystretch}{2.2}
{\begin{tabular}{|m{2.3cm}|m{2.5cm}|m{3.5cm}|m{3.5cm}|m{3.5cm}|}
\hline

\textbf{Section} & \textbf{Paper} & \textbf{Application} & \textbf{ML Algorithm Used} &  \textbf{Dataset/ Simulator Used} \\
\hline\hline
\multirow{3}{=}{\textbf{Section \ref{ssec:rfss}}\\Radio Frequency Spectrum Sensing} 
& \textit{Zhang et al.} \cite{Zhang_twc_2022} &  Cooperative Spectrum Sensing & Inductive Graph Neural Network Kriging & SPLAT! Tool \cite{splat}, Terrain
database of US Geological Survey, Simulation data \cite{Chakraborty_infocom_2018}, Crawdad WiMax \cite{crawdad_wimax} \\
\cline{2-5}
& \textit{Janu et al.} \cite{Janu_tvt_2023} & Cooperative Spectrum Sensing & Graph Convolution Network & Python-based simulator \\
\cline{2-5}
& \textit{He et al.} \cite{Haibo_WC_2019} & Distributed
cooperative spectrum sensing & GNN + RL & Own simulator \\
\hline

\multirow{4}{=}{\textbf{Section \ref{ssec:rfsc}}\\Radio frequency signal classification} 
& \textit{Xuan et al.} \cite{Xuan_tnse_2022} & Modulation classification  & Adaptive Visibility Graph Neural Network & RML2016.10a \cite{shea_eann_2016}, RML2016.10b \cite{Timothy_2016_gnur} \\
\cline{2-5}
& \textit{Qiu et al.} \cite{Qiu_twc_2023} & Radio Signal Classification & RNN + CNN + GNN & RML2016.10a \cite{shea_eann_2016}; own dataset baaed on NVIDIA Tesla T4 simulator (SigData18 and SigData36)  \\
\cline{2-5}
& \textit{Bertalanič et al.} \cite{Bertalanič_wcnc_2023} & Wireless link layer anomaly classification & Graph Isomorphism Network & Rutgers WiFi dataset \cite{crawdad_rutgers} \\
\cline{2-5}
& \textit{Pang et al.} \cite{Pang_CGNN_2021} & Network traffic classification & Chained Graph Neural Network & Application traffic dataset, Malicious traffic dataset, Encrypted traffic dataset \cite{DraperGil2016CharacterizationOE} \\
\hline

\end{tabular}}
\label{tab:gnn_sa_table}
\end{table*}

\subsection{Radio Frequency Spectrum Sensing} \label{ssec:rfss}

\begin{figure*}[t!]
\centering
\includegraphics[width=0.75\textwidth]{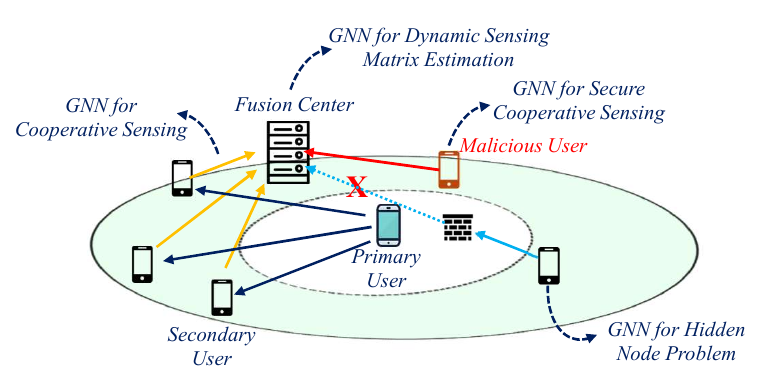}
\caption{Applications of GNN for RF spectrum sensing tasks.}
\label{fig:gnn_rf_spec1}
\end{figure*}

In the context of cognitive radios, spectrum sensing is performed to detect the presence of primary users (licensed users) in a given frequency band to enable opportunistic access for secondary users (unlicensed users). GNNs can analyze the received signals across different frequency bands and spatial locations, effectively capturing the spatial and spectral correlations among different spectrum channels. By learning the complex patterns in the received signals, GNNs can improve the accuracy of spectrum sensing, even in dynamic and heterogeneous environments. Figure~\ref{fig:gnn_rf_spec1} shows common applications of GNN for RF spectrum sensing tasks. In cooperative sensing scenarios \cite{Zhang_twc_2022}  \cite{Janu_tvt_2023} \cite{Haibo_WC_2019}, multiple sensors or nodes work together to gather and share information, improving the overall sensing performance beyond what individual sensors could achieve alone. This concept is prevalent in wireless communication networks, cognitive radio systems, distributed sensor networks, and other collaborative sensing applications. 

In \cite{Janu_tvt_2023}, the authors use a GCN-based cooperative spectrum sensing technique to address the hidden-node problem. The hidden node problem is primarily caused when a secondary user moves far away from the primary user or the fusion center thereby not participating in the decision-making process. Compared to traditional CNNs which assume the sensing matrices created by the fusion center as static but in reality, they are dynamically changing in the real environment. Thus needing a framework to feed dynamic graphs to the GCN for learning the features for accurate spectrum state classification. In \cite{Haibo_WC_2019}, the authors investigate the use of GNN in conjunction with reinforcement learning techniques to optimize the energy efficiency of cooperative spectrum sensing wireless communication systems. The GNN is used to generate a feature vector from a distributed sensing node using graph-structured data and Q-learning based reinforcement learning is used to find the optimal solution.

\begin{figure*}[t!]
\centering
\includegraphics[width=0.98\textwidth]{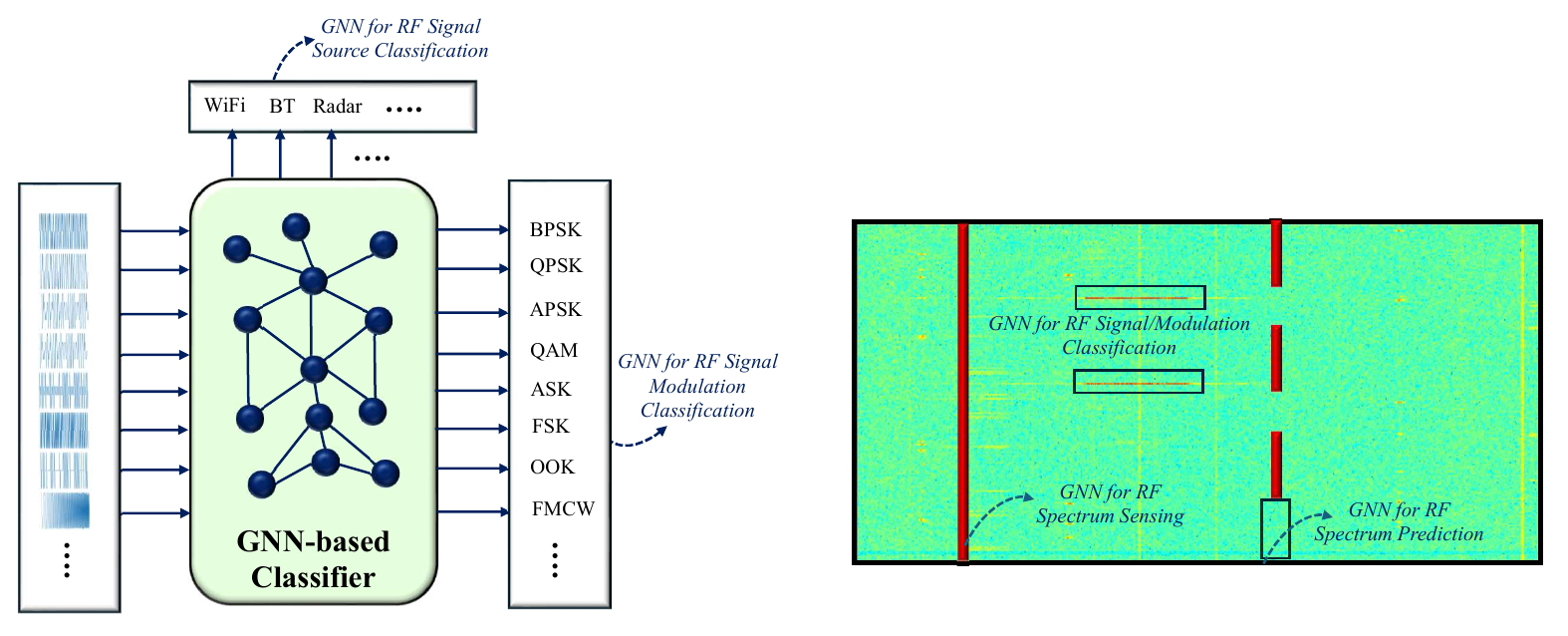}
\caption{Applications of GNN for RF signal classification and prediction tasks.}
\label{fig:gnn_rf_spec2}
\end{figure*}

\subsection{Radio Frequency Signal Classification}\label{ssec:rfsc}

While detecting the presence of active and inactive spectrum is an important problem, it is equally important to identify the type of signal being transmitted. This is especially important in applications that are critical and vulnerable to external interference. Furthermore, it is important to identify if the signals are friendly or adversary. This classification can usually be done based on the known signals that the incumbent system expects while any other unknown signals can be flagged as an adversary. ML is known for its classification and prediction capabilities across domains. Traditionally, CNNs \cite{AlexGoogleNet_AMC, AJagannath21ICC,oshea2018,Ajagannath2022SPIE, wirelesstech} and RNNs have dominated RF signal classification tasks such as modulation classification, signal type classification, RF Fingerprinting among others \cite{wirelesstech, Ajagannath2022ComST2022, AJagannathTCCN2023}.  More recently, GNNs are also being leveraged for these tasks including but not limited to \cite{Xuan_tnse_2022, Qiu_twc_2023}. Figure~\ref{fig:gnn_rf_spec2} shows common applications of GNN for RF signal classification and prediction tasks. Furthermore, GNN has also impacted other classification tasks such as anomaly classification \cite{Bertalanič_wcnc_2023} and traffic classification \cite{Pang_CGNN_2021}. 

AvgNet \cite{Xuan_tnse_2022} is an Adaptive Visibility Graph algorithm that maps time series data into graphs and can be used to classify various signal modulations. The main advantage of this algorithm is that it can automatically adapt into best-matching graphs, reducing the time consumed during the mapping process and aggregating both the local and global information of the signal sample. The proposed model is verified using simulation and the results show that AvgNet can effectively classify 11 modulations namely Amplitude Modulation - Double Sideband (AM-DSB), Amplitude Modulation - Single Sideband (AM-SSB), Phase Shift Keying (8PSK), Binary Phase Shift Keying (BPSK), Continuous Phase Frequency Shift Keying (CPFSK), Gaussian Frequency Shift Keying (GFSK), Pulse Amplitude Modulation (PAM4), Quadrature Amplitude Modulation (QAM16, QAM64), Quadrature Phase Shift Keying (QPSK) and Wide-band Frequency Modulation (WBFM). 

Another example of a signal modulation classification framework is DeepSIG \cite{Qiu_twc_2023} which is a hybrid heterogeneous modulation classification architecture designed based on the combination of Recurrent Neural Network (RNN), CNN and GNN. This framework can process radio
signals with heterogeneous input formats such as in-phase (I) and quadrature (Q) samples, images mapped from IQ samples, and graphs converted from IQ samples. In addition to classifying the signals listed for \cite{Xuan_tnse_2022}, DeepSIG can also classify Minimum Shift Keying (MSK), Gaussian Minimum Shift Keying (GMSK), Orthogonal Frequency Division Multiplexing (OFDM), OFDM Quad-Carrier Quadrature Phase Shift Keying (OFDM-QPSK), Offset Quadrature Phase Shift Keying (OQPSK), OFDM Binary PSK (OFDM-BPSK), OFDM Quadrature Amplitude Modulation (OFDM-QAM16), Amplitude Shift Keying (ASK), Amplitude and PSK (APSK), Amplitude Modulation (AM), Frequency Modulation (FM), AM-MSK, FM-MSK, Pulse Amplitude Modulation (PAM), Continuous-Phase Modulation (CPM) and On-off key (OOK).

In addition to classifying the signal modulation, it is also important to detect and correct link failures and abnormal network behavior that may affect the large wireless network infrastructure. To address this problem, in \cite{Bertalanič_wcnc_2023}, the authors present a new method that can detect wireless link anomalies based on GNN. Specifically, in this work, the authors use GIN architecture. The motivation for using GIN is due to the fact that the graphs of similar-looking anomalies have similar nodes and edges making these graphs isomorphic in nature. The effectiveness of the algorithm was verified using a real-world WiFi testbed.

GNN also finds its application for traffic classification
scenarios where it is important to differentiate between traffic generated by various applications while identifying any encrypted or malicious traffic. To enable this, in \cite{Pang_CGNN_2021}, the authors present CGNN, a Chained GNN to effectively classify various real-world traffic scenarios and categorize them into the different applications as well as label them as ``normal", ``encrypted", or ``malicious" traffic. Through extensive simulation, it is shown that compared to CNN and other state-of-the-art techniques, GNN, specifically CGNN outperforms them in terms of three key metrics namely \textit{precision, accuracy, and recall}. 

\subsection{Lessons Learned}
In this section, we present the key takeaways and lessons learned from the applications of GNN for Spectrum Awareness discussed in Sections~\ref{ssec:rfss} and \ref{ssec:rfsc}. 
\begin{enumerate}
    \item Wireless spectrum environments are inherently complex, featuring multiple users, heterogeneous devices, and dynamic frequency allocations. The interactions between these elements form a structured graph, where nodes represent transmitters, receivers, or frequency bands, and edges capture interference, connectivity, or usage relationships. GNNs excel at learning from graph-structured data, making them well-suited for modeling spectrum environments where traditional machine learning models struggle to account for the intricate dependencies among users and devices. By leveraging message-passing mechanisms, GNNs can infer spectrum availability, predict interference patterns, and optimize spectrum allocation more effectively than conventional approaches.
    \item Spectrum occupancy is rarely random; it follows spatiotemporal patterns. Devices in close proximity tend to experience similar spectrum conditions, and usage trends evolve over time due to periodic activities (e.g., office hours, peak traffic times). Traditional models often fail to capture these correlations, leading to suboptimal spectrum predictions. GNNs, however, naturally encode spatial and temporal dependencies by aggregating information from neighboring nodes over multiple graph layers. This enables them to recognize patterns such as congestion hotspots, propagation effects, and historical usage trends, improving spectrum awareness and prediction accuracy.
    \item Wireless networks are in constant flux, with devices frequently joining and leaving the network, frequency bands becoming occupied or vacated, and interference levels fluctuating. Static models struggle to adapt to such real-time changes, whereas GNNs, when combined with recurrent or adaptive learning techniques, can dynamically update their representations based on evolving graph structures.
    \item Training machine learning models for spectrum awareness often requires labeled datasets, where spectrum occupancy, interference events, and device activities are annotated. However, obtaining labeled data in wireless environments is challenging due to privacy concerns where spectrum usage data may contain sensitive information related to user behavior and communications as well as high data collection costs where Deploying sensors and monitoring devices to label spectrum states is expensive and resource-intensive. GNNs can mitigate data scarcity issues through self-supervised and semi-supervised learning techniques, where they learn from partially labeled graphs or leverage domain knowledge to enhance model generalization. Transfer learning \cite{Suetrong_acc_2024, dai_tkde_2023} can also be employed, allowing models trained in one spectrum environment to be adapted to another with minimal labeled data.
    \item Interference is a major issue in spectrum management, as unauthorized transmissions, overlapping frequency bands, and external noise can degrade network performance. Traditional interference detection methods rely on threshold-based techniques or statistical models, which often fail to capture complex interference patterns arising from diverse sources such as co-channel interference, adjacent-channel leakage, and unintentional electromagnetic interference. GNNs can identify non-trivial interference patterns that traditional models miss. By learning from historical spectrum data and network topology, GNNs can predict interference hotspots, classify interference types, and suggest mitigation strategies such as frequency reallocation or power adjustments.
    \item In conclusion, GNNs have shown great potential in improving spectrum awareness, particularly in handling the complex, dynamic, and interconnected nature of wireless networks. As the field evolves, we can expect GNNs to play a critical role in optimizing spectrum use, especially in increasingly crowded and diverse wireless environments.
\end{enumerate} 

\begin{table*}[htbp]
\renewcommand{\arraystretch}{1.3}
\centering
\caption{Comparison of surveyed papers using GNN for Networking Applications.}
{\begin{tabular}{|m{2.3cm}|m{2.5cm}|m{3.5cm}|m{3.5cm}|m{3.5cm}|}
\hline

\textbf{Section} & \textbf{Paper} & \textbf{Application} & \textbf{ML Algorithm Used} &  \textbf{Dataset Used} \\
\hline
\hline

\multirow{5}{=}{\textbf{Section \ref{ssec:pred}}\\Network and Signal Characteristics Prediction} 
& \textit{Soto et al.} \cite{soto_sensor_2021} & WLAN performance prediction & Graph Convolutional Network & ITU-T AI Challenge \cite{Francesc_dataset}, Komondor simulator \cite{Barrachina_wd_2019} \\
\cline{2-5}
& \textit{Zhou et al.} \cite{Hongkuan_infocom_2023} & WLAN performance prediction & Dynamic Temporal Graph Neural Network & Komondor simulator \cite{Barrachina_wd_2019}\\
\cline{2-5}
& \textit{Rattaro et al.} \cite{Rattaro_urucon_2021} & Wireless RSSI prediction & Random Dot Product
Graphs + GNN & Plan Ceibal \cite{plan_ceibal} \\
\cline{2-5}
& \textit{Sun et al.} \cite{Sun_tmc_2022} & Mobile data traffic prediction  & Graph-based
Temporal Convolutional Network & Own Wi-Fi dataset \\
\cline{2-5}
& \textit{Sun et al.} \cite{sun_cc_2022} & Channel prediction & GNN & WirelessInSite \cite{wirelessinsite}, Blender \cite{blender} \\
\cline{2-5}
& \textit{Liu et al.} \cite{Xiaoyang_link_capsnet_2023} & Link prediction & Graph Capsule Network & Six real networks and three citation networks  \\
\hline

\multirow{7}{=}{\textbf{Section \ref{ssec:rout}}\\Routing Optimization} 
& \textit{Yang et al.} \cite{yang_iotj_2022} & Joint routing and scheduling optimization & GCN + RL & NetworkX \cite{NetworkX} \\
\cline{2-5}
& \textit{Lu et al.} \cite{lu_sensor_2023} & Multi-path routing & GraphSAGE & NS-3 simulator \cite{ns3_sim} \\
\cline{2-5}
& \textit{Rusek et al.} \cite{Rusek_jsac_2020} & Network modeling and optimization & GNN & OMNET ++ simulator \cite{varga2001discrete}\\
\cline{2-5}
& \textit{Almasan et al.} \cite{Paul_CC_2022}  & Routing optimization  & GNN + RL & OpenAI Gym \cite{opeai_gym} \\
\cline{2-5}
& \textit{Xu et al.} \cite{xu_cbd_2022} & Wireless networking routing optimization & GNN + RL  & Own python-based simulator \\
\cline{2-5}
& \textit{Swaminathan et al.} \cite{Avinash_cc_2021} & Routing optimization  & GNN + RL & Own simulator \\
\cline{2-5}
& \textit{Yan et al.} \cite{Binghao_fgcs_2022} & Multi-path routing & GNN & Own simulator \\
\hline

\multirow{3}{=}{\textbf{Section \ref{ssec:cong}}\\Congestion Control} 
& \textit{Kirby et al.} \cite{Kirby_CongestionNet_2019} & Routing congestion prediction & Graph Attention Networks & Own python-based simulator \\
\cline{2-5}
& \textit{LaMar et al.} \cite{LaMar_CONAIR_2021} & Intent-based routing  & GNN & EXata Network Emulator Software \cite{exata} \\
\cline{2-5}
& \textit{Ferriol-Galmés et al.} \cite{Ferriol_RouteNet_2023} & Network modeling & GNN & OMNET-based simulator \cite{varga2001discrete}, Routenet-Fermi \cite{routenet_fermi} \\
\hline

\multirow{8}{=}{\textbf{Section \ref{ssec:mec}}\\Mobile Edge Computing} 
& \textit{Asheralieva et al.} \cite{Asheralieva_tmc_2023}  & Multi-access
edge computing & GNN + RL & OPNET-based simulator \cite{opnet} \\
\cline{2-5}
& \textit{Zeng et al.} \cite{Zeng_2023_jsac}  & Distributed edge servers processing & GNN & SIoT \cite{Claudio_siot}, YelpChi \cite{yelp} \\
\cline{2-5}
& \textit{Zhou et al.} \cite{Zhou_2023_dac}& Network automated design & GNN & Own python-based simulator \\
\cline{2-5}
& \textit{Sun et al.} \cite{sun_iotj_2023}  & Task offloading & GNN + RL & Own simulator \\
\cline{2-5}
& \textit{Li et al.} \cite{li_tnsm_2023}  & Edge computing, task offloading  & GNN + RL & Own simulator \\
\cline{2-5}
& \textit{Ma et al.} \cite{ma_icc_mogr_2024} & Multi-task Offloading & Graph Convolutional Network & Own simulator \\
\cline{2-5}
& \textit{Zhang et al.} \cite{Zhang_2022_icc}  & Service offloading & GNN + RL & Own python-based simulator \\
\cline{2-5}
& \textit{Yang et al.} \cite{Shu_cc_2023} & Resource scheduling and Task offloading & Graph Convolutional Network & Own simulator \\
\hline

\multirow{7}{=}{\textbf{Section \ref{ssec:dt}}\\Digital Twin Networks} 
& \textit{Wang et al.} \cite{wang_tii_2022} & Network slicing management & GNN & NSFNET and GEANT2 \cite{geant2_nsfnet_data} \\
\cline{2-5}
& \textit{Naeem et al.} \cite{Naeem_gcw_2023} & Network Slicing & GNN + RL & Own simulator \\
\cline{2-5}
& \textit{Ferriol-Galmés et al.} \cite{Miquel_ecn_2022}  & Network optimization & GNN & TwinNet \cite{Miquel_github} \\
\cline{2-5}
& \textit{Ferriol-Galmés et al.} \cite{Miquel_icassp_2022}& Network modeling & GNN &  OMNet++ Simulator \cite{varga2001discrete} \\
\cline{2-5}
& \textit{Yu et al.} \cite{yu_jsac_2023} & Self-healing mechanism in 6G edge networks. & GNN & OMNet++ based modeling dataset \cite{varga2001discrete} \\
\cline{2-5}
& \textit{Zhang et al.} \cite{zhang_jsac_2023} & Power allocation and user association & GNN & Own simulator \\
\hline

\multirow{5}{=}{\textbf{Section \ref{ssec:uavn}}\\Unmanned Aerial Networks} 
& \textit{Liu et al.} \cite{Liu_iotj_2021} & Air Quality Sensing Framework & GCN + LSTM & Ground and Aerial sensing data \cite{aqi_data} \\
\cline{2-5}
& \textit{Zhang et al.} \cite{Zhang_twc_2023} & Cooperative trajectory design & Graph Convolutional Neural Network + RL & Python-based simulator \cite{an_github} \\
\cline{2-5}
& \textit{An et al.} \cite{An_access_2024} & UAV Trajectory Prediction & Dynamic GNN & Python-based simulator \\
\cline{2-5}
& \textit{Mou et al.} \cite{Mou_jsac_2022} &  Self-healing UAV swarm network & Graph Convolutional Network & Python-based simulator \cite{mou_github} \\
\cline{2-5}
& \textit{Wang et al.} \cite{Wang_remsen_2022} &  Joint flying relay location and routing optimization & GNN + LSTM & Own simulator \\
\hline

\end{tabular}}
\label{tab:gnn_ntwk_table}
\end{table*}

\section{GNN for Networking}\label{sec:gnn_net}

With the evolving number of wireless devices, the research community is working towards finding ways to improve the network performance to meet the quality of service requirements of the users. Many ML-based solutions have been proposed in an effort to achieve the goal of better network performance. GNNs have gained significant attention due to their ability to model complex relationships in network structures and has been successfully applied to various networking challenges, offering promising improvements in performance and decision-making capabilities. In this section we aim to provide a comprehensive overview of how GNNs are shaping the future of networking to create more intelligent, efficient, and adaptive networking infrastructures. To this end, we will focus on applications of GNN for various networking challenges such as Network and Signal Characteristics Prediction (Section~\ref{ssec:pred}), Routing Optimization (Section~\ref{ssec:rout}), Congestion Control (Section~\ref{ssec:cong}). We will discuss the application of GNN for Mobile Edge Computing (Section~\ref{ssec:mec}). Within the networking domain, there are two rapidly growing fields of research namely Digital Twin Networks and Unmanned Aerial Networks. In this section, we will discuss the application of GNN for Digital Twin Networks in Section~\ref{ssec:dt} and Unmanned Aerial Networks in Section~\ref{ssec:uavn}. Table~\ref{tab:gnn_ntwk_table} provides a high-level comparison of surveyed paper using GNNs for wireless networking applications. We also provide the dataset or the simulator used for performance evaluation.

\subsection{Network and Signal Characteristics Prediction}\label{ssec:pred}

Prediction applications include but not limited to Wireless LAN (WLAN) performance prediction \cite{soto_sensor_2021, Hongkuan_infocom_2023}, Received Signal Strength Indicator (RSSI) prediction \cite{Rattaro_urucon_2021}, network traffic prediction \cite{Sun_tmc_2022}, channel prediction \cite{sun_cc_2022} and link prediction \cite{Xiaoyang_link_capsnet_2023}. In \cite{soto_sensor_2021}, the authors present a data-driven approach using GCN to predict the performance of next-generation WiFi networks. The designed model can predict the throughput of densely deployed WiFi nodes using Channel Bonding (CB). CB is a technique where two or more channels are combined thereby increasing the throughput between multiple wireless devices. In \cite{Hongkuan_infocom_2023}, the authors present HTNet, a specialized Heterogeneous temporal GNN that can extract features from dynamic WLAN deployment to predict its throughput. Through analysis of the unique graph structure of WLAN deployment graphs, HTNets can achieve the lowest prediction error in real-world experiments. 

In \cite{Rattaro_urucon_2021} the authors use a combination of Random Dot Product Graphs (RDPG) and GNNs to address the problem of predicting wireless received signal strength (RSS) of wireless networks. RDPG is a spectral-based embedding method that provides a flexible framework for generating random graphs with various structural properties. Finally, in \cite{Sun_tmc_2022}  the authors use a time-evolving graph to formulate the time-evolving nature of the user movements and propose a Graph-based Temporal Convolutional Network (GTCN) to predict the future traffic of each node in the wireless network. The GTCN can effectively learn intra- and inter-time spatial correlation between each node through the node aggregation method as well as can efficiently model the temporal dynamics of mobile traffic trends from different nodes through the temporal convolutional layer. In \cite{sun_cc_2022}, the authors present an environment-information based channel prediction method to effectively predict the large-scale and small-scale fading parameters using GNN. First, scatterer-centered communication environment graphs are used to classify the scattering sources affecting the channel characteristics into effective, primary, and non-primary scatterers. The scatterers that produce multi-path and influence the channel characteristics are referred to as effective scatterers while the scatters that produce single propagation points are referred to as primary scatterers and all other scatterers are non-primary scatterers. After successful classification, the GNN-based channel prediction module accomplishes the task of predicting the large-scale and small-scale fading parameters using fully-connected propagation environment graphs and single-connected propagation environment graphs. Furthermore, a GNN-based multi-target regression model is designed for continuous channel parameters prediction. In \cite{Xiaoyang_link_capsnet_2023}, the authors propose a link prediction approach using GCapsNet called Graph Conversion Capsule Link (GCCL) which is capable of operating on graphs with and without node features. This is achieved by first using GNN to generate the node embeddings and then transforming the node embedding to a node pair feature map. Then CapsNet is applied to learn the feature representation of the feature map. Through this transformation, the link prediction problem is similar to a binary classification problem where the two decisions are either the two nodes are likely to have an edge in the near future or not.

\subsection{Routing Optimization}\label{ssec:rout}

In the previous section we discussed the applications of GNN for network and signal characteristics prediction. The next step is to use the predicted characteristics and decide on the optimal routing path to further improve the network performance. To this end, in this section, we will briefly introduce the process of routing and then delve into the applications of GNN for routing optimization. There are various routing protocols designed to facilitate the exchange of routing information among routers and enable them to make informed decisions about packet forwarding. The objective of routing protocols is to determine the best path or route for data packets from a source to a destination in a network based on predefined performance metrics (e.g. minimizing latency or minimizing energy consumption). Thus it is important to have an optimized routing algorithm that can improve \textit{efficiency} by selecting the best routes, thereby reducing network congestion and improving overall network performance, resulting in better \textit{Resource Utilization}. 

Traditionally, techniques such as reinforcement learning, and recurrent neural networks were commonly used for routing optimization. Recently researchers have studied the applicability of GNN for routing optimization due to its ability to model and learn from graph-structured data. The key advantage of GNN compared to other methods is that by representing the network topology as a graph, where nodes represent routers or switches and edges represent links, GNNs can learn to predict the best paths for forwarding packets based on factors such as link capacities, congestion levels, latency and much more. The application of GNN for routing optimization has gained significant traction in the research community \cite{yang_iotj_2022, lu_sensor_2023,Rusek_jsac_2020,Paul_CC_2022,xu_cbd_2022,Avinash_cc_2021,Binghao_fgcs_2022, hu_ieeetnnls_july_2025}. 

\begin{figure*}[t]
\centering
\includegraphics[width=0.9\textwidth]{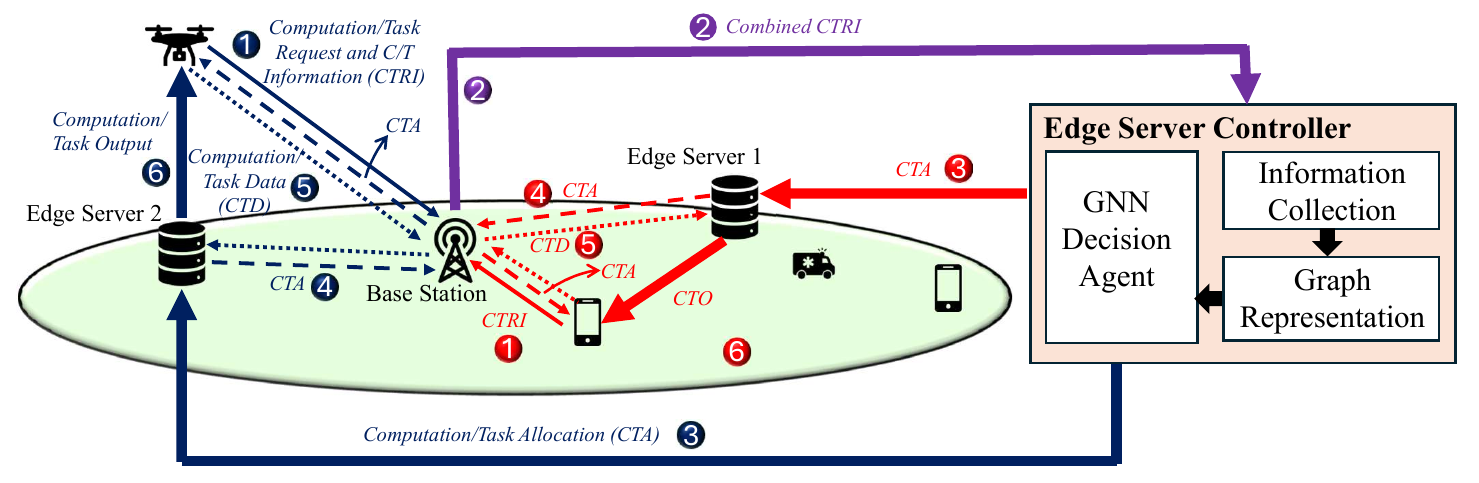}
\caption{Workflow Diagram of GNN-enabled Mobile Edge Computing.}
\label{fig:gnn_mec}
\end{figure*}

For example, in \cite{yang_iotj_2022}, the authors use a GCN-based DRL algorithm to jointly optimize traffic scheduling and routing. The key advantage of this algorithm is that there is no prior knowledge of the system, and all the information is obtained in run-time by interacting with the environment. In \cite{lu_sensor_2023}, they propose a reliable routing optimization algorithm using GraphSAGE. The algorithm consists of two components. First, a multi-path routing algorithm using breadth-first search method that can continuously predict the shortest path for different network topologies. Second is GraphSAGE-based routing performance optimization algorithm that creates network labels for training GraphSAGE and is used to learn graph features to predict the network performance for each path and select the best path for optimal network performance. This proposed method can minimize network congestion, improve network throughput, and reduce end-to-end delay and packet loss rate compared to traditional shortest-path routing algorithms.

RouteNet \cite{Rusek_jsac_2020} is a novel network model-based GNN that can understand the complex relationships between topology, routing and input traffic and accurately predict the estimate of per-source/destination per-packet delay and packet loss. RouteNet can be generalized to arbitrary topologies, routing schemes, and traffic intensity thanks to GNN's ability to learn and model graph-structured information. In \cite{Paul_CC_2022} the authors explore the advantage of integrating GNN with Deep Reinforcement learning (DRL) to overcome the challenges of using DRL which is not well suited to learning from information structured as graphs and the limited generalizability of DRL. The proposed GNN-based DRL algorithm can learn and generalize to any given network topology. The proposed algorithm not only works for scenarios seen during the training process but also for network topologies that were not part of the training set. Similarly, in \cite{xu_cbd_2022}, the authors propose GRL-NET, a combination of GNN and DRL-based intelligent routing algorithm. Compared to using conventional methods to construct DRL agent, in this work, GNN is used due to its ability to learn from complex relationships among network topologies.

In \cite{Avinash_cc_2021}, the authors present a GNN-based routing algorithm that leverages global information from a software-defined network to accurately predict the optimal path with the minimum average delay between the source and the destination node. In this work, GNN is primarily used because of its ability to generalize to different network topologies and varying data traffic scenarios. In \cite{Binghao_fgcs_2022}, the authors propose a multi-path routing algorithm based on GNN to improve end-to-end transmission efficiency. First, a flow splitting scheme is designed to achieve multi-path transmission, and second, GNN is used to predict the link delay and select the most-efficient efficient path. The proposed algorithm outperforms the existing solutions in terms of time overhead, end-to-end delay, throughput, and generalizability to different topologies.

\subsection{Congestion Control} \label{ssec:cong}

While routing optimization tries to improve the network performance, as more and more devices join the network and all devices request data with specific QoS requirements, it is important to serve all the users simultaneously which could cause congestion in the network. Even though routing optimization finds the optimal path, it does not necessarily control the flow of data within the route. Therefore it is important to control the data flow which is called congestion control. In this section, we first introduce congestion control and then look into the application of GNN to address this challenge. Congestion control refers to the techniques and mechanisms used in computer networks to regulate the flow of data packets, preventing network congestion, and ensuring efficient data transmission. When network resources such as bandwidth, buffers, or processing capacity are overwhelmed with data, congestion occurs, leading to packet loss, increased latency, and reduced throughput. Congestion control aims to manage this situation to maintain network stability and performance. While traditional ML algorithms are extensively used for congestion control, now the research community is investigating the use of GNN to efficiently control the congestion in small and large-scale networks. Some of the research work includes CongestionNet \cite{Kirby_CongestionNet_2019}, CONAIR \cite{LaMar_CONAIR_2021} and RouteNet-Fermi \cite{Ferriol_RouteNet_2023}. 

In \cite{Kirby_CongestionNet_2019}, the authors present CongestionNet, a graph-based deep learning method using GAT to quickly predict routing congestion hotspots. GAT was used because the GAT model allows the network to make higher frequency decisions compared to other methods and also performs better on inductive inference problems where there are no labels during the inference time. CONAIR \cite{LaMar_CONAIR_2021} is a  proactive Congestion Aware Intent-Based
Routing architecture that allows selection of best communication link based on quality of service (QoS) using GNN. GNN allows the prediction of the future state of the network based on QoS metrics at each input node within the network. Furthermore, this approach can be applied to dynamic networking topologies with models trained on different scenarios. In \cite{Ferriol_RouteNet_2023}, the authors introduce RouteNet-Fermi, a custom GNN model that can be used to accurately predict the delay, jitter, and packet loss in the network. One key advantage of this method is that it can also provide accurate estimates in networks that were not in the training set.

\subsection{Mobile Edge Computing}\label{ssec:mec}
In the previous section we mentioned that devices are constantly gathering information and need to process them to extract the relevant information and take necessary actions to achieve its goal. While large capable types of machinery can process huge amounts of data, machinery that are battery-constrained, low-power with limited computation and storage devices cannot effectively process and store the data. This raises the need for an effective computing and storage mechanism that can be leveraged by such devices. One of the common solutions to this challenge is called Mobile Edge Computing. In this section, we will briefly introduce mobile edge computing before discussing how GNN is applied in various applications. Mobile Edge Computing (MEC) is a decentralized computing paradigm that brings computation and data storage closer to the location where it is needed, rather than relying solely on centralized cloud data centers. In edge computing, data is processed and analyzed locally on edge devices or servers, often at or near the data source, such as IoT devices, sensors, and other endpoints. Graphs are commonly used to represent the network topology and find its applications in Multi-Edge Cooperative Computing \cite{Yujiao_corais_2024} and Edge Cloud system for prioritized services  \cite{Yuanming_edgematrix_2022}, to name a few where edge computing and GNNs can be combined to create intelligent, real-time decision-making systems at the edge of the network. This combination is particularly powerful for applications that involve analyzing data from a network of interconnected devices or sensors, where GNNs can capture complex relationships and patterns in the data. Figure~\ref{fig:gnn_mec} shows the workflow of GNN-enabled MEC. The workflow can be divided into six steps. First, the user equipment (such as a cellular device or cellular-enabled UAV) sends the computation/task request and relevant information (CTRI) to the base station. Second, the base stations aggregate all the task information and QoS requirements and forwards the combined CTRI to the Edge Server Controller (ESC). Third, The ESC collects the information, generates the graph representation and the GNN decision agent (usually coupled with an RL) decides the best edge server capable of executing the task that satisfies the user QoS and sends a computation/task allocation (CTA) message to the edge server. Fourth, The BS then relays this CTA message to the user. Fifth, the user sends the task data to the edge server through the BS. Sixth, the edge server executes the task and sends the computation/task output (CTO) back to the user. Each of these steps is clearly labeled with numbered blue circles for cellular-enabled UAVs, numbered red circles for cellular users, and numbered purple circles for the combined messages.

In general, MEC can be divided into two parts. The first is Secure and Cost-efficient Computation \cite{Asheralieva_tmc_2023, Zeng_2023_jsac,Zhou_2023_dac} and the second is task offloading \cite{sun_iotj_2023,li_tnsm_2023,ma_icc_mogr_2024,Zhang_2022_icc,Shu_cc_2023}. In Section~\ref{ssec:iotdatafusion}, we discussed the use of data from different sensors for prediction tasks. These data are typically processed in a centralized server or a group of edge-servers. The final data obtained may be malicious when the server is under attack by an adversary, or the edge server may be malfunctioning due to other factors. In order to correctly recover the final output in either of these situations, it is important to introduce additional security measures. In \cite{Asheralieva_tmc_2023}, the authors propose the use of a generalized graph NN that would help identify the edge server types and categorize them into faithful and malicious. This approach has lower complexity and faster convergence than conventional neural networks. In \cite{Zeng_2023_jsac}, the authors build a novel modeling framework for optimizing system cost based on distributed GNN processing over heterogeneous edge servers. They also developed an incremental graph layout strategy to address the dynamic evolution of GNN's input data graph and an adaptive scheduling algorithm to balance the tradeoff between overhead and system performance. HGNAS  \cite{Zhou_2023_dac}, is the first Hardware-aware Graph Neural Architecture Search framework for resource-constrained edge devices. The key feature of HGNAS is that it can automatically search for optimal GNN architectures that maximize given performance metrics such as task accuracy or computation efficiency. 

Task offloading in MEC refers to the process of transferring computational tasks from mobile devices (such as smartphones, tablets, or IoT devices) to nearby edge servers or nodes located at the edge of the network infrastructure. In traditional cloud computing models, all computation is typically performed in centralized data centers, which can result in high latency and network congestion, especially for latency-sensitive applications running on mobile devices. Task offloading in MEC aims to alleviate these issues by bringing computation closer to the end-users, reducing latency, and improving the overall quality of service. GNN and Task offloading have been extensively studied and are applied to various applications \cite{sun_iotj_2023,li_tnsm_2023,ma_icc_mogr_2024,Zhang_2022_icc,Shu_cc_2023}. For example, in \cite{sun_iotj_2023} the authors propose a task offloading mechanism based on graph
neural network called graph reinforcement learning-based offloading (GRLO) framework is paired with an actor-critic network to overcome the challenges of Heuristic algorithms and deep reinforcement learning-based approaches that heavily depend on accurate mathematical models for the MEC system, and deep reinforcement learning techniques does not make fair use of the relationship between devices in the network. Similarly, in \cite{li_tnsm_2023}, the authors propose a meta-reinforcement learning task offloading algorithm called GASTO to overcome the challenges associated with the generalization and robustness of offloading algorithms in dynamic and uncertainties in the real-world environment. In \cite{ma_icc_mogr_2024}, the authors propose a task offloading mechanism based on GCN which generates the graph and analyzes the integrated relationship from features such as task characteristics, network conditions, and available resources at the edge. GCN enables improved decision-making, enhanced resource utilization, and optimized performance in the edge network.
In \cite{Zhang_2022_icc} the authors present an end-to-end fine-grained computing offloading model with the goal of achieving better load balancing. To this end, a deep graph matching method based on GNN is used to select the best node for offloading and can be applied for dynamic and large-scale networks and achieve faster execution speed. In comparison to existing methods for task offloading and load balancing, the proposed method can reduce the network load imbalance efficiently and due to its shorter execution time, the proposed scheme is suitable for real-time scenarios with multiple sub-tasks and edge nodes. Similarly, in \cite{Shu_cc_2023}, the authors propose a GCN-RL based algorithm for Container Scheduling for NextG IoT user task offloading and resource scheduling architecture with the objective of minimizing average task completion time for fine-grained and dynamic container scheduling problems.

\begin{figure*}
\centering
\includegraphics[width=0.8\textwidth]{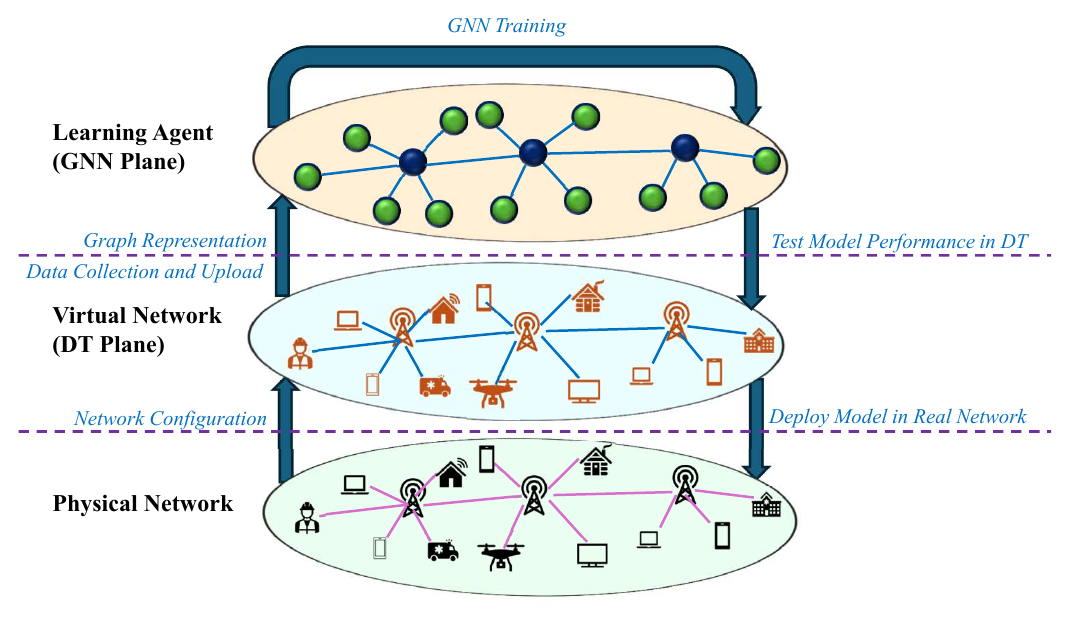}
\caption{Workflow diagram of GNN-enabled DT networks.}
\label{fig:gnn_dt}
\end{figure*}

\subsection{Digital Twin Networks} \label{ssec:dt}
In prior sections we discussed the common challenges and applications that used GNN that are not necessarily exclusive to NextG networks, but they are still important for any wireless networks. To shed some light on some NextG applications and the applications of GNN, in this section, we discuss one of the exciting and extensively researched topics by the community called Digital Twins. Digital Twins (DT) is considered one of the most promising enabling techniques for future IoT and next-generation networks \cite{JagannathWiseML22_DT}. Digital Twins are a virtual replica of the real world which enables researchers to create an accurate model of the real-world environment to test and improve their ML algorithms before integrating with real-world hardware. Figure~\ref{fig:gnn_dt} shows an example workflow diagram of the GNN-enabled DT network.

Over the past few years, DT has gained significant interest from the research community spanning across different domains. GNN is envisioned to be a powerful framework for analyzing and extracting information from complex network data \cite{Isah_icaiic_2024,Perdomo_iccw_2023}. In addition to traditional optimization and ML algorithms, GNN has been extensively used together with DT in applications such as network optimization \cite{wang_tii_2022, Naeem_gcw_2023, Miquel_ecn_2022, Miquel_icassp_2022}, 6G Edge Networks \cite{yu_jsac_2023} and resource allocation \cite{zhang_jsac_2023}.

For example, in \cite{wang_tii_2022}, the authors present a scalable DT of 5G network to model the interactions between multiple network slices into a graph and use a GNN-based virtual representation model that captures the hidden inter-dependencies among nodes based on the aggregation of information from multiple neighbors. Here, the DT is used to explore the complex traffic generated by the network slices and accurately predict the end-to-end slice latency. The DT designed based on GNN is shown to have higher accuracy in estimating the end-to-end latency for different topologies, traffic distributions, and QoS constraints that were not part of the training process. Similarly, in \cite{Naeem_gcw_2023}, the authors propose a DT-enabled deep distributed Q-network (DDQN) framework that constructs a digital replica of physical slicing-enabled beyond 5G networks that can simulate complex environment and predict its dynamic characteristics. Here, the DT is represented using graph and GNN to learn the complicated relationships of the network slice and the network states are forwarded to the DDQN agent to learn the optimal network slicing policy. It is shown that the proposed DT model achieves optimal and autonomous network management of complex slicing-enabled beyond 5G networks. 

In \cite{Miquel_ecn_2022}, the authors propose TwinNet, a GNN-based DT model that can take into account the complex relationship between different service level agreements metrics such as queuing policy, network topology, routing configuration, and input traffic. It is shown that GNN-enabled TwinNet can generalize to its input parameters, topologies, routing and queuing configuration, and can accurately predict the end-to-end path delays in 106 unseen real-world topologies from Internet Topology Zoo \cite{int_top_zoo}. Flow-aware Digital Twin (FlowDT) \cite{Miquel_icassp_2022}, is a GNN-based framework to model computer networks, where \textit{flow} refers to the set of packets that share common characteristics such as the same protocols (UDP/TCP), IP addresses or port numbers of source and destination node. FlowDT supports the creation and destruction of flows and also provides accurate estimates of per-flow metrics such as delay and jitter. 

In \cite{yu_jsac_2023}, the authors analyze the self-healing mechanism in 6G edge networks using DT. Self-healing 6G networks refer to the next-generation of wireless communication infrastructure that possesses advanced capabilities to automatically detect, diagnose, and mitigate issues or disruptions within the network without human intervention. These networks are designed to be highly resilient and reliable, capable of maintaining connectivity even in challenging conditions or in the face of malicious attacks. GNN is used to design a performance prediction mechanism that can accurately predict the network performance and detect any abnormal links or nodes. In \cite{zhang_jsac_2023}, the authors present a DT architecture for Terahertz (THz) wireless network and represent the DT network as a graph. The generated graph is used in a GNN-based multi-set distributed message propagation algorithm to solve power allocation and user association for THz networks.

\begin{figure*}[t]
\centering
\includegraphics[width=0.8\textwidth]{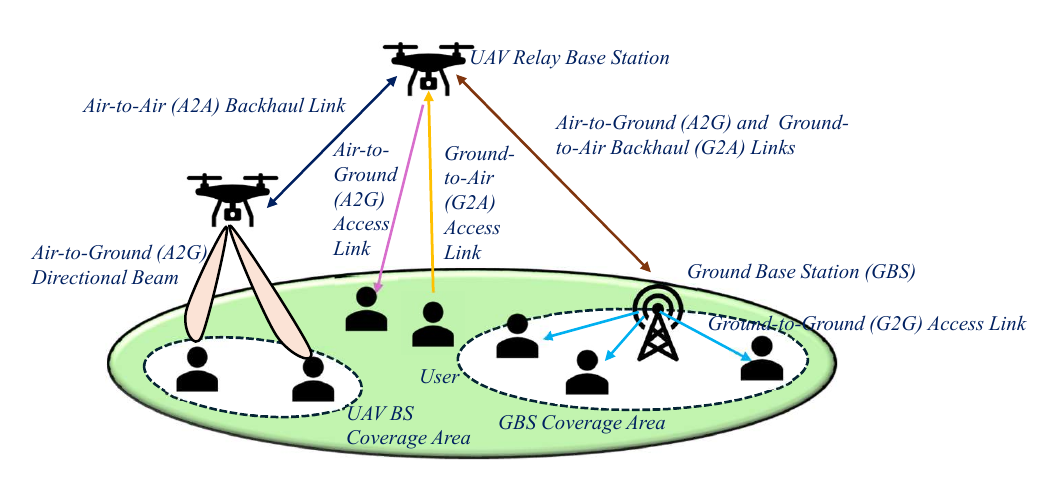}
\caption{Typical components of UANs.}
\label{fig:uan1}
\end{figure*}

\begin{figure*}
\centering
\includegraphics[width=0.98\textwidth]{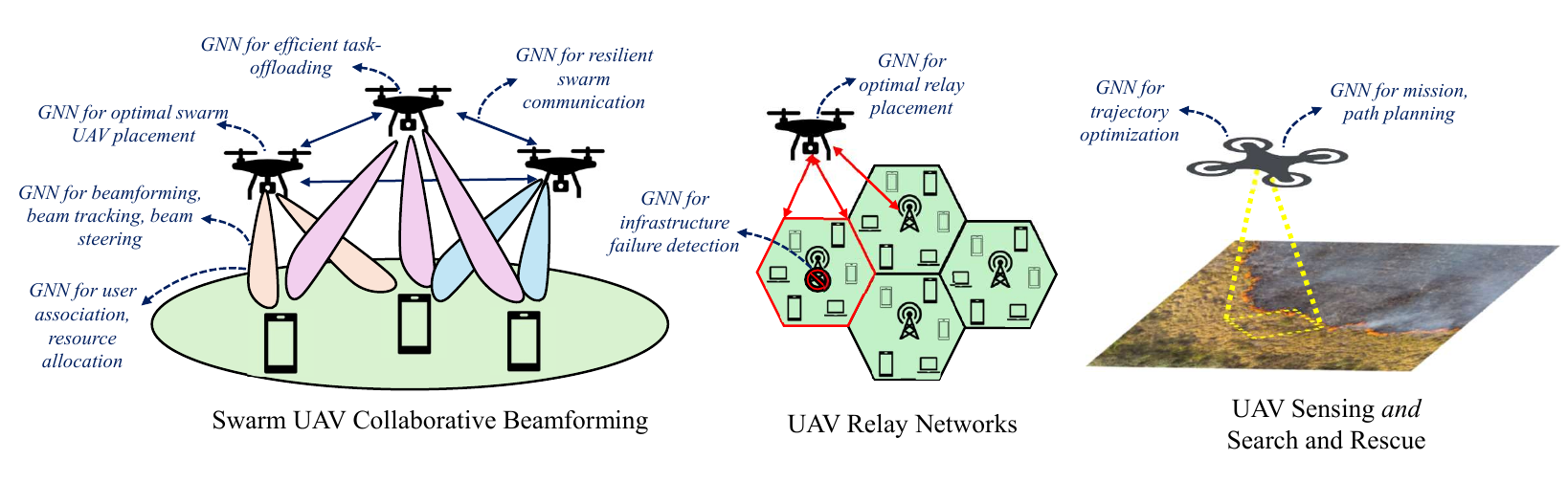}
\caption{Applications of GNN for UAN tasks.}
\label{fig:uan2}
\end{figure*}

\subsection{Unmanned Aerial Networks} \label{ssec:uavn}
While traditional base stations offer good network coverage over longer distances, there arise situations where the existing infrastructure may be damaged due to unforeseen circumstances. This has motivated researchers to find a solution that can serve as a temporary solution while the existing infrastructure is brought back to working order. This has given rise to Unmanned Aerial Networks. In this section, we will first discuss unmanned aerial networks and its applications and then discuss how GNNs are used to address some of the challenges. Unmanned Aerial Networks (UANs) refer to networks formed by interconnected Unmanned Aerial Vehicles (UAVs), also known as drones, operating cooperatively to achieve various objectives. Due to the portable nature of the UAVs, they can be dynamically deployed when required and play a crucial role in establishing temporary connectivity in the event of disaster or infrastructure failure and enable high data rate network connectivity. Figures~\ref{fig:uan1} and \ref{fig:uan2} show the Typical components of UANs and applications of GNN for UAN tasks, respectively. ML techniques such as RNNs and RL are commonly used for UAN problems such as spectrum access and flight control for microwave, millimeter wave, and Terahertz frequency bands  \cite{Jagannath20UAVBook, skm_secon_2020, PoloskyRobocom22, skm_tnet_2022, skm_tnet_2023,skm_secon_2021}, UAV swarm collaborative beamforming \cite{skm_comnet_2023,skm_icc_2021,skm_wcnee_2020,skm_tmc_2022}. In addition to traditional ML techniques GNNs are also commonly used for UANs problems such as Sensing \cite{Liu_iotj_2021} and networking applications such as trajectory planning \cite{ Zhang_twc_2023, An_access_2024}, resilient UAV swarm communication \cite{Mou_jsac_2022},  Flying Relay Location and Routing Optimization \cite{Wang_remsen_2022}.

In \cite{Liu_iotj_2021}, the authors propose a GCN and GAT-based LSTM model to accurately predict the real-time and future Air Quality Index (AQI) using aerial-ground sensing framework. This model uses a lightweight Dense-MobileNet model \cite{Andrew_mobilenet_2017} to achieve energy-efficient end-to-end learning from features extracted from images captured by the UAV. This framework allows for federation across various institutions to collaboratively learn a well-trained global model to accurately and securely monitor the AQI and also expand the scope of UAV swarm monitoring. In \cite{Zhang_twc_2023}, the authors present a GCN-based framework to solve the cooperative trajectory design problem formulated based on modeling the relationship between the ground terminal and  UAV-base stations (UAV-BS). This framework enables UAV-BS to efficiently manage the time-varying local observation such as current location, average throughput, and scheduling and facilitate cooperation between UAV-BSs through information exchange. Due to the lack of global information and a globally optimal solution, a Q-learning-based multi-agent RL algorithm is used by the UAV-BS to learn a distributed trajectory policy which is modeled as a Markov game. Here, the message passing by the GCN framework ensures cooperative behavior among UAV-BS and significantly enhances the network's performance. In \cite{An_access_2024}, the authors propose a multi-dimensional trajectory prediction model for UAV swarms by integrating GNN with a dynamic graph neural network (DynGN). The DynGN uses an encode-decoder structure with a GCN to process the dynamic adjacency matrix, extract spatial features, and gated recurrent unit to model the temporal dynamics and capture the evolution of UAV interactions over time. This model processes both the dynamically changing network configuration and the trajectory data of the UAV swarm simultaneously. 

In \cite{Mou_jsac_2022}, the authors study the self-healing problem of UAV swarm network to re-establish network connectivity in the event of unpredictable destruction. These destruction may affect one or more UAVs simultaneously and can either be one-off events or recurring events. To address the one-off events, a GCN-based framework is designed to find the recovery topology of the UAV swarm network in a timely manner. This framework is then extended by incorporating a monitoring mechanism to detect UAV destruction and a self-healing trajectory planning algorithm based on GCN to address the recurring events of destruction. Finally, a meta-learning scheme is designed for the GCN to reduce the execution time complexity and ensure faster convergence of GCN. In \cite{Wang_remsen_2022}, the authors propose a GNN-based framework for the optimization of UAV locations and relay paths in UAV-relayed 6G IoT networks. A two-stage GNN model is developed to decouple the problem of optimal relay path selection and UAV position optimization and is solved by a separate GNN model. Here, both stages are trained using reinforcement and unsupervised learning with no training data or prior knowledge of optimal solutions. First, the locations of the UAVs are optimized and second, the best relay path is determined based on the optimized location of the UAVs from the first step. Results show that the proposed model performs better than brute search with lower time complexity, scalable to very large networks with dynamically changing environments. 

\begin{figure*}[t]
\centering
\includegraphics[width=1\textwidth]{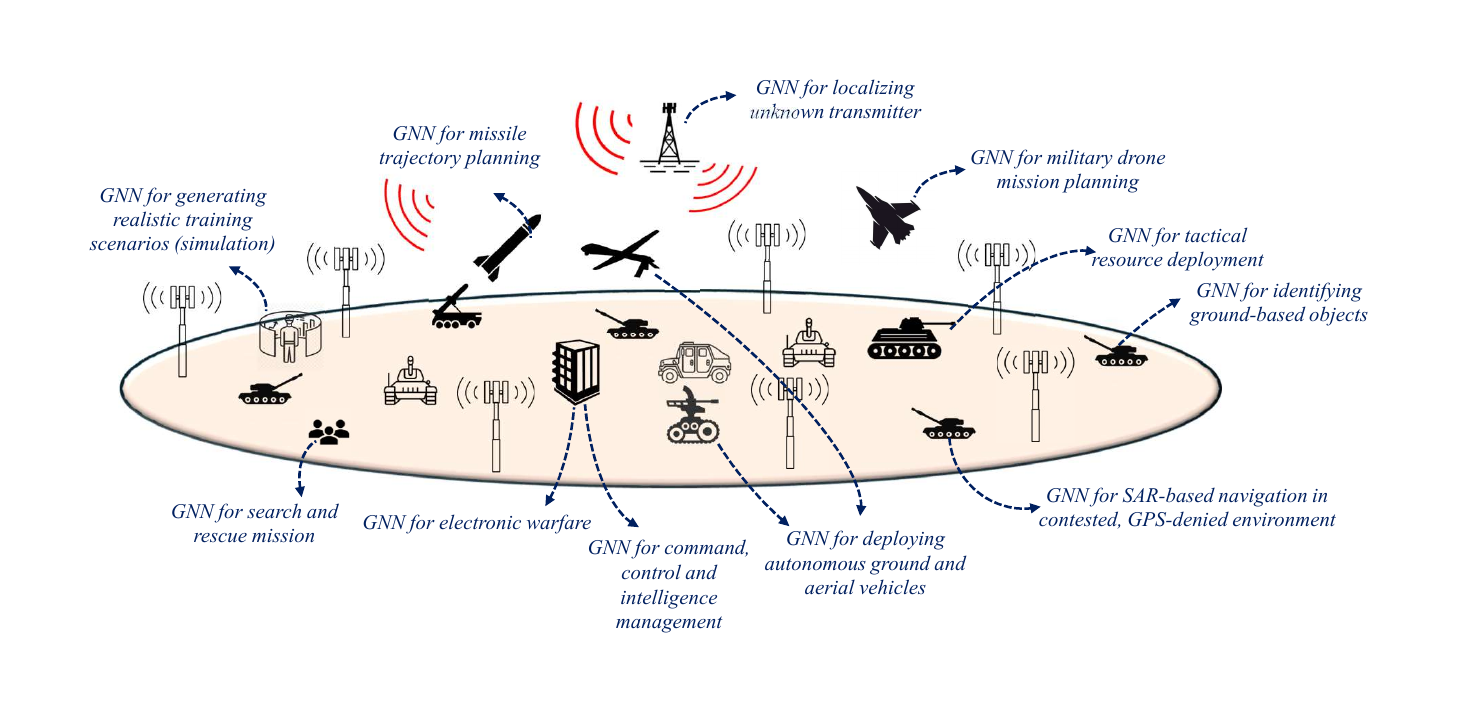}
\caption{Application of GNN for tactical system tasks.}
\label{fig:tact}
\end{figure*}

\subsection{Lessons Learned}
In this section, we present the key takeaways and lessons learned from the applications of GNN for Networking discussed in Sections~\ref{ssec:pred} - \ref{ssec:uavn}.
\begin{enumerate}
    \item Despite their advantages, GNNs face scalability challenges as network size increases. The computational complexity of GNNs grows with the number of nodes and edges, making it difficult to apply them directly to large-scale networks like the Internet or hyper-scale data centers. Large networks require extensive message passing between nodes, leading to increased memory and processing requirements. Techniques such as sampling \cite{liu_jas_2022, Weichen_enn_2023},  and distributed GNN architectures \cite{md_ichpcnsa_2021, zheng_ia3_2020} can be explored to mitigate these scalability constraints.
    \item Networking operates across multiple layers of the OSI model, and performance optimization often requires a holistic, cross-layer approach. GNNs have the potential to model interactions between different layers, such as how physical-layer signal interference affects network-layer routing decisions. However, most research in GNN-based networking has focused on single-layer optimization, such as routing or traffic prediction. The exploration of multi-layer GNN models could lead to breakthroughs in end-to-end network performance optimization, including adaptive modulation, dynamic bandwidth allocation, and interference-aware routing.
    \item UAV networks such as drone swarms and airborne communication relays, introduce unique challenges due to their rapidly changing topologies. Traditional static network models struggle to adapt to such dynamic environments. GNNs, with their ability to process graph structures dynamically, can model UAV network topologies in real-time, enabling efficient routing, link prediction, and connectivity management. They can be particularly useful for optimizing communication in ad hoc UAV networks where maintaining connectivity is a key challenge.
    \item In conclusion, GNNs represent a transformative approach to network management and optimization, offering powerful capabilities in understanding and predicting complex network behaviors. Their ability to model network topologies, adapt to dynamic environments, and optimize multiple layers simultaneously makes them highly attractive for applications such as traffic engineering, anomaly detection, and resource allocation. However, significant challenges remain, particularly in scaling GNNs to large networks, ensuring real-time processing, and designing energy-efficient architectures. Addressing these challenges through novel algorithmic and hardware advancements will be key to unlocking the full potential of GNNs in next-generation networking. As research in this domain progresses, GNNs are poised to play a crucial role in building intelligent, resilient, and self-optimizing networks of the future.
\end{enumerate}

\section{GNN for Tactical Systems}\label{sec:gnn_tact}

Modern warfare and defense operations rely on tactical systems to ensure effective coordination, secure communication, and resilient decision-making in highly dynamic and adversarial environments. Unlike conventional communication systems, tactical networks must adapt in real time, detect and respond to threats, and ensure operational integrity despite interference or adversarial attacks. These requirements extend beyond the capabilities of traditional NextG networks, necessitating intelligent, resilient, and adaptive solutions. Tactical systems rely on interdependent components such as distributed sensors, mobile units, and adversarial entities which is hard to be captured and modelled using traditional NNs but GNNs can be used effectively to model these complex interactions. To this end, in this section, we discuss the application of GNN for Tactical Communication and Sensing Systems (Section~\ref{ssec:tact}), Target Recognition (Section~\ref{ssec:target}), and Localization (Section~\ref{ssec:localize}). Figure~\ref{fig:tact} shows applications of GNN for tactical system tasks. Tactical systems are an extension of NextG networks where it is not enough to be aware of the spectrum and improve the QoS, but also be able to take countermeasures to detect, and localize any nodes or adversary causing interference or affecting the integrity of the NextG networks. Table~\ref{tab:gnn_tac_table} provides a high-level comparison of surveyed paper using GNNs for tactical applications. We also provide the dataset or the simulator used for performance evaluation.

\begin{table*}[!ht]
\centering
\caption{Comparison of surveyed papers using GNN for Tactical Systems Applications.}
\renewcommand{\arraystretch}{2.2}
{\begin{tabular}{|m{2.3cm}|m{2.5cm}|m{3.5cm}|m{3.5cm}|m{3.5cm}|}
\hline

\textbf{Section} & \textbf{Paper} & \textbf{Application} & \textbf{ML Algorithm Used} &  \textbf{Dataset/Simulator Used} \\
\hline
\hline

\multirow{9}{=}{\textbf{Section \ref{ssec:tact}}\\Tactical Communication and Sensing Systems} 
& \textit{Nie et al.} \cite{Nie_cac_2020} & Military decision support & Context-surrounding GNN with numbers & DBP15K \cite{hou2020measuring}, own Milassistant dataset \\
\cline{2-5}
& \textit{Xu et al.} \cite{Xu_csecs_2023} & Mission planning & GNN & Own python-based simulator \\
\cline{2-5}
& \textit{Sant'Ana da Silva et al.} \cite{Silva_tnsm_2023} & Decision making in intelligent transportation systems  &  GNN   & FB15k \cite{Bordes_anips_2013} \\
\cline{2-5}
& \textit{Liu et al.} \cite{Qi_itsc_2022} & Decision-making in interactive traffic scenarios  & GNN + RL & Own python-based simulator  \\
\cline{2-5}
& \textit{Yang et al.} \cite{yang_mdpis_2022} & Decision-making in autonomous driving & GNN + RL & SUMO simulator \cite{SUMO2018} \\
\cline{2-5}
& \textit{Liu et al.} \cite{Liu_wsc_2021} & Behavior Prediction  & GNN + RL & RIDE \cite{Hartholt_ride}, OpenAI Gym \cite{opeai_gym} \\
\cline{2-5}
& \textit{Tang et al.} \cite{Rugang_eeaai_2024} & Real-time resource dispatching  & GNN + RL & Own simulator \\
\cline{2-5}
& \textit{Li et al.} \cite{Biyue_etret_2024}  & Airspace complexity prediction & Spatio-temporal graph neural network & Chinese domestic airspace traffic data  \\
\cline{2-5}
& \textit{Okine et al.} \cite{Okine_tnsm_2024} & Tactical wireless sensor networks & GNN + RL & Own python-based simulator \\
\hline

\multirow{6}{=}{\textbf{Section \ref{ssec:target}}\\Target Recognition} 
& \textit{Wijeratne et al.} \cite{Sasindu_spie_dcs_2023}  & Target recognition & GraphSAGE + CNN &  MSTAR \cite{mstar_afrl}  \\
\cline{2-5}
& \textit{Ye et al.} \cite{Tian_spie_dcs_2023}  & SAR image classifier  & GNN & MSTAR \cite{mstar_afrl} \\
\cline{2-5}
& \textit{Zhu et al.} \cite{Zhu_senlet_2020}  & Target classification & Graph Convolutional Network & MSTAR \cite{mstar_afrl} \\
\cline{2-5}
& \textit{Xu et al.} \cite{xu_tgrs_2024} & Hyperspectral target detection & GNN + CNN &  HYDICE dataset, AVRIS San Diego Dataset \cite{avris_1998}, Airport Beach Urban Dataset   \cite{avris_nasa}\\
\cline{2-5}
& \textit{Chen et al.} \cite{chen_tgrs_2024}  & Target detection & Graph Convolution Networks + Graph Transformer & AVRIS San Diego Dataset \cite{avris_1998}, Urban Dataset \cite{avris_nasa}, MUUFL Gulfport Dataset \cite{MUUFL}  \\
\cline{2-5}
& \textit{Yang et al.} \cite{yang_taes_2023} & Aerial target pose estimation & Motion
flow-based Graph Convolutional Network & Microsoft Visual Studio air combat simulation platform  \\
\hline

\multirow{4}{=}{\textbf{Section \ref{ssec:localize}}\\Localization} 
& \textit{Lezama et al.} \cite{Lezama_2021_urucon}& Wi-Fi fingerprinting based indoor localization & Graph Convolutional Network & MNAV \cite{antonio_museum_2020} ; UJIIndooeLoc \cite{Torres_pin_2014} \\
\cline{2-5}
& \textit{Zhang et al.} \cite{Zhang_2023_iotj}& WiFi indoor localization & Domain adversarial graph convolutional network & Microsoft Indoor Location Competition 2.0 data set \cite{microsoft_indoor} \\
\cline{2-5}
& \textit{Yan et al.} \cite{yan_2021_icassp}  & Large-scale network localization & Graph Convolutional Network & Own simulator \cite{yan_github} \\
\cline{2-5}
& \textit{Boyaci et al.} \cite{Boyaci_2022_isg} & Detection and localization of stealth false
data injection attacks & Auto-Regressive Moving Average Graph Filters-based GNN & NYISO  \cite{nysio} \\
\hline

\end{tabular}}
\label{tab:gnn_tac_table}
\end{table*}

\subsection{Tactical Communication and Sensing Systems} \label{ssec:tact}

Tactical Communication and Sensing Systems refer to communication networks and technologies specifically designed for military or tactical operations. These systems aim to provide reliable, secure, and efficient communication among various military assets, including soldiers, vehicles, sensors, and other connected devices, in dynamic and challenging environments. GNNs can be used to enhance situational awareness on the battlefield. By modeling the battlefield as a graph where nodes represent military assets (e.g., vehicles, soldiers) and edges represent relationships or interactions (e.g., proximity, communication), GNNs can analyze the evolving battlefield situation in real-time. This can aid in decision-making and mission planning \cite{Nie_cac_2020, Xu_csecs_2023, Silva_tnsm_2023, Qi_itsc_2022, yang_mdpis_2022}, training simulations \cite{Liu_wsc_2021}, human-intelligence collaboration \cite{Rugang_eeaai_2024}, airspace complexity prediction \cite{Biyue_etret_2024} and  tactical mobile networking \cite{Okine_tnsm_2024}.

In \cite{Nie_cac_2020}, the authors propose context-surrounding graph neural networks with numbers (CS-GNN-N) for knowledge-based reasoning for military decision support. One of the major components is \textit{GNN-based learning} which utilizes two smoothness metrics, namely feature smoothness and label smoothness to measure the quantity and quality of neighborhood information of nodes. In \cite{Xu_csecs_2023}, the authors present a GNN-based mission planning approach for unmanned ground vehicles (UGV) to generate optimal mission planning results using recursive graph grammar to describe the task assignment process of multiple UGVs as an irregular graph structure and by coupling both task assignment as well as path planning. The performance of the proposed algorithm was verified both using simulation and real-world UGV robots. In  \cite{Silva_tnsm_2023}, the authors propose \textit{GraphNearbyAlert} (GNA) framework to effectively broadcast cooperative awareness messages to a group of vehicles. The results show that it is sufficient to transmit very little information about the topology of the graph to extract information from the shared data and can be used for decision-making in cooperative transportation systems while preserving privacy. In \cite{Qi_itsc_2022} the authors present a GCN framework integrated with deep Q network-based RL-based to achieve an efficient and reliable multi-agent decision-making system for safe and efficient operation of connected autonomous vehicles in intelligent transportation systems. The GCN generates the features of the network based on the topology and is used by the RL to generate optimal decisions. In \cite{yang_mdpis_2022}, the authors present an improved single-agent GNN-based algorithm for decision-making in autonomous vehicles, where the GNN is first used to generate features for a multi-agent system and then transferred to a single-agent system. The advantage of this approach is that the newly trained single-agent model can then be used to train scenarios with multiple agents. This design significantly reduces the time complexity of training a multi-agent model and improves the convergence speed of the model.  

In \cite{Liu_wsc_2021}, the authors introduce a novel GNN-based behavior prediction model to train and develop synthetic characters for roles in the Rapid Integration and Development Environment (RIDE). RIDE is a unity-based military training simulation environment that facilitates rapid development and prototyping of simulated environments in direct service of the Army and other Department of Defense entities. Through simulation experiments, the authors show that a GNN model trained using pseudo-human behavior can be used as a behavior predictor and help with better cooperation between human agents. In \cite{Rugang_eeaai_2024}, the authors propose a GNN-based DRL framework to autonomously accomplish real-time resource dispatching and allow human intervention to accomplish complex tactics. In this design, the GNN is used to generate an integrated scheduling model of detection, tracking, and interception and transformed into a sequential decision problem. The GNN model is integrated with the proximal policy optimization (PPO) algorithm to learn the air defense environment. This framework enables human-intelligence collaborative dynamic scheduling for emergency response scenarios. In \cite{Biyue_etret_2024}, the authors propose a multi-modal adaptive STGN framework to simultaneously explore spatio-temporal dependencies in the airspace network with the objective to effectively learn the diverse spatial relationship and adaptively adjust the impact of different spatial models. The results show that the framework can harness differing spatial models and achieve high generalization performance across different temporal patterns compared to state-of-the-art methods. In \cite{Okine_tnsm_2024}, the authors present a GNN-based multi-agent DRL routing algorithm for multi-sink tactical mobile sensor networks to overcome link-layer jamming attacks. The proposed algorithm captures the hop count to the nearest sink (i.e., ), the one-hop delay, the next hop's packet loss, and the energy cost of packet forwarding. The results show that the proposed scheme outperforms compared to other state-of-the-art algorithms.

\subsection{Target Recognition}\label{ssec:target}

While Tactical systems discussed in the previous section help in making critical decisions and mission planning, it is also important to manage and keep track of deployed resources. These resources can either be personnel or deployed ground and aerial resources. In this section, we will discuss several applications where GNNs are commonly used for target recognition. In addition to its use for decision-making, GNN has widely been used for automatic target recognition. Synthetic Aperture Radar (SAR) images are commonly utilized in military applications for automatic target recognition (ATR). SAR is a remote sensing technology that uses radar to create high-resolution images of the Earth's surface. SAR works by emitting radio waves at a target and then measuring the reflection of those waves. Unlike optical imaging, SAR can operate day or night and in various weather conditions because it doesn't rely on visible light. Due to this ability, SAR images are widely used in the military for ATR \cite{Sasindu_spie_dcs_2023, Tian_spie_dcs_2023, Zhu_senlet_2020}.  In addition to SAR images there are also other techniques used for target recognition such as Hyperspectral images \cite{xu_tgrs_2024, chen_tgrs_2024} and Infrared thermal imaging \cite{yang_taes_2023}. In \cite{Sasindu_spie_dcs_2023}, authors use GNN to identify ground-based objects such as battle tanks, personnel carriers, and missile launchers, among others. By identifying the type of objects, it would allow the military to determine if the objects are friendly or adversarial. However, the classification is susceptible to the noise level in the SAR images which may degrade the performance of the GNN classifier. The effect of an adversarial attack on GNN-based SAR image classifier has been studied extensively in \cite{Tian_spie_dcs_2023}. Furthermore, extraction of features from the original SAR image requires a large amount of data and is time-consuming. To overcome this issue, in  \cite{Zhu_senlet_2020}, the authors use SAR image target pixel gray-scale decline by a graph representation. Here, the gray-scale SAR image is divided into sub-intervals and a node is assigned to represent each pixel with declined order of pixel gray-scale in the sub-interval. The raw SAR image is then transformed to construct a graph structure which is used by the GCN to extract the features of the graph structured data and then classify the target. 

Hyperspectral imaging (HSI) is a process of capturing images with a wide spectrum of light across many narrow wavelengths. Due to this, compared to traditional imaging, HSI can help extract more detailed and accurate information such as precise color and material identification. In \cite{xu_tgrs_2024}, the authors propose CFGC, a cognitive fusion of GNN and CNN for enhanced hyperspectral target detection (HTD), where CNN is used to extract multi-scale features and then a depth-wise separable
convolution-self attention (DSC-SA) mechanism is used to reduce the number of CNN parameters while enhancing the feature space. The GNN local aggregation and global attention block (LAGAB) \cite{Zhonghao_lagab_2023} mechanism to extract the global information. Compared to state-of-the-art detection algorithms results show that CFGC achieves high robustness and better detection performance. Similarly, in \cite{chen_tgrs_2024}, the authors propose a spectral graph transformer-based HTD method where a pixel spectrum is first constructed as a spectral graph and a GCN is used to extract the local information of the spectrum. Then, a self-attention transformer is used to learn the global information of the spectrum/ The framework also integrates a graph contrast clustering algorithm to enable high spectral discrimination ability. In \cite{yang_taes_2023}, the authors develop a GCN-based adaptive graph reasoning method for aerial target pose estimation modeled based on the eye-tracking properties of humans. The GCN is employed to derive the dynamic prediction of the object motion based on the spatio-temporal and visual data. The GCN uses the historical pose trajectories to predict the corresponding pose in the next time instant. This framework can help detect the target as well as its trajectory.

\subsection{Localization}\label{ssec:localize}

In the previous section, we discussed the application of GNN for target recognition. The obvious next step is to accurately pinpoint the location of the deployed resources as well as keep track of the movement through a process called localization. Localization refers to the process of determining the location or position of an object or entity within a given environment. It is a fundamental task in various fields, including robotics, navigation, mobile computing, and wireless communication. Accurate localization is crucial for applications such as GPS-Based Navigation, Augmented Reality (AR), and autonomous vehicles. GNNs are used for localization in certain contexts due to their ability to capture spatial relationships and dependencies among different elements in a graph-based representation of the environment. GNNs can integrate information from different types of sensors or data sources within the same graph structure. This is beneficial in scenarios where multiple sensors (e.g., GPS, Wi-Fi, LIDAR, cameras) are available for localization. Localization data can be noisy and uncertain due to factors like signal interference, multi-path effects, and environmental changes. GNNs can learn to handle such noise and uncertainty by aggregating information from neighboring nodes and making robust predictions. 

GNNs are widely used for indoor localization \cite{Lezama_2021_urucon} \cite{Zhang_2023_iotj}, large-scale network localization \cite{yan_2021_icassp} and false data injection attack localization  \cite{Boyaci_2022_isg}. For example, in \cite{Lezama_2021_urucon} graphs are generated using the RSSI from Wi-Fi access points to localize the device into certain predefined areas using GCN. Similarly, in \cite{Zhang_2023_iotj}, Zhang et al. proposed a novel WiFi domain adversarial graph convolutional network model for indoor localization using RSSIs between way-points and WiFi access points. The difference in the two works is that in the latter unlabeled data and multiple data domains are considered. It also uses a semi-supervised domain adversarial training to efficiently use the unlabeled data as well as different data distributions across data domains. While GNNs are widely used for indoor localization, they also find application in large-scale network localization such as \cite{yan_2021_icassp} where a network with 500 nodes was considered in comparison to 25 nodes/devices in \cite{Lezama_2021_urucon}. In \cite{Boyaci_2022_isg}, the authors propose a GNN-based approach to identify the presence of a false data injection attack as well as localize the source of the attack. 

\subsection{Lessons Learned}
In this section, we present the key takeaways and lessons learned from the applications of GNN for Tactical Systems discussed in Sections~\ref{ssec:tact} - \ref{ssec:localize}.
\begin{enumerate}
    \item Tactical environments, such as battlefields and military operations, involve highly dynamic and complex interactions among various entities, including soldiers, vehicles, and drones. These interactions are inherently interconnected, as communication and coordination between units play a crucial role in mission success. GNNs provide enhanced situational awareness, enabling military units to make more informed decisions. By accurately capturing the evolving state of the tactical environment, GNNs assist in mission planning, unit coordination, and threat assessment, ultimately improving operational effectiveness.
    \item In battlefield scenarios, target recognition is often complicated by ambiguous or cluttered surroundings, where multiple objects can be closely positioned or occluded. Traditional object detection methods may struggle to differentiate between targets and non-targets in such scenarios. GNNs improve target recognition by leveraging spatial and relational information between detected objects. By considering contextual relationships such as the expected proximity of friendly and enemy assets, GNNs can refine detection accuracy and reduce false positives or negatives. This capability is particularly beneficial in surveillance and reconnaissance tasks, where precision and rapid identification are critical.
    \item Effective localization and tracking of assets and targets are essential in tactical operations. These tasks often involve multiple sensors, including GPS, radar, and communication nodes, which must collaboratively determine real-time positions. GNNs improve localization accuracy even in GPS-denied environments \cite{liu_segcn_2024, zha_sens_2023, wang_remsens_2024}. This ensures that commanders maintain accurate knowledge of friendly and enemy unit positions, thereby optimizing resource deployment and mission execution.
    \item Modern military operations require seamless interoperability across multiple domains, including land, sea, air, space, and cyber. Traditional communication systems often struggle to integrate information from these diverse domains due to differences in data formats, protocols, and communication architectures. GNNs facilitate cross-domain interoperability by modeling interactions between heterogeneous data sources within a unified graph framework. This capability allows for coordinated decision-making across different operational theaters, ensuring that forces can act cohesively even in highly distributed environments.
    \item Training and simulation play a vital role in preparing military personnel for real-world operations. GNNs can be used to model various tactical scenarios, allowing commanders and troops to simulate battlefield conditions and assess different strategies. By analyzing past mission data and generating realistic engagement models, GNN-based simulations provide an advanced training environment where personnel can refine their decision-making skills under diverse operational conditions. This leads to better preparedness and adaptability in actual combat situations.
    \item In conclusion, GNNs presents a great opportunity for enhancing tactical communication systems by providing superior modeling and real-time decision-making abilities. By integrating GNNs thoughtfully within tactical systems, military forces can achieve enhanced coordination, strategic advantage, and operational success in dynamic and high-stakes environments.
    
\end{enumerate}

\section{Opportunities and Roadmap ahead} \label{ssec:opp_roadmap}

As seen in this article, GNNs have been widely adopted for a wide range of applications but several areas are in search of breakthroughs to make GNNs more impactful. This provides an opportunity for the research community to further develop new features for GNN as well as apply GNN for even broader research areas. To this end, in this section, we discuss the possible avenues of research to improve the capabilities of GNN and research areas where GNN will play a crucial role in IoT and NextG networks. Figure~\ref{fig:research} shows the key areas of research opportunities.

\begin{enumerate}
    \item \textbf{Scalable and Reliable GNN.} \textit{Scalability} is one of the crucial aspects to gauge the success of any ML algorithm and GNN is not an exception. As of now, most of the GNN algorithms are tested on a smaller network topology and present good results with a slightly larger network to show the scalability. However, with the evolving number of IoT devices and wireless devices, it is crucial to develop algorithms that can scale up to thousands or hundreds of thousands of devices without any possible degradation in performance. GNN promises to be a great tool that can enable scalable ML models, it is important for the research community to further investigate and rigorously test their algorithm from the perspective of scalability on real networks. \textit{Reliability} is another important area that requires further attention. As of now, most of the GNN-based algorithms designed assume a perfect environment with pre-defined interference, channel, and other relevant models for system design. It is important to design GNNs that are robust to adversarial attacks, sensor malfunction, data corruption, and other unforeseen conditions. This is especially important when there are millions of connected devices and data from one or more sensors should not impact the whole network. This can be addressed by systematically identifying, quantifying, and rectifying anomalies or security risks. This promises to be an interesting area of research and will significantly benefit in maturation and adoption of GNN across different domains.
    
    \item \textbf{Capturing Network Complexity.}
    GNNs are often trained on a smaller network size and relatively smaller dataset which are insufficient to capture the underlying complexity of the network. Another possible situation with collected datasets is that they can be imbalanced. This imbalance can occur when more data points are collected for a perfect scenario without any abnormal or noisy cases. When this data is used for training a GNN model, the developed model will work well for perfect cases while failing in cases with noise or any perturbations. Techniques such as Data Augmentation \cite{data_aug_2021}, and Generative Adversarial Network \cite{GAN_2014} can be integrated with GNN as one of the possible solutions to address the challenge of limited data. The research community is also encouraged to explore the integration of Transfer Learning, Meta Learning, Edge-perturbation, Motif-Sampling as additional possible solutions.

\begin{figure}[t]
\centering
\includegraphics[width=0.48\textwidth]{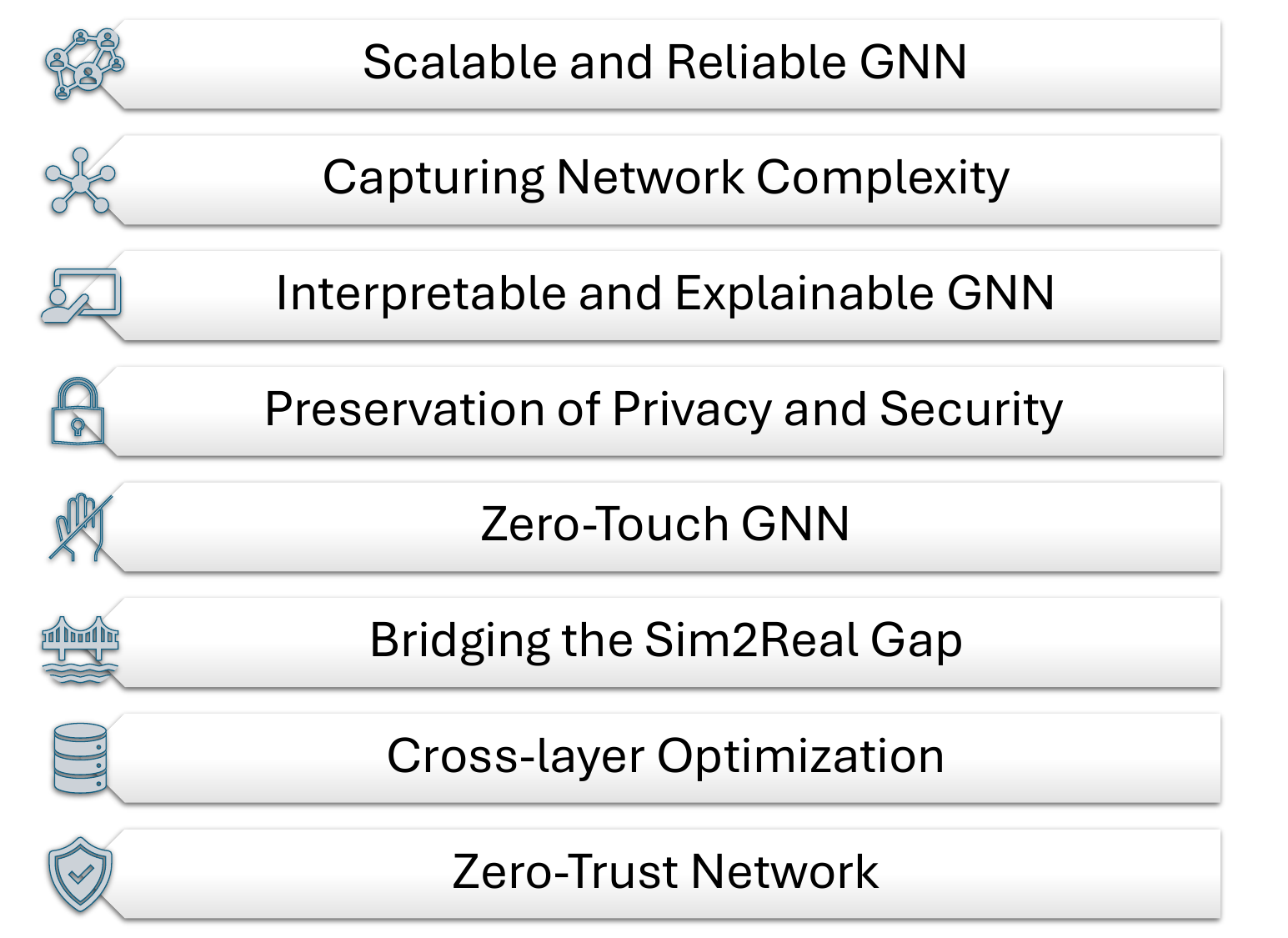}
\caption{Overview of Research Opportunities.}
\label{fig:research}
\end{figure}

    \item \textbf{Interpretable and Explainable GNN.}
    One of the major challenges with any ML-based algorithms is the lack of interpretability. \textit{Interpretability} can be defined as the measure of how much the user can understand the decision of the model. In other words, if the user can understand and predict the output of the ML algorithm then the ML algorithm/model is said to be highly interpretable. Similarly, the users should also be able to understand the factors that affect the performance of the GNN model and ensure that the decisions align with human intuition and domain knowledge. However, in the wireless domain (specifically IoT), with more nodes and heterogeneous data, it is challenging to design a highly interpretable and explainable GNN model.A good starting point for research along the design of interpretable and explainable GNN is GNNExplainer \cite{gnnexplainer}, the first-general, model-agnostic approach that can provide interpretable explanations for any GNN-based model on any graph-based learning tasks. This line of research is very important to further enhance GNN's features, making it a reliable and preferred ML technique across many domains. Another technique for explainable GNN is called Integrated-Gradients (IG) methods \cite{Mukund_IG_PMLR_2017}. A version of IG was used for GNN-based segmentation networks \cite{wu_integrated_2024} where IG is used to explain a DualGCN model applied in the image segmentation domain.

    \item \textbf{Preservation of Privacy and Security.}
    The current applications of GNN for IoT and wireless networks collect data from different sensors and users as well as data shared between sensors (or users). While this is not an area of concern for simulation-based implementations for proof-of-concepts implementations, it is important to integrate privacy and security measures into the GNN algorithm. One of the most promising and well-known techniques for preserving privacy and security is Federated Learning (FL). In FL, instead of sharing the data between nodes/sensors, only the model weights are shared. This enables collaborative learning without compromising data privacy and security. An example of GNN using FL is FedGraphNN\cite{FedGraphNN} which is an open FL system for research on Federated GNNs. The integration of GNN with Federated Learning is an interesting and important area of research for future IoT and next-generation wireless networks. There are also other techniques such as Homomorphic Encryption \cite{Sgaglione_wetice_2019}, Differential Privacy \cite{pei_tdsc_2025}, that can be used to preserve the data privacy for GNN models.

    \item  \textbf{Zero-Touch GNN.}
   In the current implementation of GNNs, the algorithm designer has to grapple with the design of GNN making the process tedious for complex network scenarios. Automated Machine Learning (AutoML) \cite{automl} promises to be a solution to overcome the tedious design process. AutoML is an enabling technique to improve the implementation of ML models such as GNN by automatically learning the important steps of the ML model design process. There is some progress towards Zero-Touch GNN such as Auto-GNN \cite{autognn} which helps in finding the optimal GNN architecture within a predefined search space and Auto-STGCN \cite{autostgcn} which helps in generating optimal STGCN models automatically by quickly scanning over the parameter search space. There are also other techniques such as Graph Neural Network Neural Architecture Search (GraphNAS) \cite{GraphNAS_2019}, GraphNAS++ \cite{gao_plusplus_2023}, Parallel Graph Architecture Search GraphPAS \cite{chen_pas_2021}, Differentiable Architecture Search (DARTS) \cite{liu_darts_2019} that can be used to design best GNN models automatically. Zero-Touch GNN will play an important role in increasing the efficiency of the GNN modeling process and requires significant attention from the research community to make it a widely adopted strategy in designing the GNN model for the future of wireless networks.
    
    \item \textbf{Bridging the Sim2Real Gap.} In most of the existing literature, the researchers tend to focus more on two aspects of validating the performance of designed the GNN model. First is the use of open-source datasets and second is the simulation environment. This creates a gap between the performance of the proposed GNN algorithm in a simulation environment and the real-world environment. There are several National Science Foundation (NSF) Platforms for Advanced Wireless Research (PAWR) platforms such as Colosseum \cite{colosseum}, Platform for Open Wireless Data-driven Experimental Research (POWDER) \cite{powder}, Cloud Enhanced Open Software Defined Mobile Wireless Testbed for City-Scale Deployment (COSMOS) \cite{cosmos}, Agriculture and Rural Communities (ARA) \cite{ara}, Aerial Experimentation and Research Platform for Advanced Wireless (AERPAW) \cite{aerpaw} and other testbeds such as ARENA \cite{arena}, CloudRAFT \cite{CloudRAFT}, UnionLabs \cite{UnionLabs}. All these platforms are available to the research community to conduct real-world experiments. In addition there are also other open-source platforms such as Open Network Automation Platform (ONAP) \cite{onap}, which is a unified platform for orchestrating, managing, and automating network and edge computing services across telecom operators, cloud providers, and enterprises while Central Office Re-architected as a Datacenter (CORD) \cite{cord}  transforms legacy central offices into cloud-native edge data centers using SDN/NFV to support low-latency, flexible service delivery.  Furthermore, most articles reviewed in this survey fails to discuss how the proposed GNN frameworks could integrate with existing 3GPP or IEEE protocols. It is recommended that future research discuss this integration to demonstrate the possibility of adopting the proposed solution in real-world large-scale commercial networks. The researchers can share the test scenario, code, and dataset with the community to accelerate the growth of GNN as well as next-generation wireless networks.

    \item \textbf{Cross-Layer Optimization.} In wireless networks, each layer is interconnected, and changes made in one layer have direct or indirect effects on other layers. Currently, most of the state-of-the-art literature of GNN focuses on optimizing only one layer, for example, the physical layer \textit{sensing} or network layer \textit{routing}. However, it is important to jointly optimize more than one layer to achieve better network performance \cite{lin2006tutorial, JagannathComnet22}. A rare example of GNN-based cross-layer optimization is GCLR \cite{gclr_acc_2020} which is a GNN-based cross-layer optimization system for Multi-path TCP protocol.  This framework helps in predicting the expected throughput and the best routing path based on the given network topology and multi-path route candidate set. While traditional optimization has been extensively used for cross-layer optimization problems \cite{Jagannath18TMC, LDing10TVT, Jagannath16GLOBECOM, JagannathMILCOM21, colonnese2017cross, Jagannath19WOWMoM}, GNN promises to be one of the best solutions for large-scale, dynamically changing networks.  

    \item \textbf{Zero-Trust Networking.} Security is always an active research area for IoT and NextG networks \cite{mao2023security, Ramezanpour22Comnet} and the Zero-Trust approach is one key direction the industry is moving towards. Zero-Trust architecture (ZTA) \cite{zta_nist} is a framework that works on the underlying principle that no one accessing the network and its resources can be blindly trusted. The objective is to prevent data breaches and other cybersecurity attacks. While there are system checks in place such as authorizing users based on their unique ID and password, ZTA's main function is to authorize each user's requests on a case-by-case basis instead of authorizing the user. The key elements of ZTA are dynamic risk assessment and trust evaluation. Application of Artificial Intelligence has been significantly explored by researchers \cite{hussain_wc_2024, hosney_icci_2022} as well as intelligent ZTA for 6G networks \cite{izta_comnet_2022,sedjelmaci_ieeen_2023, enright_fnwf_2022}. Given the vast features of GNN, it promises to be a key enabler for ZTA-enabled systems. GNN can be used for Intelligent Network Security State Analysis (INSSA) \cite{izta_comnet_2022}, Trust Evaluation \cite{huo_tnnls_2023}. While there has been preliminary research done on GNN for Zero-Trust systems, it remains an exciting area of research towards 6G and NextG networks. 
\end{enumerate}

\section{Conclusion}\label{sec:conclusion}
In this article, we presented a comprehensive survey of GNN with a focus on the IoT and NextG networks. First, we provided a detailed description of GNN by starting with the common terminologies and their definitions before elaborating on its architecture. To complete this introduction to GNN, we provide detailed insight into the commonly used GNN variants. Second, we discussed the applications of GNN for IoT where we reviewed the existing literature focusing on IoT Network Intrusion Detection and IoT Data Fusion. Third, we looked into how GNNs have been leveraged for spectrum awareness and reviewed the existing literature focusing on RF signal classification and RF spectrum sensing. Fourth, we delved into applications of GNN for networking. Here, we discussed the existing literature focusing on applications of GNN for network and signal characteristics prediction, routing optimization, congestion control, mobile edge computing, digital twin networks, and unmanned aerial networks. Fifth, we focused on applications of GNN for tactical systems where we discussed existing literature on tactical communication and sensing systems, target recognition, and localization. Finally, based on the detailed survey that we have conducted, we have listed the possible research directions for improving GNN features as well as open research areas that will benefit from GNN's versatility.

\bibliographystyle{ieeetran}
\bibliography{bibliography}

\begin{IEEEbiography}
[{\includegraphics[width=1.1in,height=1.1in,clip,keepaspectratio]{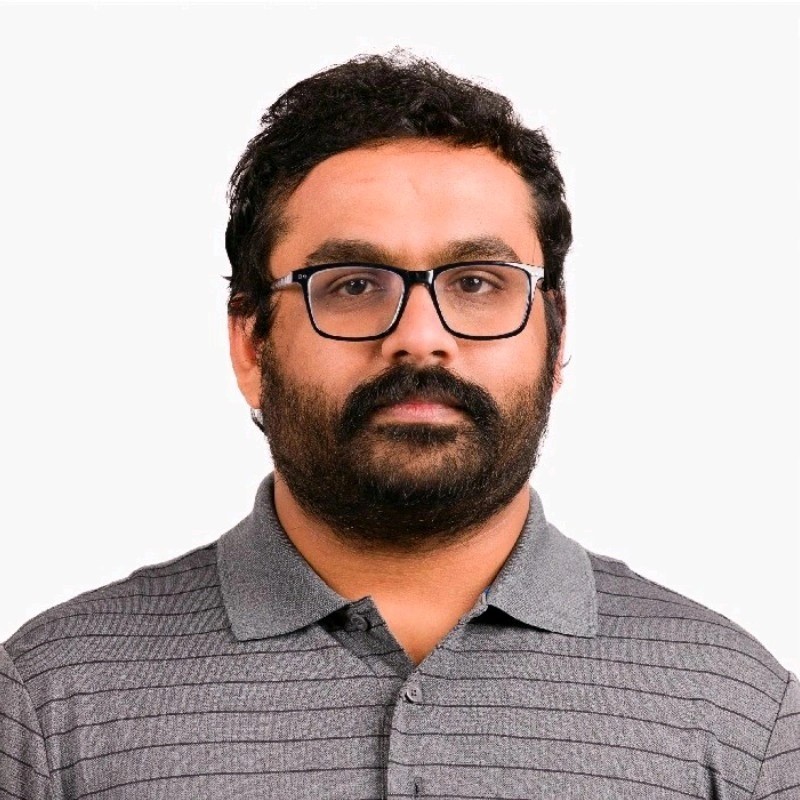}}]{Sabarish Krishna Moorthy} received his Ph.D. and MS degree in Electrical Engineering from the State University of New York at Buffalo. His PhD focussed on designing solutions for automating resource orchestration in software-defined broadband flying networks. He is currently working as a Senior Scientist in the Marconi-Rosenblatt AI Innovation Lab at ANDRO Computational Solutions, LLC. He also worked at Bosch Research and Technology Center as a Robotics AI Intern. Dr. Krishna Moorthy is an active reviewer for several conferences and journals. He also serves as a Technical Program Committee member for IEEE MILCOM. His areas of interest include next-generation spectrum technologies, network design automation, machine learning, software-defined networking, millimeter and Terahertz networks, UAV networking, and Zero-Touch Theories and Algorithms.
\end{IEEEbiography}

\begin{IEEEbiography}[{\includegraphics[width=1.1in,height=1.1in,clip,keepaspectratio]{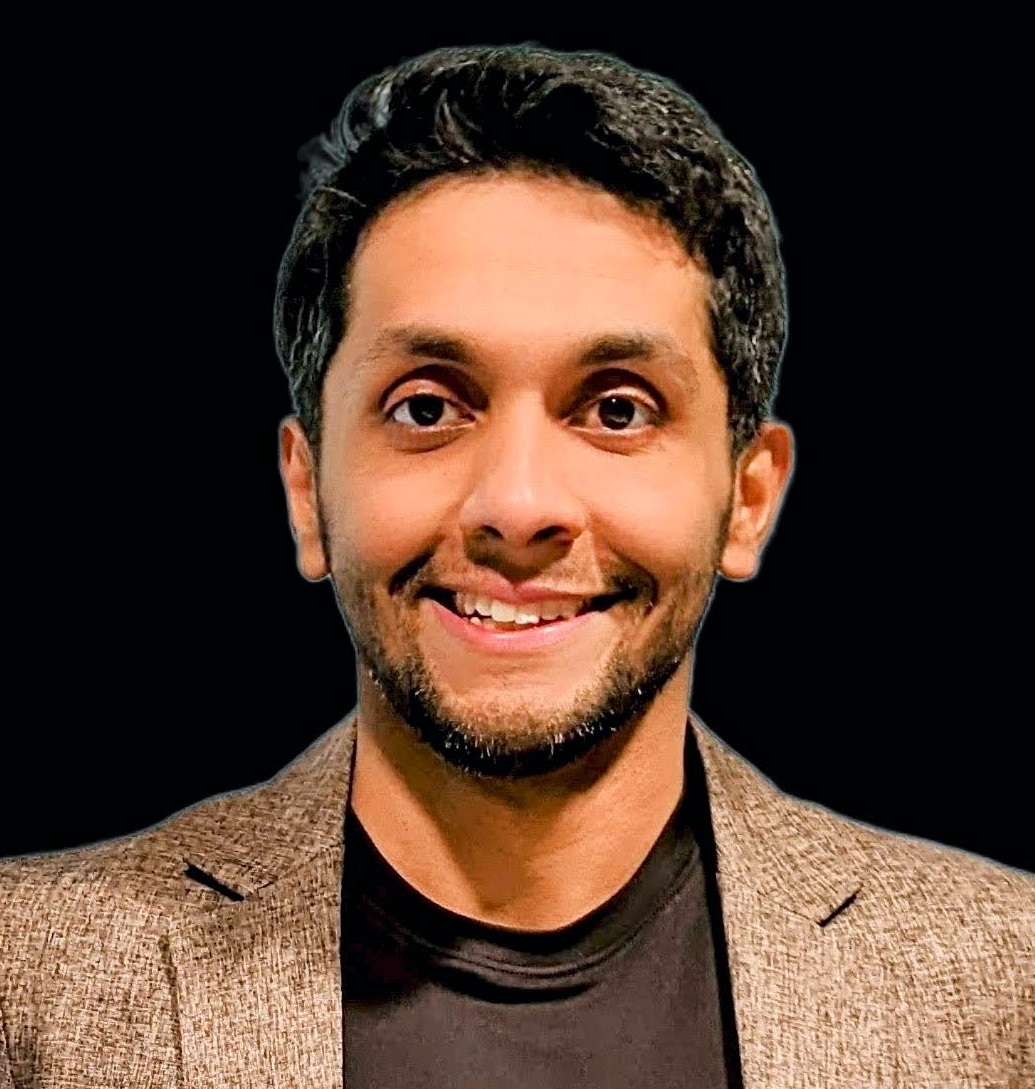}}]{Jithin Jagannath}
is the Chief Technology Officer and Chief Scientist at ANDRO Computational Solutions, LLC, and the Founding Director of ANDRO's Marconi-Rosenblatt AI Innovation Lab. He is also the Adjunct Assistant Professor in the Department of Electrical Engineering at the University at Buffalo. He received his Ph.D. degree in Electrical Engineering from Northeastern University. He has been the Principal Investigator in research efforts for several customers including the U.S. Army, Defense Advanced Research Projects Agency (DARPA), U.S. Navy, United States Special Operations Command, U.S. Department of Homeland Security (DHS), and Air Force Office of Scientific Research (AFOSR) in the field of Radio Frequency machine learning, Beyond 5G, signal processing, RF signal intelligence, cognitive radio, computer vision, and AI-enabled wireless. Dr. Jagannath serves on the editorial board of IEEE Internet of Things Journal and Computer Networks (Elsevier), and serves on Technical Program Committees (TPCs) of several leading IEEE conferences.

Dr. Jagannath is frequently invited to deliver keynote addresses and participate in panels on machine learning and Beyond 5G wireless communication at leading conferences and events, both domestically and internationally. His engagements include major IEEE conferences, the Advancement of Artificial Intelligence (AAAI) Symposium, and the AUSA Annual Exposition and Meeting. Additionally, he has served on the review panel for the Fulbright STEM Impact Award 2023/24 project review under the Polish-U.S. Fulbright Commission and has been appointed as the IEEE Signal Processing Society Representative to the Internet of Things Journal Steering Committee. He is the co-inventor of 16 U.S. Patents (granted and pending). Dr. Jagannath was the recipient of the 2021 IEEE Region 1 Technological Innovation Award for innovative contributions in machine learning techniques for the wireless domain. He was awarded the 2024 George. W. Thorn Award. He is also the recipient of the AFCEA International Meritorious Rising Star Award for achievement in engineering and the AFCEA 40 Under Forty award.
\end{IEEEbiography}

\end{document}